\documentclass[twocolumn,DIV25,9pt]{scrartcl}
\date{}
\pdfoutput=1

\usepackage{ctable}
\usepackage{float}
\usepackage{graphicx}
\usepackage{amssymb}
\usepackage{array}
\usepackage{url}
\usepackage{multirow}
\usepackage[round]{natbib}
\usepackage{algorithm}
\usepackage{algorithmic}
\usepackage{subfig}
\usepackage{gensymb}
\usepackage{palatino}
\usepackage{amsmath}
\usepackage{tweaklist}
\usepackage{fixltx2e}

\definecolor{linkblue}{rgb}{0, 0.3, 0.6}

\RequirePackage{fancyhdr}
\newtheorem{definition}{Definition}

\renewcommand{\itemhook}{\setlength{\topsep}{0pt}%
  \setlength{\itemsep}{0pt}}

\date{}
\begin{document}

\title{\flushleft{\fontsize{21}{12}\selectfont{}Fast Damage Recovery in Robotics with the T-Resilience Algorithm}}
% reviewers : Bongard, Burgard, Ijspeert, whiteson , Bentley?

\author{Sylvain Koos, Antoine Cully and Jean-Baptiste Mouret% <-this % stops a space
\thanks{Sylvain Koos, Antoine Cully and Jean-Baptiste Mouret are with the ISIR, Universit\'e Pierre et Marie Curie-Paris 6, CNRS UMR 7222, F-75252, Paris
Cedex 05, France. Contact: mouret@isir.upmc.fr}% <-this % stops a space
}

\maketitle

\thispagestyle{fancy}
\pagestyle{fancy}

\begin{abstract}
 \bfseries Damage recovery is critical for autonomous robots that need
 to operate for a long time without assistance. Most current methods
 are complex and costly because they require anticipating each
 potential damage in order to have a contingency plan ready. As an
 alternative, we introduce the T-resilience algorithm, a new algorithm
 that allows robots to quickly and autonomously discover compensatory
 behaviors in unanticipated situations. This algorithm equips the
 robot with a self-model and discovers new behaviors by learning to
 avoid those that perform differently in the self-model and in
 reality. Our algorithm thus does not identify the damaged parts but
 it implicitly searches for efficient behaviors that do not use
 them. We evaluate the T-Resilience algorithm on a hexapod robot that
 needs to adapt to leg removal, broken legs and motor failures; we
 compare it to stochastic local search, policy gradient and the
 self-modeling algorithm proposed by Bongard et al. The behavior of
 the robot is assessed on-board thanks to a RGB-D sensor and a SLAM
 algorithm. Using only 25 tests on the robot and an overall running
 time of 20 minutes, T-Resilience consistently leads to substantially
 better results than the other approaches.
\end{abstract}

%\maketitle
%legged robots \citep{Raibert1986, Hauser2008, Kajita2008}

\section{Introduction}
\label{sec:intro}
\begin{figure}
  \centering
  \subfloat[Normal state.]{\includegraphics[width=0.23\textwidth]{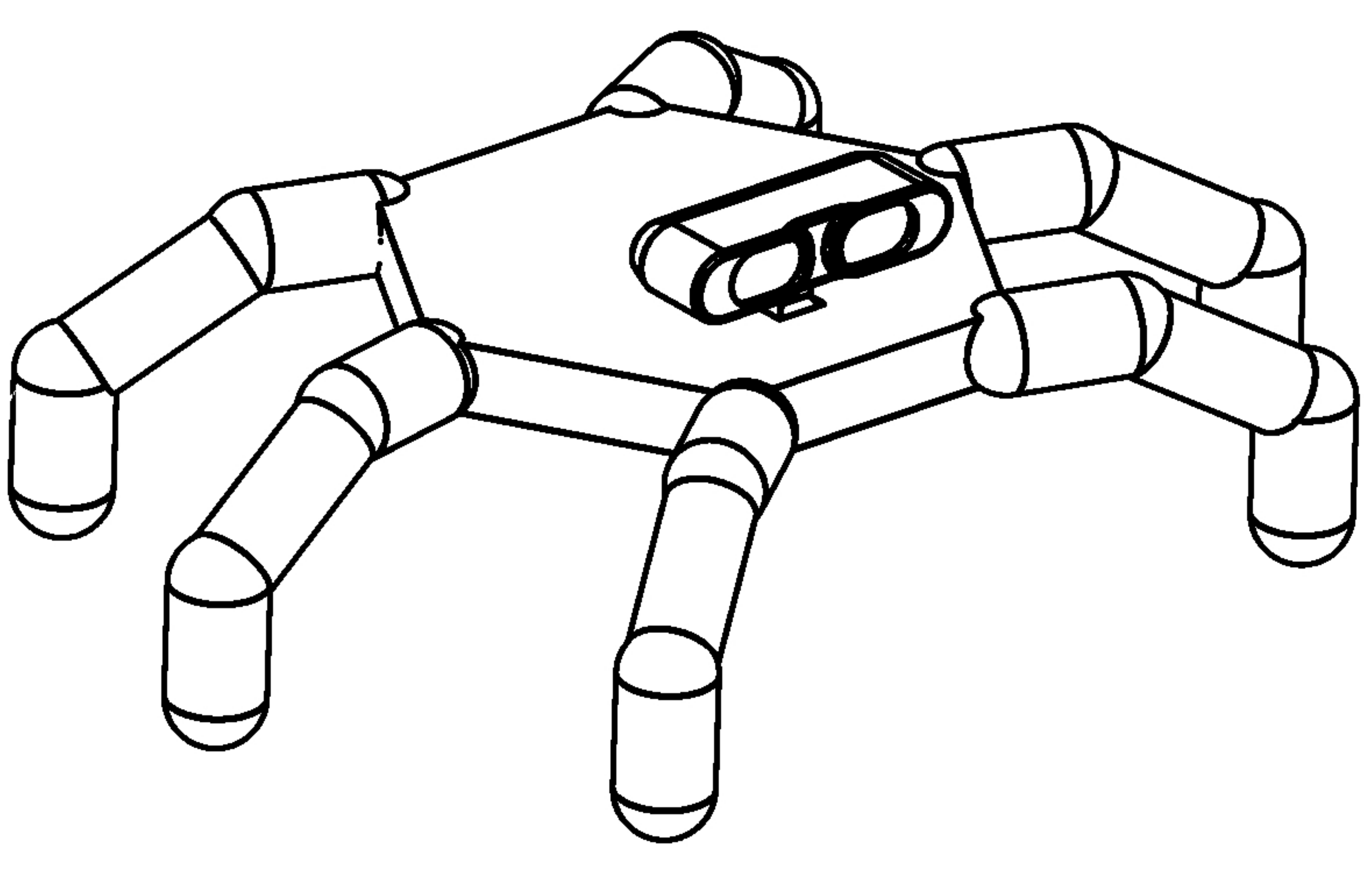}}
  \hfill
  \subfloat[Two legs ripped out.]{\includegraphics[width=0.23\textwidth]{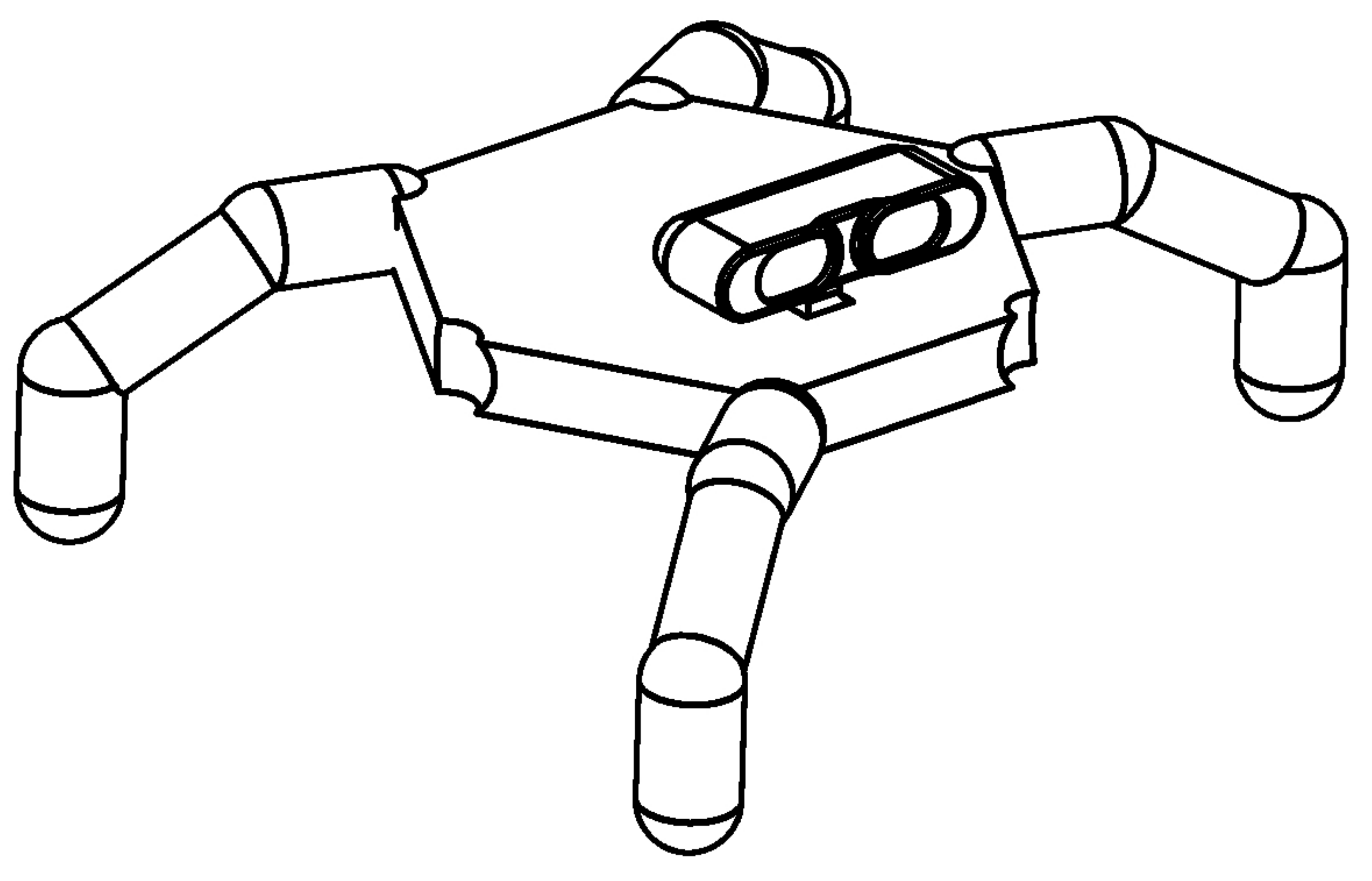}}
\\%\hfill
\subfloat[One broken leg.]{\includegraphics[width=0.23\textwidth]{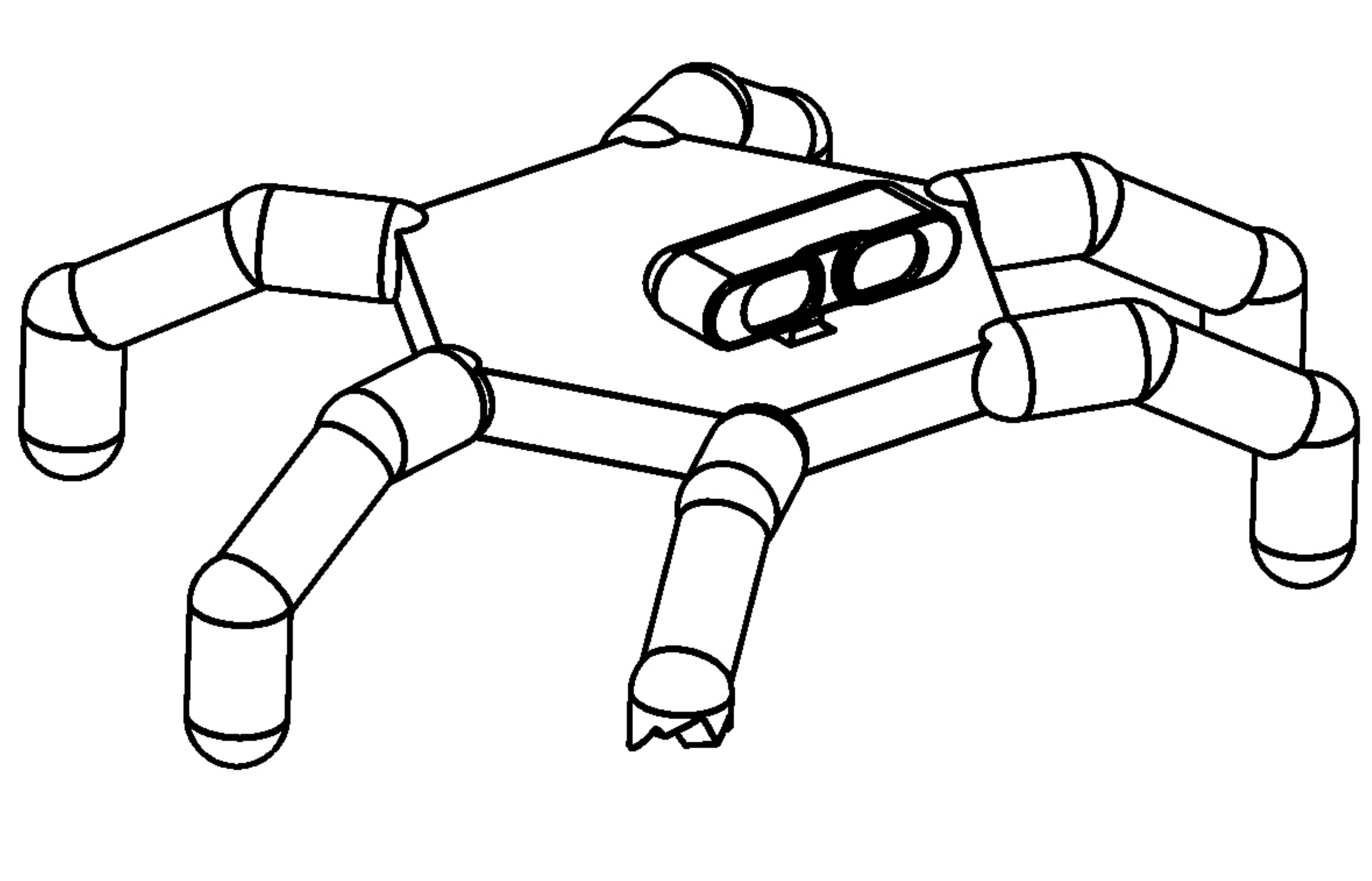}}
\hfill
\subfloat[Two unpowered motors.]{\includegraphics[width=0.23\textwidth]{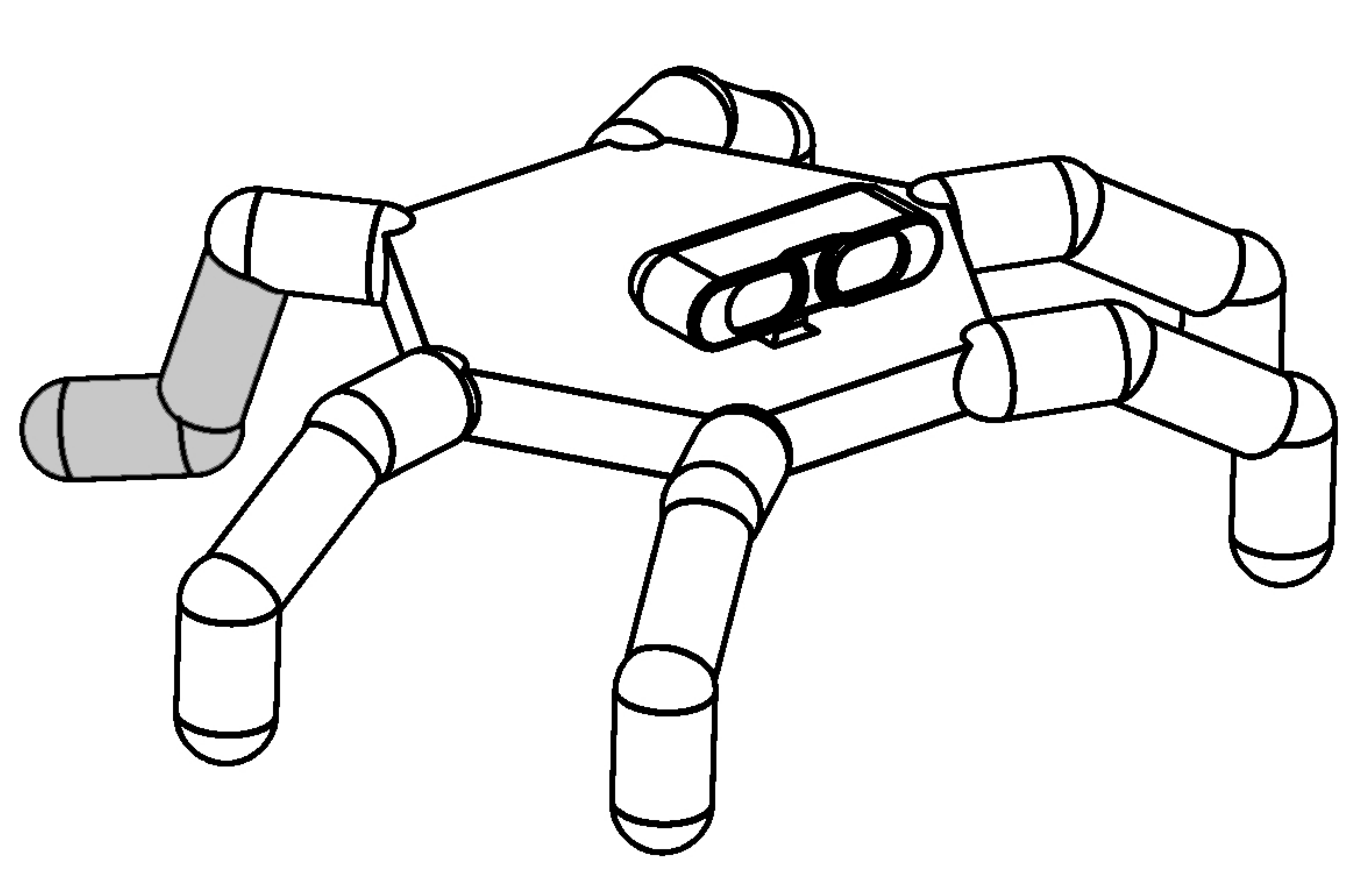}}
\caption{\label{fig:scenarios}Examples of situations in which an
  autonomous robot needs to discover a qualitatively new behavior to
  pursue its mission: in each case, classic hexapod gaits cannot be
  used. The broken leg example (c) is a typical damage that is hard to
  diagnose by direct sensing (because no actuator or sensor is
  damaged).}
\end{figure}

Autonomous robots are inherently complex machines that have to cope
with a dynamic and often hostile environment. They face an even more
demanding context when they operate for a long time without any
assistance, whether when exploring remote
places~\citep{Bellingham2007} or, more prosaically, in a house without
any robotics expert~\citep{Prassler2008}. As famously pointed out by
\cite{Corbato2007}, when designing such complex systems, ``[we should
  not] wonder \emph{if} some mishap may happen, but rather ask
\emph{what} one will do about it when it occurs''. In autonomous
robotics, this remark means that robots must be able to pursue their
mission in situations that have not been anticipated by their
designers. Legged robots clearly illustrate this need to handle the
unexpected: to be as versatile as possible, they involve many moving
parts, many actuators and many sensors~\citep{Kajita2008}; but they
may be damaged in numerous different ways. These robots would
therefore greatly benefit from being able to autonomously find a new
behavior if some legs are ripped off, if a leg is broken or if one
motor is inadvertently disconnected~(Fig.~\ref{fig:scenarios}).

Fault tolerance and resilience are classic topics in robotics and
engineering. The most classic approaches combine intensive testing
with redundancy of components \citep{Visinsky1994,
  koren2007fault}. These methods undoubtedly proved their usefulness
in space, aeronautics and numerous complex systems, but they also are
expensive to operate and to design. More importantly, they require the
identification of the faulty subsystems and a procedure to bypass
them, whereas both operations are difficult for many kinds of faults
-- for example mechanical failures. Another classic approach to fault
tolerance is to employ robust controllers that can work in spite of
damaged sensors or hardware
inefficiencies~\citep{Goldberg2001,Caccavale2002,
  Qu2003,Lin2007}. Such controllers usually do not require diagnosing
the damage, but this advantage is tempered by the need to integrate the
reaction to all faults in a single controller. Last, a robot can embed
a few pre-designed behaviors to cope with anticipated potential
failures~\citep{gorner2010analysis,jakimovski2010situ,
  mostafa2010alternative,schleyer2010adaptable}.  For instance, if a
hexapod robot detects that one of its legs is not reacting as
expected, it can drop it and adapt the position of the other legs
accordingly~\citep{jakimovski2010situ,mostafa2010alternative}.

An alternative line of thought is to \emph{let the robot learn on its
  own} the best behavior for the current situation. If the learning
process is open enough, then the robot should be able to discover new
compensatory behaviors in situations that have not been foreseen by
its designers. Numerous learning systems have been experimented in
robotics (for reviews, see \cite{Connell1993, Argall2009,
  NguyenTuong2011, Kober2012}), with different levels of openness and
various a priori constraints. Many of them primarily aim at
automatically tuning controllers for complex
robots~\citep{kohl2004policy,tedrake2005learning,sproewitz2008learning,hemker2009efficient}
whereas only a handful of these systems has been explicitly tested in
situations in which a robot needs to adapt itself to unexpected
situations~\citep{mahdavi2003evolutionary,berenson2005hardware,bongard2006resilient}.

Finding the behavior that maximizes performance in the current
situation is a \emph{reinforcement learning
  problem}~\cite{Sutton1998}, but classic reinforcement learning
algorithms (e.g. TD-Learning, SARSA, ...) are designed for discrete
state spaces~\citep{Sutton1998,Togelius2009}. They are therefore hard
to use when learning continuous behaviors such as locomotion
patterns. Policy gradient algorithms~\citep{kohl2004policy,
  peters2008reinforcement,Peters2010} are reasonably fast learning
algorithms that are better suited for robotics (authors typically
report learning time of 20 minutes to a few hours), but they are
essentially limited to a local search in the parameter space: they
lack the openness of the search that is required to cope with truly
unforeseen situations. Evolutionary Algorithms (EAs)~\citep{deb2001,
  de2006evolutionary} can optimize reward functions in larger, more
open search spaces (e.g. automatic design of neural networks, design
of structures)~\citep{Grefenstette1999, Heidrich-Meisner2009a,
  Togelius2009, Doncieux2011,Hornby2011,Whiteson2012}, but this
openness is counterbalanced by substantially longer learning time
(according to the literature, 2 to 10 hours for simple robotic
behaviors).

%%  They demonstrated their idea on a
%% starfish-like quadruped robot that overcome the loss of a
%% leg.

All policy gradient and evolutionary algorithms spend most of their
running time in evaluating the quality of controllers by testing them
on the target robot. Since, contrary to simulation, reality cannot be
sped up, their running time can only be improved by finding strategies
to evaluate fewer candidate solutions on the robot. In their
``starfish robot'' project, \cite{bongard2006resilient} designed a
general approach for resilience that makes an important step in this
direction. The algorithm of Bongard et al. is divided into two stages:
(1) automatically building an internal simulation of the whole robot
by observing the consequences of a few elementary actions (about 15 in
the demonstrations of the paper) -- this internal simulation of the
whole body is called a \emph{self-model}\footnote{ Following the
  literature in
  psychology~\citep{Metzinger2004,Metzinger2007,Vogeley1999} and
  artificial intelligence~\citep{bongard2006resilient,Holland2003}, we
  define a self-model as a forward, internal model of the \emph{whole
    body} that is accessible to introspection and instantiated in a
  model of the environment. In the present paper, we only consider a
  minimal model of the environment (a horizontal
  plane).}~\citep{Metzinger2004,Metzinger2007,Vogeley1999,bongard2006resilient,Holland2003,Hoffmann2010};
(2) launching \emph{in this simulation} an EA to find a new
controller. In effect, this algorithm transfers most of the learning
time to a computer simulation, which makes it increasingly faster when
computers are improved~\citep{moore1975progress}.

Bongard's algorithm highlights how mixing a self-model with a learning
algorithm can reduce the time required for a robot to adapt to an
unforeseen situation. Nevertheless, it has a few important
shortcomings. First, actions and models are undirected: the algorithm
can ``waste'' a lot of time to improve parts of the self-model that
are irrelevant for the task. Second, it is computationally expensive
because it includes a full learning algorithm (the second stage, in
simulation) and an expensive process to select each action that is
tested on the robot. Third, there is often a ``reality gap'' between a
behavior learned in simulation and the same behavior on the target
robot~\citep{jakobi1995noise,Zagal2004,2012ACLI2214}, but nothing is
included in Bongard's algorithm to prevent such gap to happen: the
controller learned in the simulation stage may not work well on the
real robot, even if the self-model is accurate. Last, one can
challenge the relevance of calling into question the full self-model
each time an adaptation is required, for instance if an adaptation is
only temporarily useful.

%\footnote{For example, humans seem
%  reluctant to permanently change important parts of their self-model,
%  as witnessed by the observation that most amputees dream about
%  themselves in their intact body~\citep{Mulder2008}. Nevertheless
%  humans can quickly and temporarily adapt their body image, as shown
%  by the rubber hand experiments~\citep{Botvinick1998,Hoffmann2010}.},

In the present paper we introduce a new resilience algorithm that
overcomes these shortcomings while still performing most of the search
in a simulation of the robot. Our algorithm works with any
parametrized controller and it is especially efficient on modern,
multi-core computers. More generally, it is designed for situations in
which:

\begin{itemize}
\item behaviors optimized on the undamaged robot are not efficient
  anymore on the damaged robot (otherwise, adaptation is useless) and
  qualitatively new behavior is required (otherwise, local search
  algorithms should perform better);
\item the robot can only rely on internal measurements of its state
  (truly autonomous robots do not have access to perfect, external
  sensing systems);
 \item some damages cannot be observed or measured directly (otherwise
   more explicit methods may be more efficient).
\end{itemize}

Our algorithm is inspired by the ``transferability
approach''~\citep{2012ACLI2214,Mouret2012}, whose original purpose is
to cross the ``reality gap'' that separates behaviors optimized in
simulation to those observed on the target robot. The main proposition
of this approach is to make the optimization algorithm aware of the
limits of the simulation. To this end, a few controllers are
transferred during the optimization and a regression algorithm (e.g. a
SVM or a neural network) is used to approximate the function that maps
behaviors in simulation to the difference of performance between
simulation and reality. To use this approximated \emph{transferability
  function}, the single-objective optimization problem is transformed
into a multi-objective optimization in which both performance in
simulation and transferability are maximized. This optimization is
typically performed with a stochastic multi-objective optimization
algorithm but other optimization algorithms are conceivable.

As this paper will show, the same concepts can be applied to design a
fast adaptation algorithm for resilient robotics, leading to a new
algorithm that we called ``T-Resilience'' (for Transferability-based
resilience). If a damaged robot embeds a simulation of itself, then
behaviors that rely on damaged parts will not be transferable: they
will perform very differently in the self-model and in reality. During
the adaptation process, the robot will thus create an approximated
transferability function that classifies behaviors as ``working as
expected'' and ``not working as expected''. Hence the robot will
possess an ``intuition'' of the damages but it will not explicitly
represent or identify them. By optimizing both the transferability and
the performance, the algorithm will look for the most efficient
behaviors among those that only use the reliable parts of the
robots. The robot will thus be able to sustain a functioning behavior
when damage occurs by learning to avoid behaviors that it is unable to
achieve in the real world. Besides this damage recovery scenario, the
T-Resilience algorithm opens a new class of adaptation algorithms that
benefit from Moore's law by transferring most of the adaptation time
from real experiments to simulations of a self-model.

We evaluate the T-Resilience algorithm on an 18-DOFs hexapod robot
that needs to adapt to leg removal, broken legs and motor failures; we
compare it to stochastic local search~\citep{hoos2005stochastic},
policy gradient~\citep{kohl2004policy} and Bongard's
algorithm~\citep{bongard2006resilient}. The behavior on the real robot
is assessed on-board thanks to a RGB-D sensor coupled with a
state-of-the-art SLAM algorithm~\citep{endres12icra}.

\section{Learning for resilience}
\begin{figure}
\centering
\includegraphics[width=0.7\linewidth]{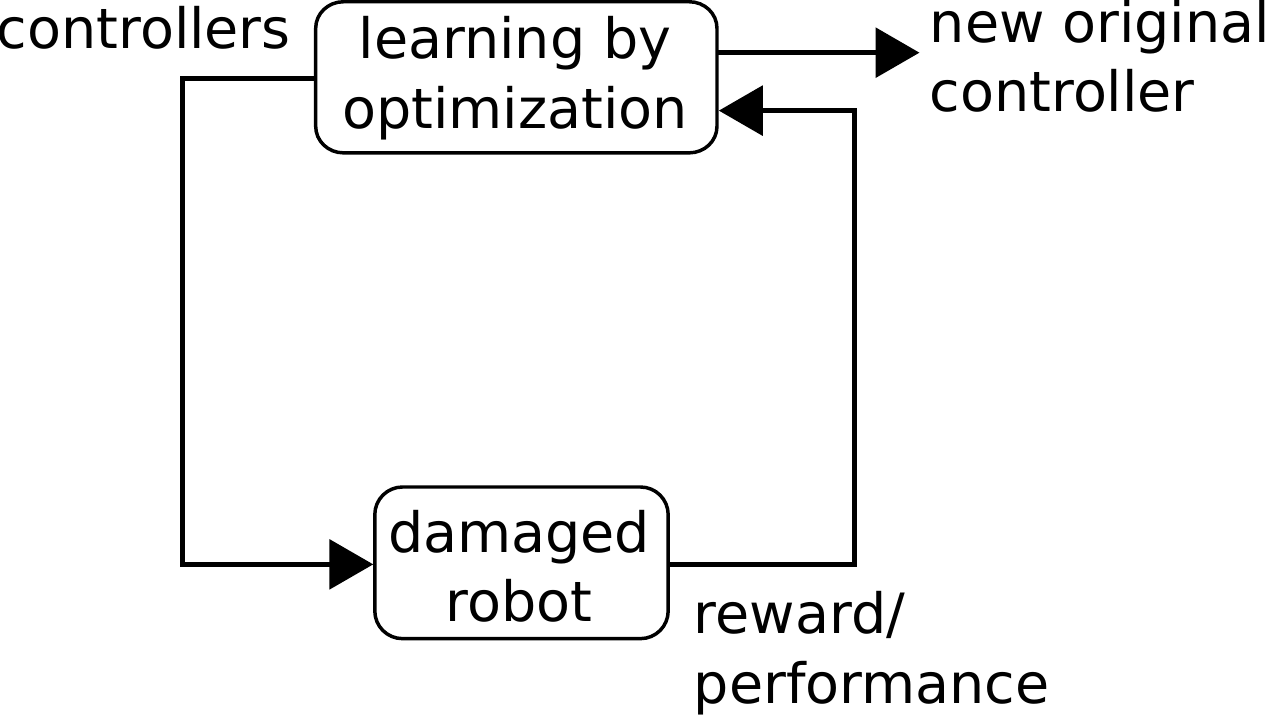}
\caption{Principle of resilience processes based on policy gradient. Controllers are optimized by measuring rewards on the robot.}
\label{fig:approaches_policy}
\end{figure}

Discovering a new behavior after a damage is a particular case of
\emph{learning} a new behavior, a question that generates an abundant
literature in artificial intelligence since its
beginnings~\citep{Turing1950}. We are here interested in reinforcement
learning algorithms because we consider scenarios in which evaluating
the performance of a behavior is possible but the optimal behavior is
unknown. However, classic reinforcement learning algorithms are
primarily designed for discrete states and discrete
actions~\citep{Sutton1998,Peters2010}, whereas autonomous robots have
to solve many continuous problems (e.g. motor control). Two
alternative families of methods are currently prevalent for continuous
reinforcement learning in robotics (table \ref{table:direct_methods}):
policy gradient methods and evolutionary algorithms. These two
approaches both rely on optimization algorithms that directly optimize
parameters of a controller by measuring the overall performance of the
robot (Fig.~\ref{fig:approaches_policy}); learning is thus here
regarded as an optimization of these parameters.

\subsection{Policy gradient methods}
\ctable[star,
  cap = {},
  caption = {Typical examples of learning algorithms that have been used on legged robots.},
  label   = {table:direct_methods},
]{lccccccc}{
  \tnote[]{$^\star$Behavior used to initialize the learning algorithm.}
  \tnote[]{$^\dag$ DOFs: number of controlled degrees of freedom.} 
  \tnote[]{$^\ddag$ param: number of learned control parameters.}
  \tnote[]{$^1$ Nelder-Mead descent.\ \ $^2$ Powell method.\ \ $^3$ Design and Analysis of Computer Experiments.\ \ $^4$ Multi-agent reinforcement learning}% Hebbian learning. 
}{
  \FL
  approach/article & starting beh.$\ ^\star$ & learning time & robot & DOFs$^\dag$ & param.$^\ddag$ & reward\\%  & real tests
  \FL
  Policy Gradient Methods &&&&&&\\
  \cline{1-1}
  \cite{kimura2001reinforcement} & no info.  & 80 min. & quadruped & 8 & 72 & internal\\% & 10000
  \cite{kohl2004policy} & walking  & 3 h & quadruped & 12 & 12 & external\\%external checked(landmark = vision ext)! % external? & 1035
  \cite{tedrake2005learning} & standing  & 20 min. & bidepal & 2 & 46 & internal\\% & 1000 already stable controllers
  \FL
  % Natural Actor-Critic &&&&&\\ % cf. Peters, Schaal 2OO8 (idem pour critiques de DPS)
  % Guenter et al. Reinforcement Learning for Imitating Constrained Reaching Movements
  % Ueno et al. Fast and Stable Learning of Quasi-Passive Dynamic Walking by an Unstable Biped Robot based on Off-Policy Natural Actor-Critic
  % \cline{1-1}
  % \FL
  Evolutionary Algorithm &&&&&&\\
  \cline{1-1}
  \cite{chernova2004evolutionary} & random & 5 h & quadruped & 12 & 54 & external\\ %  & 4000
  \cite{zykov2004evolving} & random & 2 h & hexapod & 12 & 72 & external\\ %  & 500
  \cite{berenson2005hardware} & random & 2 h & quadruped & 8 & 36 & external\\ % external= aucune mention de moyen intern  & 200
  \cite{hornby2005autonomous} & non-falling & 25h & quadruped & 19 & 21 & internal\\ % & 30 * 500
  \cite{mahdavi2006innately}  & random & 10 h & snake & 12 & 1152 & external \\ %  600
  \cite{barfoot2006experiments} & random & 10 h & hexapod & 12 & 135 & external \\
  \cite{Yosinski2011} & random & 2 h & quadruped & 9 & 5 & external\\% random search equivalent to others !!!  & 540
  \FL
  Others &&&&&&\\
  \cline{1-1}
  \cite{weingarten2004automated}$\ ^1$ & walking & $>$ 15 h & hexapod (Rhex-like) & 6 & 8 & external\\ % external? & 300-500
  % \cite{manoonpong2007adaptive}$\ ^2$ & walking & 14 sec. & biped & 3 & 6 & internal\\ %  & 3
  \cite{sproewitz2008learning}$\ ^2$ & random & 60 min. & quadruped & 8 & 5 & external\\ %  80-180 % only 1 (quadrupedal)
  \cite{hemker2009efficient}$\ ^3$ & walking & 3-4 h & biped & 24 & 5 & external \\%  & 80
  \cite{barfoot2006experiments}$\ ^4$ & random & 1h & hexapod & 12 & 135 & external
  \LL
}

Policy gradient methods~\citep{sutton2000policy,
  peters2008reinforcement, Peters2010} use iterative stochastic
optimization algorithms to find a local extremum of the reward
function. The search starts with a controller that can be generated at
random, designed by the user or inferred from a demonstration. The
algorithm then iteratively modifies the parameters of the controller
by estimating gradients in the control space and applying slight
changes to the parameters.

Typical policy gradient methods iterate the following steps:
\begin{itemize}
\item generation of $N$ controllers in the neighborhood of the current
  vector of parameters (by variating one or multiple parameter values
  at once);
\item estimation of the gradient of the reward function in the control space;
\item modification of parameter values according to the gradient information.
\end{itemize}

These steps are iterated until a satisfying controller is found or
until the process converges. Policy gradient algorithms essentially
differ in the way gradient is estimated. The most simple way is the
finite-difference method, which independently estimates the local
gradient of the reward function for each
parameter~\citep{kohl2004policy,tedrake2005learning}: considering a
given parameter, if higher (resp. lower) values lead to higher rewards
on average on the $N$ controllers tested during the current iteration,
the value of the parameter is increased (resp. decreased) for the next
iteration. Such a simple method for estimating the gradient is
especially efficient when parameters are mostly independent. Strong
dependencies between the parameters often require more sophisticated
estimation techniques.

Policy gradient algorithms have been successfully applied to
locomotion tasks in the case of
quadruped~\citep{kimura2001reinforcement, kohl2004policy} and biped
robots~\citep{tedrake2005learning} but they typically require numerous
evaluations on the robot, most of the times more than 1000 trials in a
few hours (table \ref{table:direct_methods}). To make learning
tractable, these examples all use carefully designed controllers with
only a few degrees of freedom. They also typically start with
well-chosen initial parameter values, making them efficient algorithms
for imitation learning when these values are extracted from a
demonstration by a human~\citep{Kober2010}. Recent results on the
locomotion of a quadruped robot suggest that using random initial
controllers would likely require many additional experiments on the
robot~\citep{Yosinski2011}. Consistent results have been reported on
biped locomotion with computer simulations using random initial
controllers that make the robot fall~\citep{nakamura2007reinforcement}
(about 10 hours of learning for 11 control parameters).
% Policy gradient methods are not efficient when exploring plateaus of the reward
% function~\citep{peters2008reinforcement}, which is likely to happen if considering random initial controllers.

\subsection{Evolutionary Algorithms}
Evolutionary Algorithms (EAs)~\citep{deb2001,de2006evolutionary} are
another family of iterative stochastic optimization methods that
search for the optima of function~\citep{Grefenstette1999,
  Heidrich-Meisner2009a}. They are less prone to local optima than
policy gradient algorithms and they can optimize arbitrary structures
(neural networks, fuzzy rules, vector of parameters,
...)~\citep{Doncieux2011,Hornby2011,Mouret2012div,Whiteson2012}.

While there exists many variants of EAs, the vast majority of them
iterate the following steps:
\begin{itemize}
\item (first iteration only) random initialization of a population of
  candidate solutions;
\item evaluation of the performance of each controller of the
  population (by testing the controller on the robot);
\item ranking of controllers;
\item selection and variation around the most efficient controllers to
  build a new population for the next iteration.
\end{itemize}

Learning experiments with EAs are reported to require many hundreds of
trials on the robot and to last from two to tens of hours (table
\ref{table:direct_methods}). EAs have been applied to quadruped
robots~\citep{hornby2005autonomous,Yosinski2011}, hexapod
robots~\citep{zykov2004evolving,barfoot2006experiments} and
humanoids~\citep{katic2003survey,palmer2009}. EAs have also been used
in a few studies dedicated to resilience, in particular on a
snake-like robot with a damaged body~\citep{mahdavi2003evolutionary}
(about 600 evaluations/10 hours) and on a quadrupedal robot that
breaks one of its leg~\citep{berenson2005hardware} (about 670
evaluations/2 hours).

Aside from these two main types of approaches, several authors
proposed to use other black-box optimization algorithms: global
methods like Nelder-Mead descent~\citep{weingarten2004automated},
local methods like Powell's method~\citep{sproewitz2008learning} or
surrogate-based optimization~\citep{hemker2009efficient}. Published
results are typically obtained with hundreds of evaluations on the
robot, requiring several hours (table~\ref{table:direct_methods}).

Regardless of the optimization technique, reward functions are, in
most studies, evaluated with external tracking devices
(table~\ref{table:direct_methods}, last column). While this approach
is useful when researchers aims at finding the most efficient
controllers (e.g. \cite{kohl2004policy,sproewitz2008learning,
  hemker2009efficient}), learning algorithms that target adaptation
and resilience need to be robust to the inaccuracies and constraints
of on-board measurements.

\subsection{Resilience based on self-modeling}
\label{sec:review-self-model}

\begin{figure}
\centering
\includegraphics[width=\linewidth]{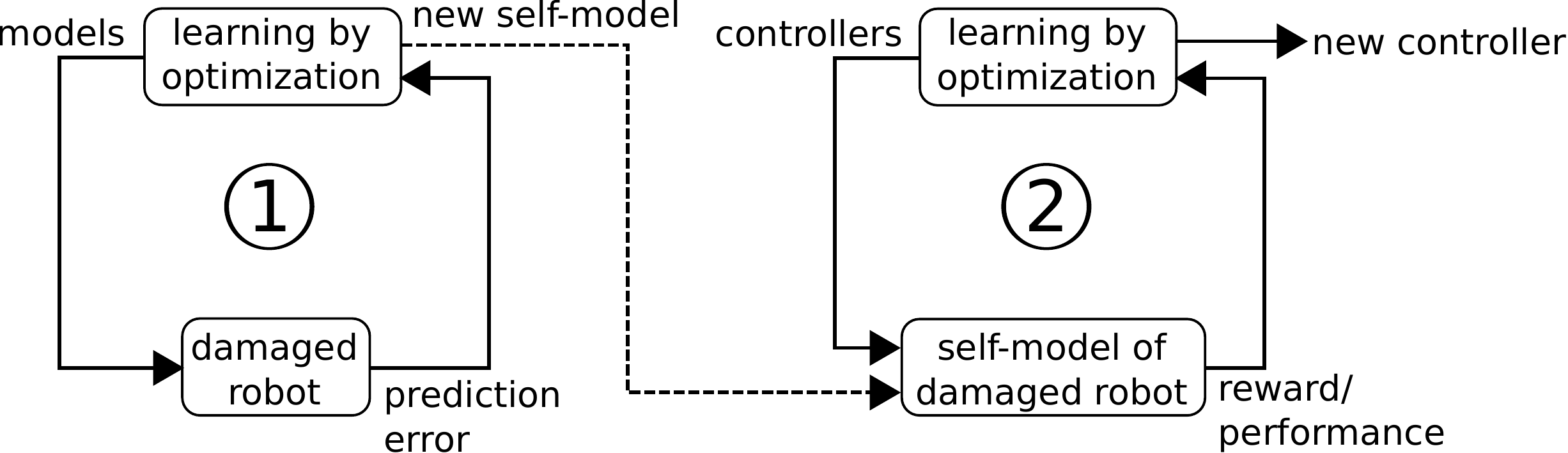}
\caption{Principle of Bongard's algorithm. (1) A self-model is learned
  by testing a few actions on the damaged robot. (2) This self-model
  is next used as a simulation in which a new controller is
  optimized.}
\label{fig:approaches_bongard}
\end{figure}

Instead of directly learning control parameters,
\cite{bongard2006resilient} propose to improve the resilience of
robots by equipping robots with a \emph{self-model}. If a disagreement
is detected between the self-model and observations, the proposed
algorithm first infers the damages by chosing motor actions and
measuring their consequences on the behavior of the robot; the
algorithm then relies on the updated model of the robot to learn a new
behavior. This approach has been successfully tested on a
starfish-like quadrupedal
robot~\citep{bongard2006resilient,zykov2008morphological}. By adapting
its self-model, the robot manages to discover a new walking gait after
the loss of one of its legs.

In Bongard's algorithm, the identification of the self-model is based
on an active learning loop that is itself divided into an \emph{action
  selection loop} and a \emph{model selection loop}
(Fig.~\ref{fig:approaches_bongard}). The action
selection loop aims at selecting the action that will best distinguish
the models of a population of candidate models. The model selection
loop looks for the models that best predict the outcomes of the
actions as measured on the robot. In the ``starfish''
experiment~\citep{bongard2006resilient}, the following steps are
repeated:
\begin{itemize}
\item[1.1.] action selection (\emph{exploration}):
\begin{itemize}
\item each of the 36 possible actions is tested on each of the $16$
  candidate models to observe the orientation of robot's body
  predicted by the model;
\item the action for which models of the population disagree at most
  is selected;
\item this action is tested on the robot and the corresponding exact
  orientation of robot's body is recorded by an external camera;
\end{itemize}
\item[1.2.] model selection loop (\emph{estimation}):
\begin{itemize}
\item a stochastic optimization algorithm (an EA) is used to optimize the population of models so that they accurately predict what was measured with the robot, for each tested action;
\item if less than $15$ actions have been performed, the action selection loop is started again.
\end{itemize}
\end{itemize}
Once the $15$ actions have been performed, the best model found so far is
used to learn a new behavior using an EA:
\begin{itemize}
\item [2.] controller optimization (\emph{exploitation}):
\begin{itemize}
\item a stochastic optimization algorithm (an EA) is used to optimize
  a population controllers so that they maximize forward displacement
  within the simulation of the self-model;
\item the best controller found in the simulation is transferred to
  the robot, making it the new controller.
\end{itemize}
\end{itemize}
The population of models is initialized with the self-model that
corresponds to the morphology of the undamaged robot. Since the
overall process only requires $15$ tests on the robot, its speed
essentially depends on the performance of the employed
computer. Significant computing times are nonetheless required for the
optimization of the population of models.

In the results reported by \cite{bongard2006resilient}, only half of
the runs led to correct self-models. As Bongard's. approach implies
identifying a full model of the robot, it would arguably require many
more tests to converge in most cases to the right morphology. For
comparison, results obtained by the same authors but in a simulated
experiment required from 600 to 1500 tests to consistently identify
the model~\citep{bongard2005nonlinear}. It should also be noted that
these authors did not measure the orientation of robot's body with
internal sensors, whereas noisy internal measurements could
significantly impair the identification of the model. Other authors
experimented with self-modeling process similar to the one of Bongard
et al., but with a humanoid robot~\citep{zagal2009self}. Preliminary
results suggest that thousands of evaluations on the robot would be
necessary to correctly identify 8 parameters of the global
self-model. Alternative methods have been proposed to build
self-models for robots and all of them require numerous tests, e.g. on
a manipulator arm with about 400 real tests~\citep{sturm2008adaptive}
or on a hexapod robot with about 240 real
tests~\citep{parker2009punctuated}. Overall, experimental costs for
building self-models appear expensive in the context of resilience
applications in both the number of tests on the real robot and in
computing time.

Furthermore, controllers obtained by optimizing in a simulation -- as
does the algorithm proposed by Bongard et al. -- often do no work as
well on the real robot than in
simulation~\citep{2012ACLI2214,Zagal2004,jakobi1995noise}. In effect,
this classic problem has been observed in the starfish
experiments~\cite{bongard2006resilient}. In these experiments, it
probably originates from the fact that the identified self-model
cannot perfectly model every detail of the real world (in particular,
slippage, friction and very dynamic behaviors).

\subsection{Concluding thoughts}

Based on this short survey of the literature, two main thoughts can be drawn:
\begin{enumerate}
\item[1.] Policy gradient methods and EAs can both be used to discover
  original behaviors on a damaged robot; nevertheless, when they don't
  start from already good initial controllers, they require a high
  number of real tests (at least a few hundred), which limits the
  speed of the resulting resilience process.
\item[2.] Methods based on self-modeling are promising because they
  transfer some of the learning time to a simulation; however building
  an accurate global model of the damaged robot requires many real
  tests; reality gap problems can also occur between the behavior
  learned with self-model and the real, damaged robot.
\end{enumerate}

\section{The T-Resilience algorithm}
\subsection{Concept and intuitions}
Following Bongard et al., we equip our robot with a self-model. A
direct consequence is that detecting the occurrence of a damage is
facilitated: if the observed performance is significantly different
from what the self-model predicts, then the robot needs to start a
recovery process to find a better behavior. Nevertheless, contrary to
Bongard et al., we propose that a damaged robot discovers new original
behaviors \emph{using the initial, hand-designed self-model}, that is
without updating the self-model. Since we do not attempt to diagnose
damages, the solved problem is potentially easier than the one solved
by Bongard et al; we therefore expect our algorithm to perform
faster. This speed increase can, however, comes at the price of
slightly less efficient post-damage behaviors.

The model of the undamaged robot is obviously not accurate because it
does not model the damages. Nonetheless, since damages can't radically
change the overall morphology of the robot, this ``undamaged''
self-model can still be viewed as a reasonably accurate model of the
damaged robot. Most of the degrees of freedom are indeed correctly
positionned, the mass of components should not change much and the
body plan is most probably not radically altered.

Imperfect simulators and models are an almost unavoidable issue when
robotic controllers are first optimized in simulation then transferred
to a real robot. The most affected field is probably evolutionary
robotics because of the emphasis on opening the search space as much
as possible: behaviors found within the simulation are often not
anticipated by the designer of the simulator, therefore it's not
surprising that they are often wrongly simulated. Researchers in
evolutionary robotics explored three main ideas to cross this
``reality gap'': (1) automatically improving
simulators~\citep{bongard2006resilient,Pretorius2012,Klaus2012}, (2)
trying to prevent optimized controllers from relying on the unreliable
parts of the simulation (in particular, by adding
noise)~\citep{jakobi1995noise}, and (3) model the difference between
simulation and reality~\citep{Hartland2006, 2012ACLI2214}.

\begin{figure*}[ht!]
\centering
\includegraphics[width=0.95\textwidth]{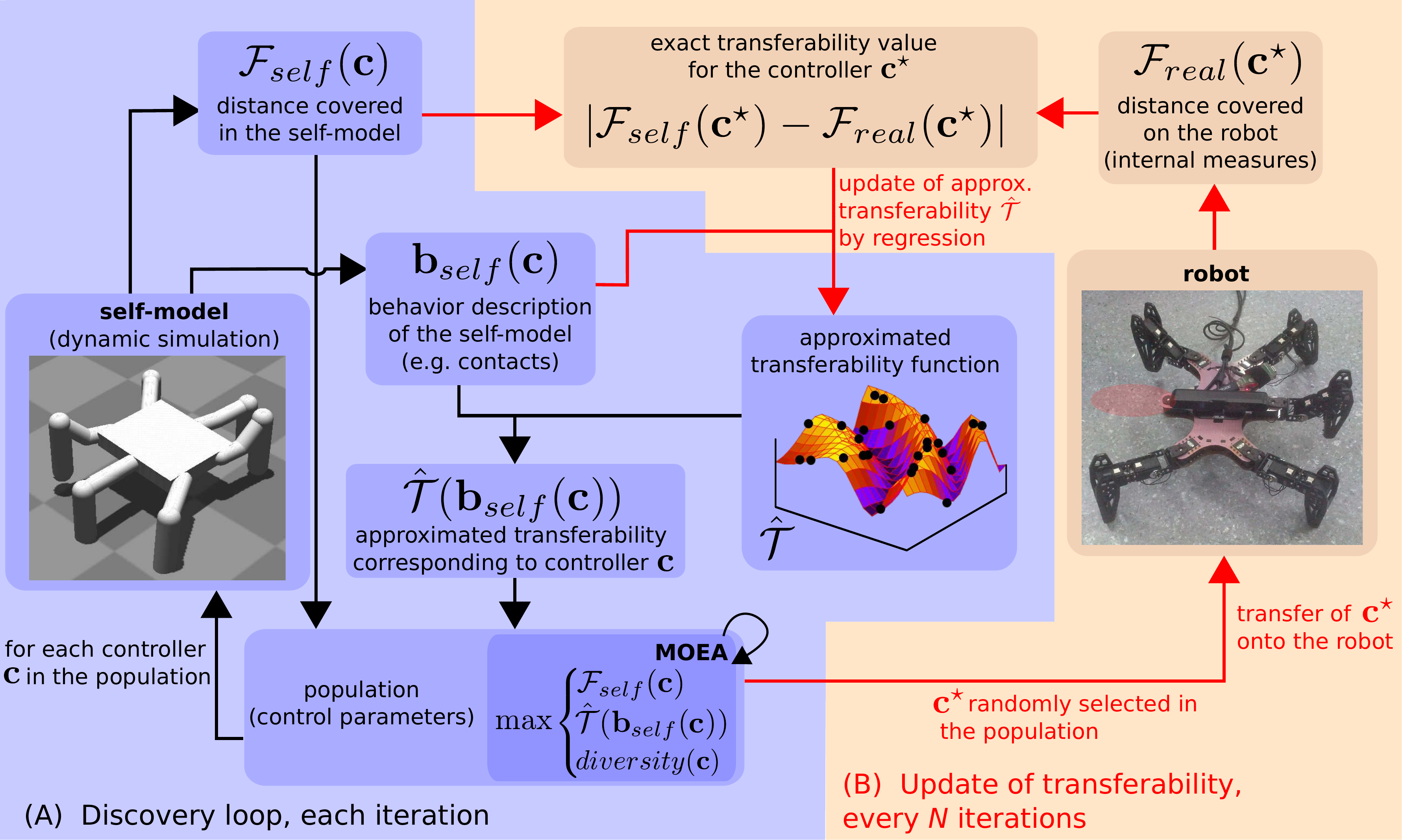}
\caption{Schematic view of the T-Resilience algorithm (see
  algorithm~\ref{algo:transf} for an algorithmic view). (A) Discovery
  loop: each controller of the population is evaluated with the
  self-model. Its transferability score is approximated according to
  the current model $\hat{\mathcal{T}}$ of the exact transferability
  function $\mathcal{T}$. (B) Transferability update: every $N$
  iterations, a controller of the population is randomly selected and
  transferred onto the real robot. The model of the transferability
  function is next updated with the data generated during the
  transfer.}
\label{fig:transferability}
\end{figure*}

Translated to resilient robotics, the first idea is equivalent to
improving or adapting the self-model, with the aforementioned
shortcomings (sections \ref{sec:intro} and
\ref{sec:review-self-model}). The second idea corresponds to
encouraging the robustness of controllers so that they can deal with
an imperfect simulation. It could lead to improvements in resilient
robotics but it requires that the designer anticipates most of the
potential damages. The third idea is more interesting for resilient
robotics because it acknowledges that simulations are never perfect
and mixes reality and simulation during the optimization. Among the
algorithms of this family, the recently-proposed transferability
approach~\citep{2012ACLI2214} explicitly searches for high-performing
controllers that work similarly in both simulation and reality. It led
to successful solutions for quadruped robot (2 parameters to optimize)
and for a Khepera-like robot in a T-maze (weights of a feed-forward
neural networks to optimize)~\citep{2012ACLI2214}.

%% \begin{figure}
%% \centering
%% \includegraphics[width=\linewidth]{img/schema_transf.pdf}
%% \caption{Principle of the T-Resilience algorithm. (A) Controllers are
%%   optimized with a stochastic multi-objective optimization algorithm
%%   that optimizes both rewards predicted by the self-model of the
%%   undamaged robot and approximated transferability values. (B) The
%%   approximated transferability function is periodically updated with a
%%   regression algorithm by transferring a controller onto the robot.}
%% \label{fig:approaches_transf}
%% \end{figure}

The main assumption of the transferability approach is that some
transferable behaviors exist in the search space. Although formulated
in the context of the reality gap, this assumption holds well in
resilient robotics. For instance, if a hexapod robot breaks a leg,
then gaits that do not critically rely on this leg should lead to
similar trajectories in the self-model and on the damaged robot. Such
gaits are numerous: those that make the simulated robot lift the
broken leg so that it never hits the ground; those that make the robot
walk on its ``knees''; those that are robust to leg damages because
they are closer to crawling than walking. Similar ideas can be found
for most robots and for most mechanical and electrical damages,
provided that there are different ways to achieve the mission. For
example, any redundant robotic manipulator with a blocked joint should
be able to follow a less efficient but working trajectory that does
not use this joint.

The transferability approach captures the differences between the
self-model and reality through the \emph{transferability
function}~\citep{Mouret2012}:
\begin{definition}[transferability function]
 A \emph{transferability function} is a function that maps, for the
 whole search space, descriptors of solutions (e.g. control parameters,
 or behavior descriptors) to a \emph{transferability score} that
 represents how well the simulation matches the reality.
\end{definition}

This function is usually not accessible but it can be learned with a
regression algorithm (neural networks, support vector machines, etc.)
by recording the behavior of a few controllers in reality and in
simulation.

\subsection{T-Resilience}
\begin{algorithm*}[h]
\caption{T-Resilience ($T$ real tests)}
\label{algo:transf}
\begin{algorithmic}
\STATE $pop \leftarrow \{c^1, c^2, \ldots, c^S\}$ (randomly generated)
\vspace{5pt}
\FOR{$i = 1 \to T$}
\vspace{5pt} 
\STATE \begin{tabular}{@{}l|l}random selection of $c^*$ in $pop$ &\\
transfer of $c^*$ on the robot&\\
estimation of performance $\mathcal{F}_{real}(c^*)$ using internal measurements & (B) Update of\\
estimation of the exact transferability value $\Big|\mathcal{F}_{self}(c^*) - \mathcal{F}_{real}(c^*)\Big|$& transferability\\
update of the approximated transferability function $\hat{\mathcal{T}}$&
\vspace{5pt} \\
$N$ iterations of MOEA on $pop$ & (A) Discovery loop\vspace{5pt}
\end{tabular}
\ENDFOR
\vspace{5pt}
\STATE selection of the new controller
\end{algorithmic}
\end{algorithm*}

To cross the reality gap, the transferability approach essentially
proposes optimizing both the approximated transferability and the
performance of controllers with a stochastic multi-objective
optimization algorithm. This approach can be adapted to make a robot
resilient by seeing the original, ``un-damaged'' self-model as an
inaccurate simulation of the damaged robot, and if the robot only uses
internal measurements to evaluate the discrepancies between
predictions of the self-model and measures on the real
robot. Resilient robotics is thus a related, yet new application of
the transferability concept. We call this new approach to resilient
robotics ``T-Resilience'' (for Transferability-based Resilience).

\paragraph{Algorithm.} 
The T-Resilience algorithm relies on three main principles
(Fig.~\ref{fig:transferability} and Algorithm \ref{algo:transf}):
\begin{itemize}
\item the self-model of the robot is not updated;
\item the approximated transferability function is learned ``on the
  fly'' thanks to a few periodic tests conducted on the robot and a
  regression algorithm;
\item three objectives are optimized simultaneously:
  \begin{displaymath}
    \textrm{maximize } \left\{\begin{array}{l}
    \mathcal{F}_{self}(\mathbf{c})\\
    \hat{\mathcal{T}}({\bf b}_{self}(\mathbf{c}))\\
    diversity(\mathbf{c})
    \end{array}
    \right.
  \end{displaymath}
\end{itemize}
where $\mathcal{F}_{self}(\mathbf{c})$ denotes the performance of the
candidate solution $\mathbf{c}$ that is predicted by the self-model
(e.g. the forward displacement in the simulation); ${\bf
  b}_{self}(\mathbf{c})$ denotes the behavior descriptor of
$\mathbf{c}$, extracted by recording the behavior of $\mathbf{c}$ in
the self-model; $\hat{\mathcal{T}}({\bf b}_{self}(\mathbf{c}))$
denotes the approximated transferability function between the
self-model and the damaged robot, which is separately learned using a
regression algorithm; and $diversity(\mathbf{c})$ is a
application-dependent helper-objective that helps the optimization
algorithm to mitigate premature convergence (\cite{Toffolo2003,
  Mouret2012div}).

Evaluating these three objectives for a particular controller does not
require any real test: the behavior of each controller and the
corresponding performance are predicted by the self-model; the
approximated transferability value is computed thanks to the
regression model of the transferability function. \emph{The update of
  the approximated transferability function is therefore the only step
  of the algorithm that requires a real test on the robot.} Since this
update is only performed every $N$ iterations of the optimization
algorithm, only a handful of tests on the real robot have to be done.

At a given iteration, the T-Resilience algorithm does not need to
predict the transferability of the whole search space, it only needs
these values for the candidate solutions of the current
population. Since the population, on average, moves towards better
solutions, the algorithm has to periodically update the approximation
of the transferability function. To make this update simple and
unbiased, we chose to select the solution to be tested on the robot by
picking a random individual from the population. We experimented with
other selection schemes in preliminary experiments, but we did not
observe any significant improvement.

%Let $I_{max}$ be
%the maximal number of iterations of the MOEA, the maximal number of
%real tests $T$ conducted on the robot during the overall process is
%the integer part of the ratio ${I_{max}} / {N}$.

Three choices depend on the application:
\begin{itemize}
  \item the performance measure $\mathcal{F}_{self}$ (i.e. the reward function);
  \item the diversity measure;
  \item the regression technique used to learn the transferability
    function and, in particular, the inputs and outputs of this
    function.
\end{itemize}

We will discuss and describe each of these choices for our resilient
hexapod robot in section \ref{sec:experiments}.

\paragraph{Optimization algorithm.} 
Recent research in stochastic optimization proposed numerous
algorithms to simultaneously optimize several objectives
\citep{deb2001}; most of them are based on the concept of Pareto
dominance, defined as follows:

\begin{definition}[Pareto dominance]A solution $p^*$ is said to dominate another solution $p$, if both
conditions 1 and 2 are true:
\begin{enumerate}
\item the solution $p^*$ is not worse than $p$ with respect to all objectives;
\item the solution $p^*$ is strictly better than $p$ with respect to at least one objective.
\end{enumerate}
\end{definition}

The non-dominated set of the entire feasible search space is the
globally Pareto-optimal set (Pareto front). It represents the set of
optimal \emph{trade-offs}, that is solutions that cannot be improved
with respect to one objective without decreasing their score with
respect to another one.

\begin{figure}
\centering
\includegraphics[width=\columnwidth]{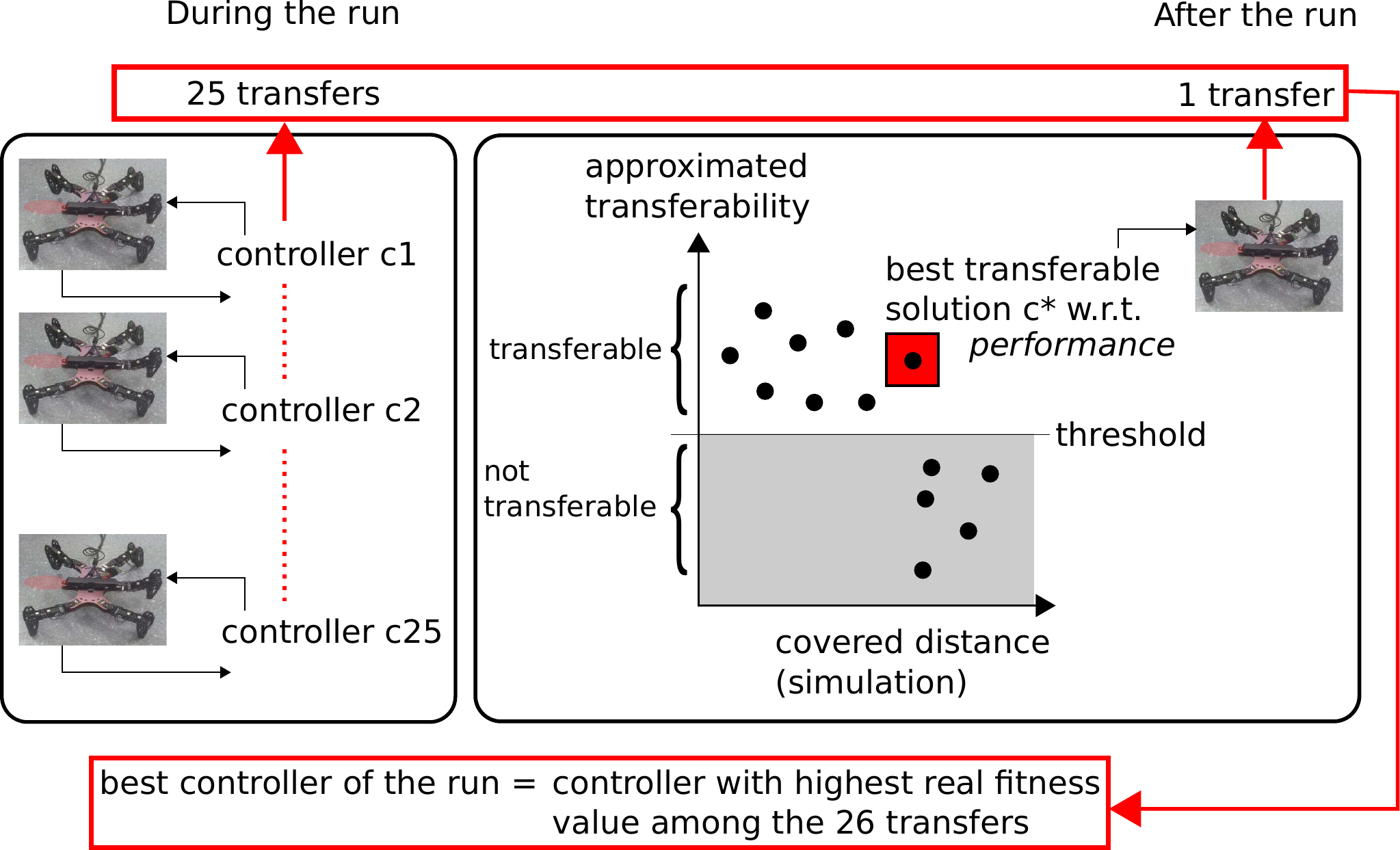}
\caption{Choice of the final solution at the end of the T-Resilience algorithm.}
\label{fig:best_choice}
\end{figure}

\begin{figure*}[h]
\centering
\subfloat[Hexapod robot.]{\includegraphics[width=0.49\columnwidth, height=4.16cm]{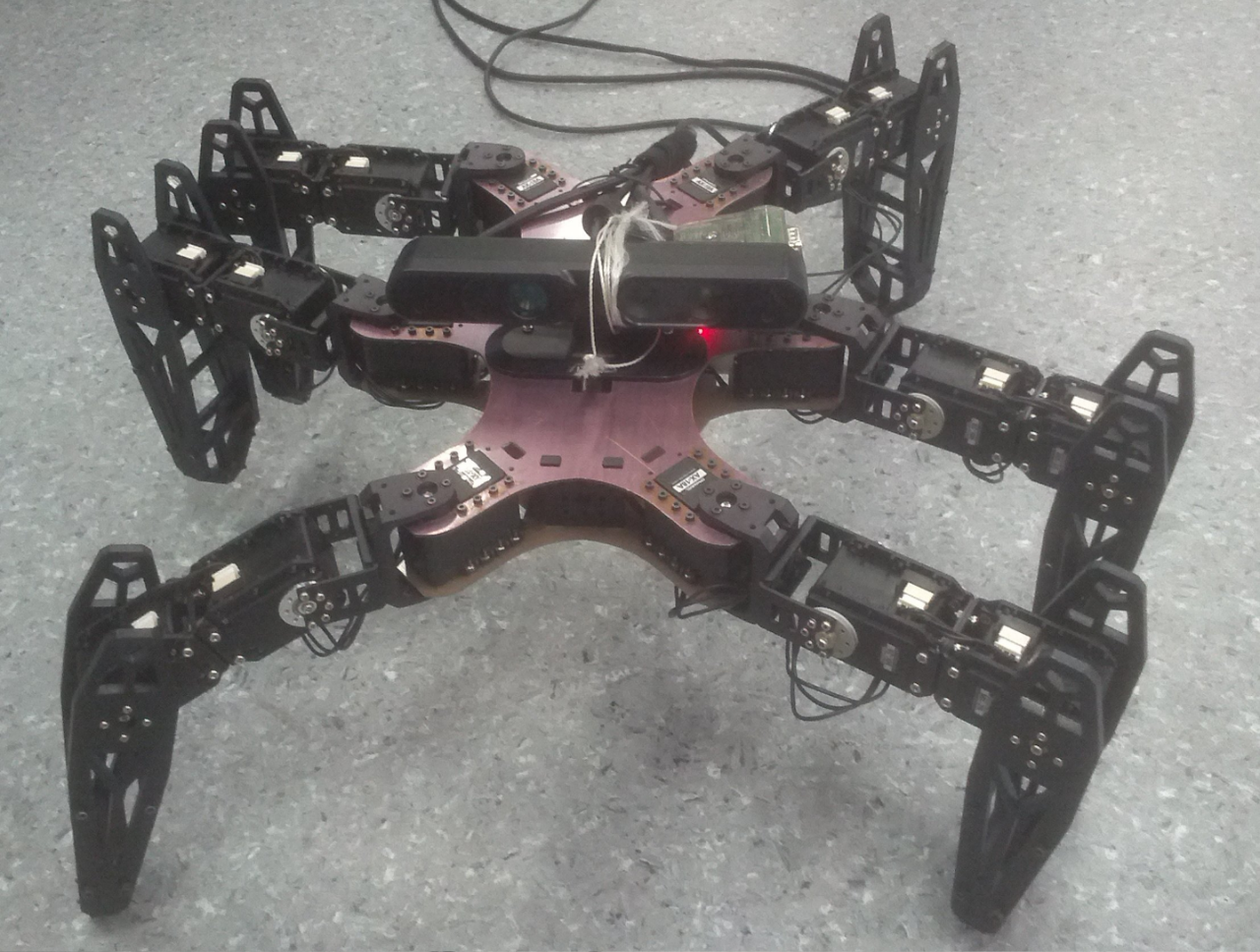}}\hfil
\subfloat[Self-model.]{\includegraphics[width=0.50\columnwidth, height=4.16cm]{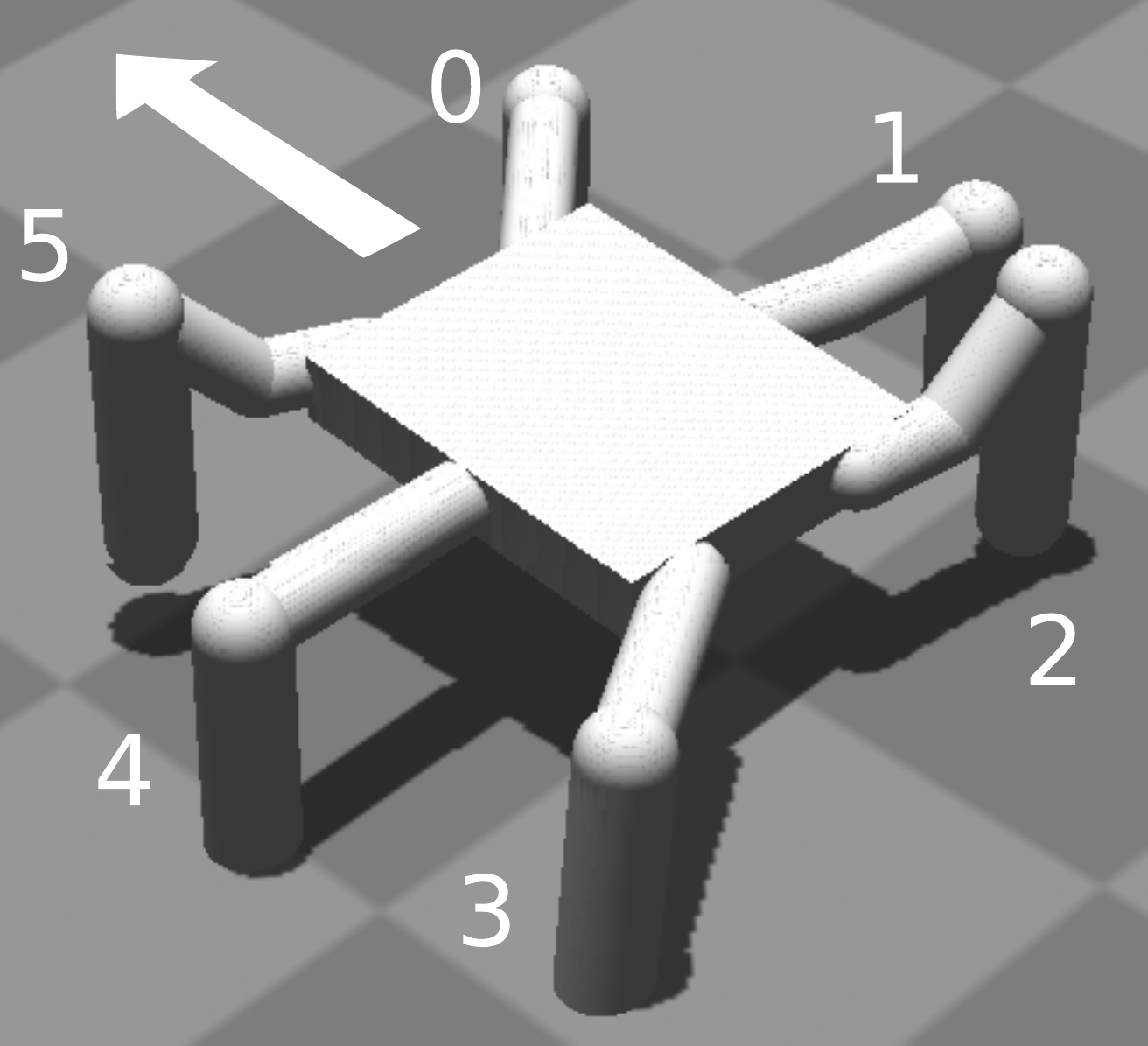}}
\subfloat[Kinematic scheme.]{\includegraphics[width=0.50\columnwidth, height=4cm]{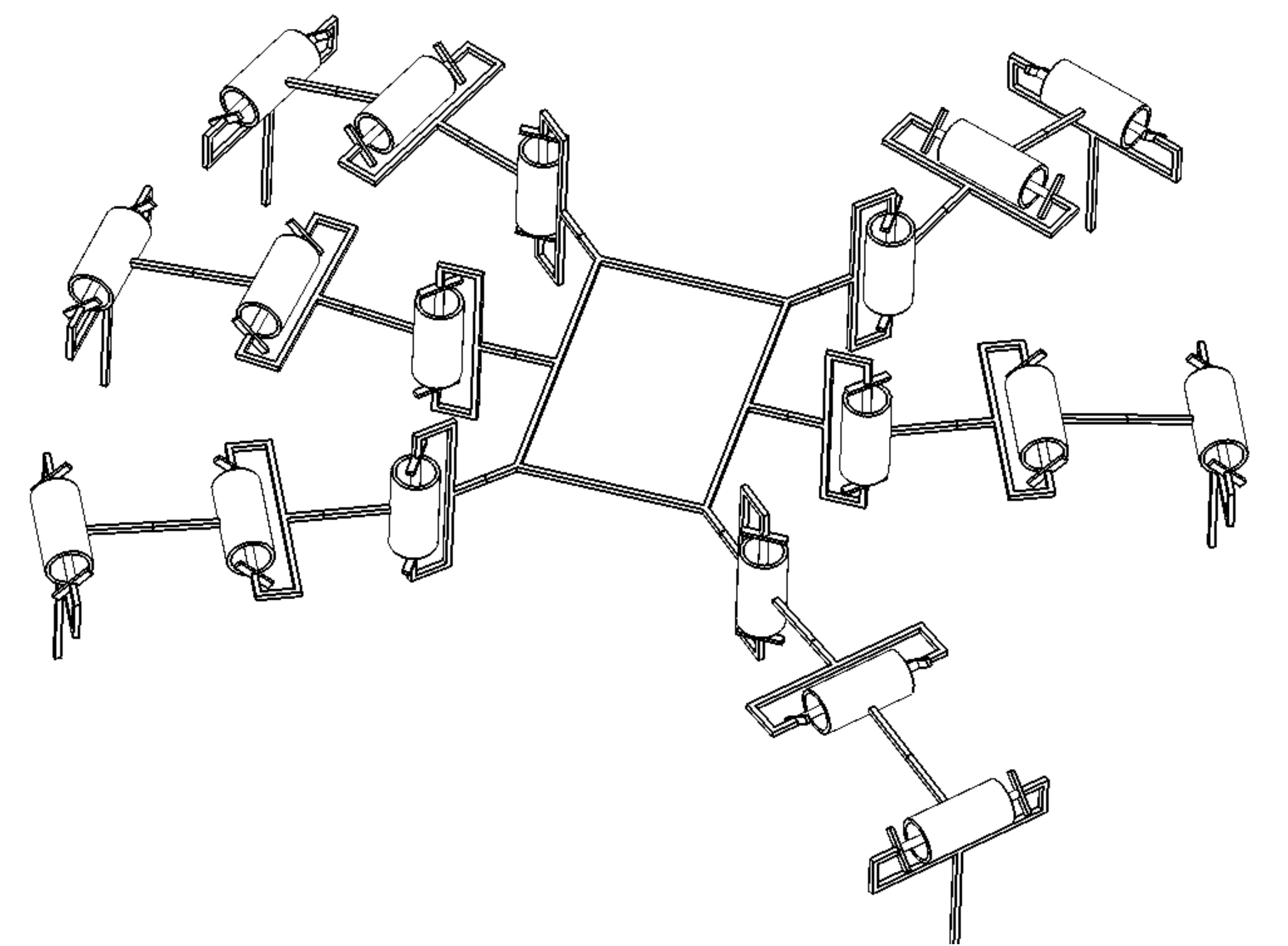}}
\subfloat[Control function.]{\includegraphics[width=0.45\columnwidth, height=4.5cm]{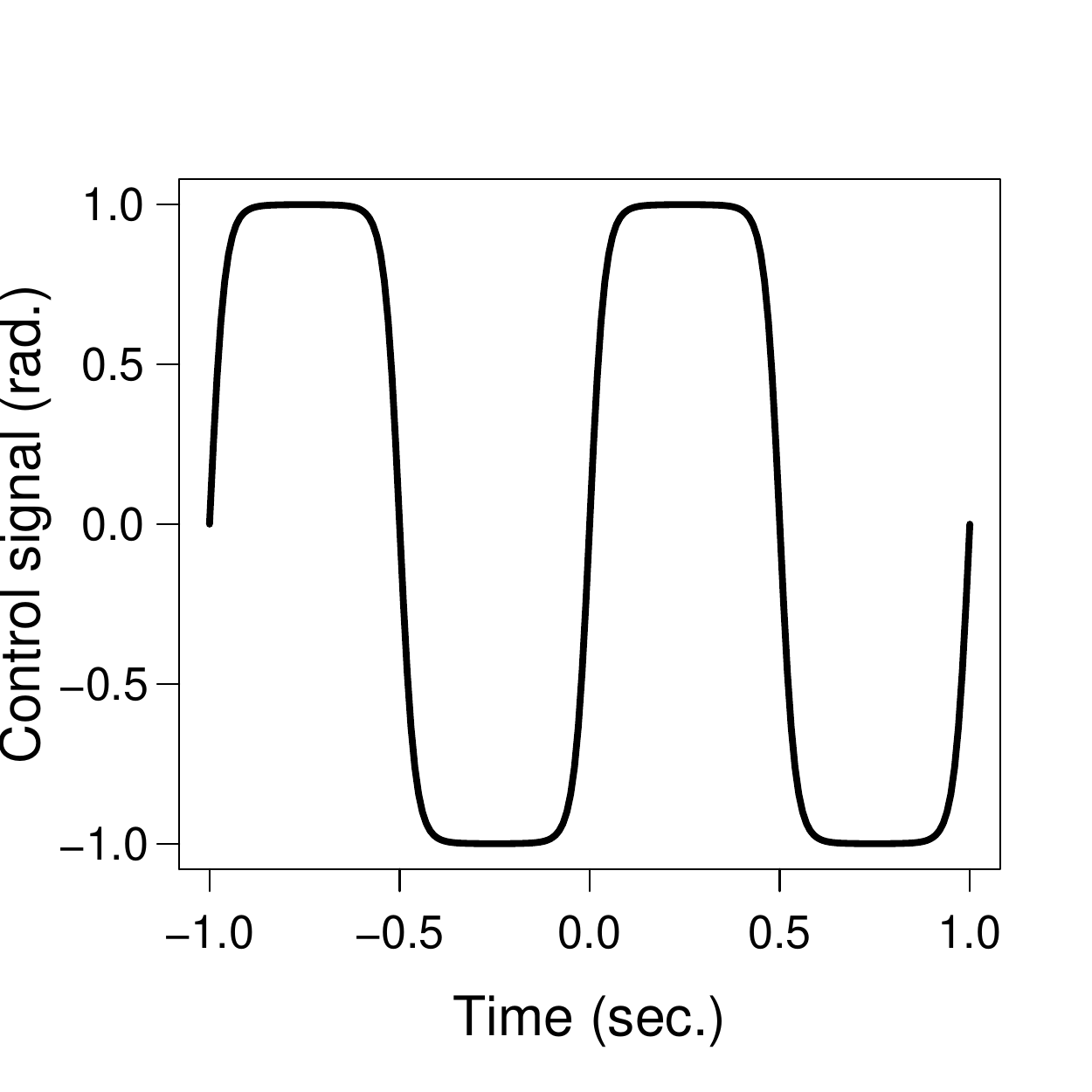}}
\caption{(a) The 18-DOF hexapod robot is equipped with a RGB-D camera
  (RGB camera with a depth sensor). (b) Snapshot of the simulation
  used as a self-model by the robot which occurs in an ODE-based
  physics simulator. The robot lies on a horizontal plane and
  contacts are simulated as well. (c) Kinematic scheme of the
  robot. (d) Control function $\gamma(t,\ \alpha,\ \phi)$ with $\alpha
  = 1$ and $\phi = 0$.}
\label{fig:robot}\label{fig:control}
\end{figure*}

Pareto-based multi-objective optimization algorithms aim at finding
the best approximation of the Pareto front, both in terms of distance
to the Pareto front and of uniformity of its sampling. This Pareto
front is found using only one execution of the algorithm and the
choice of the final solution is left to another algorithm (or to the
researcher). Whereas classic approaches to multi-objective
optimization aggregate objectives (e.g. with a weighted sum) then use
a single-objective optimization algorithm, multi-objective
optimization algorithms do not require tuning the relative importance
of each objective.

Current stochastic algorithms for multi-objective optimization are
mostly based on EAs, leading to Multi-Objective Evolutionary
Algorithms (MOEA). Like most EAs, they are intrinsically
parallel~\citep{Cantu2000}, making them especially efficient on modern
multi-core computers, GPUs and clusters~\citep{Mouret2010}. In the
T-Resilience algorithm, we rely on
NSGA-II~\citep{deb2002fast,deb2001}, one of the most widely used
multi-objective optimization algorithm (appendix
\ref{appendix:nsga2}); however, any Pareto-based multi-objective
algorithm can replace this specific EA in the T-Resilience algorithm.

At the end of the optimization algorithm, the MOEA discards diversity
values and returns a set of non-dominated solutions based on
performance and transferability. We then need to choose the final
controller. Let us define the ``transferable non-dominated set'' as
the set of non-dominated solutions whose transferability values are
greater than a user-defined threshold. To determine the best solution
of a run, the solution of the transferable non-dominated set with the
highest performance in simulation is transferred onto the robot and
its performance in reality is assessed. The final solution of the run
is the controller that leads to the highest performance on the robot
among all the transferred controllers (Fig.~\ref{fig:best_choice}).

\section{Experimental validation}
\label{sec:experiments}

\subsection{Robot and parametrized controller}\label{section:robot_and_control}

The robot is a hexapod with 18 Degrees of Freedom (DOF), 3 for each
leg (Fig.~\ref{fig:robot}(a,c)). Each DOF is actuated by
position-controlled servos (6 AX-12 and 12 MX-28 Dynamixel actuators,
designed by Robotis). The first servo controls the horizontal
orientation of the leg and the two others control its elevation. The
kinematic scheme of the robot is pictured on Figure~\ref{fig:control}
c.

A RGB-D camera (Asus Xtion) is screwed on top of the robot. It is used
to estimate the forward displacement of the robot thanks to a RGB-D
SLAM algorithm~\citep{endres12icra}\footnote{We downloaded our
  implementation from: \url{http://www.ros.org/wiki/rgbdslam}} from
the ROS framework~\citep{Ros2009}\footnote{\url{http://www.ros.org}}.

The movement of each DOF is governed by a periodic function that
computes its angular position as a function $\gamma$ of time $t$,
amplitude $\alpha$ and phase $\phi$ (Fig.~\ref{fig:control}, d):
\begin{equation}
\gamma(t,\ \alpha,\ \phi) = \alpha\cdot\tanh\left(4\cdot\sin\left(2\cdot\pi\cdot(t + \phi)\right)\right)\label{eq:tanh}
\end{equation}
where $\alpha$ and $\phi$ are the parameters that define the amplitude
of the movement and the phase shift of $\gamma$,
respectively. Frequency is fixed.

Angular positions are sent to the servos every 30 ms. The main feature
of this particular function is that, thanks to the $\tanh$ function,
the control signal is constant during a large part of each cycle, thus
allowing the robot to stabilize itself. In order to keep the ``tibia''
of each leg vertical, the same control signal is used for the two last
servos. Consequently, positions sent to the $i^{th}$ servos are:
\begin{itemize}
\item $\gamma(t,\ \alpha^i_1,\ \phi^i_1)$ for DOF 1;
\item $\gamma(t,\ \alpha^i_2,\ \phi^i_2)$ for DOFs 2 and 3.
\end{itemize}

There are $4$ parameters for each leg ($\alpha^i_1$, $\alpha^i_2$,
$\phi^i_1$, $\phi^i_2$), therefore each controller is fully described
by $24$ parameters. By varying these $24$ parameters, numerous gaits
are possible, from purely quadruped gaits to classic tripod gaits.

This controller is designed to be as simple as possible so that we can
show the performance of the T-Resilience algorithm in a
straightforward setup. Nevertheless, the T-Resilience algorithm does
not put any constraint on the type of controllers and many other
controllers are conceivable (e.g. bio-inspired central pattern
generators like \cite{sproewitz2008learning} or evolved neural
networks like in \citep{Yosinski2011,Clune2011}).

\subsection{Reference controller}
A classic tripod gait~\citep{Wilson1966,saranli2001rhex,
  schmitz2001biologically, ding2010locomotion, steingrube2010self} is
used as a reference point. This reference gait considers two tripods:
legs 0, 2, 4 and legs 1, 3, 5 (see Figure~\ref{fig:robot} for
numbering). It is designed to always keep the robot balanced on at
least one of these tripods. The walking gait is achieved by lifting
one tripod, while the other pushes the robot forward (by shifting
itself backward). The lifted tripod is then placed forward in order to
repeat the cycle by inverting the tripods. This gait is static, fast
and similar to insect gaits~\citep{Wilson1966,Delcomyn1971}. The
parameters of this reference controller are available in appendix
\ref{appendix:ref}.

\subsection{Implementation choices for T-Resilience}
\label{sec:impl_choices}
\paragraph{Performance function.} 
The mission of our hexapod robot is to go forward as fast as possible,
regardless of its current state and of any sustained damages. The
performance function to be optimized is the forward displacement of
the robot predicted by its self-model. Such a high-level function does
not constrain the features of the optimized behaviors, so that the
search remains as open as possible, possibly leading to original
gaits~\citep{nelson2009fitness}:

\begin{equation}
\mathcal{F}_{self}(\mathbf{c}) = p^{t=E,SELF}_x(\mathbf{c}) - p^{t=0, SELF}_x(\mathbf{c})
\end{equation}
where $p^{t=0, SELF}_x(\mathbf{c})$ denotes the x-position of the
robot's center at the beginning of the simulation when the parameters
$\mathbf{c}$ are used and $p^{t=E, SELF}_x(\mathbf{c})$ its x-position
the end of the simulation.

Because each trial lasts only a few seconds, this performance function
does not strongly penalize gaits that do not lead to straight
trajectories. Using longer experiments would penalize these
trajectories more, but it would increase the total experimental time
too much to perform comparisons between approaches. Other performance
functions are possible and will be tested in future work.

\paragraph{Diversity function.}
The diversity score of each individual is the average Euclidean
distance to all the other candidate solutions of the current
population. Such a parameter-based diversity objective enhances the
exploration of the control space by the population~\citep{Toffolo2003,
  Mouret2012div} and allows the algorithm to avoid many local
optima. This diversity objective is straightforward to implement and
does not depend on the task.

\begin{equation}
diversity(\mathbf{c}) =  \frac{1}{N} \sum_{y \in P_n} \sqrt{\sum_{j = 1}^{24} (\mathbf{c}_j - \mathbf{y}_j)^2}
\end{equation}
where $P_n$ is the population at generation $n$, $N$ the size of $P$
and $\mathbf{c}_j$ the $j^{th}$ parameter of the candidate solution
$\mathbf{c}$. Other diversity measures (e.g. behavioral measures, like
in \citep{Mouret2012div}) led to similar results in preliminary
experiments.

\paragraph{Regression model.}
When a controller $\mathbf{c}$ is tested on the real robot, the
corresponding exact transferability score $\mathcal{T}$ is computed as
the absolute difference between the forward performance predicted by
the self-model and the performance estimated on the robot based on the
SLAM algorithm.

% and the RGB-D camera:

\begin{equation}
\mathcal{T}(\mathbf{c}) = \Big|p_{t=E}^{SELF}(\mathbf{c}) - p_{t=0}^{REAL}(\mathbf{c}) \Big|
\end{equation}
The transferability function is approximated by training a SVM model
$\hat{\mathcal{T}}$ using the $\nu$-Support Vector Regression
algorithm with linear kernels implemented in the library
\emph{libsvm}\footnote{\url{http://www.csie.ntu.edu.tw/~cjlin/libsvm}}~\citep{chang2011libsvm}
(learning parameters are set to default values).

\begin{equation}
\hat{\mathcal{T}}({\bf b}_{self}(\mathbf{c})) = \textrm{SVM}(b_{t=0}^{(1)}, \cdots, b_{t=E}^{(1)}, \cdots, b_{t=0}^{(6)}, \cdots, b_{t=E}^{(6)})
\end{equation}
where $E$ is the number of time-steps of the control function (equation \ref{eq:tanh}) and:
\begin{equation}
  b_{t}^{(n)} = \left\{
  \begin{array}{ll} 
  1 & \textrm{ if leg $n$ touches the ground at that time-step}\\ 
  0 & \textrm{ otherwise}
  \end{array}
  \right.
\end{equation}

We chose to describe gaits using contacts\footnote{When choosing the
  input of a predictor, there is a large difference between using the
  control parameters and using high-level descriptors of the
  behavior~\citep{Mouret2012div}. Intuitively, most humans can predict
  that a behavior will work on a real robot by watching a simulation,
  but their task is much harder if they can only see the
  parameters. More technically, predicting features of a complex
  dynamical system usually requires simulating it. By starting with
  the output of a simulator, the predictor avoids the need to
  re-invent physical simulation and can focus on discrimination.},
because it is a classic representation of robotic and animal gaits
(e.g. \cite{Delcomyn1971}). On the real robots, we deduces the
contacts by measuring the torque applied by each servo.

We chose SVMs to approximate the transferability score because of the
high number of inputs of the model and because there are many
available implementations. Contrary to other classic regression models
(neural networks, Kriging, ...), SVMs are indeed not critically
dependent on the size of the input
space~\citep{smola1997support,smola2004tutorial}. They also provide
fast learning and fast prediction when large input spaces are used.

\paragraph{Self-model.}

The self-model of the robot is a dynamic simulation of the undamaged
six-legged robot in Open Dynamics Engine (ODE)\footnote{Open Dynamics
  Engine: \url{http://www.ode.org}} on a flat ground
(Fig.~\ref{fig:robot}b).

\paragraph{Main parameters.}
For each experiment, a population of 100 controllers is optimized
for 1000 generations. Every 40 generations, a controller is
randomly selected in the population and transferred on the robot, that
is we use 25 real tests on the robot in a run. Each test takes place
as follows:
\begin{itemize}
\item the selected controller is transferred and evaluated for 3
  seconds on the robot while the RGB-D camera records both color and
  depth images at 10 Hz;
\item a SLAM algorithm estimates the forward displacement of the robot
  based on the data of the camera;
\item the estimate of the forward displacement is provided to the main
  algorithm.
\end{itemize}

At each generation, each parameter of each selected candidate solution
has a $10\%$ chance of being incremented or decremented, with both
options equally likely; five values are available for each $\varphi$
($0, 0.25, 0.5, 0.75, 1$) and for each $\alpha$ ($0, 0.25, 0.5, 0.75,
1$).

To select the final solution, we fixed the transferability threshold
at 0.1 meter.

\subsection{Test cases and compared algorithms}

\begin{figure}
\centering
\includegraphics[width=\columnwidth]{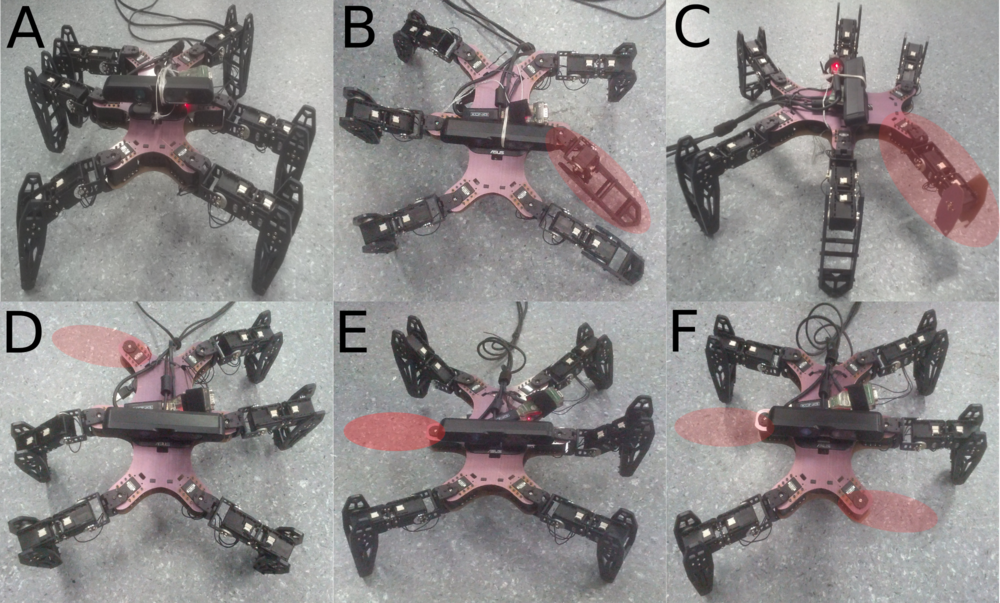}
\caption{Test cases considered in our experiments. (A) The hexapod robot is not damaged. (B) The left middle leg is no longer powered. (C) The terminal part of the front right leg is shortened by half. (D) The right hind leg is lost. (E) The middle right leg is lost. (F) Both the middle right leg and the front left leg are lost.}
\label{fig:failures_robot}
\end{figure}

To assess the ability of T-Resilience to cope with many different
failures, we consider the six following test
cases~(Fig.~\ref{fig:failures_robot}):
\begin{itemize}
\item A. the hexapod robot is not damaged;
\item B. the left middle leg is no longer powered;
\item C. the terminal part of the front right leg is shortened by half;
\item D. the right hind leg is lost;
\item E. the middle right leg is lost;
\item F. both the middle right leg and the front left leg are lost.
\end{itemize}

We compare the The T-Resilience algorithm to three representative
algorithms from the literature (see appendix \ref{appendix:details}
for the exact implementations of each algorithm):
\begin{itemize}
\item a stochastic local search~\citep{hoos2005stochastic}, because of
  its simplicity;
\item a policy gradient method inspired from~\cite{kohl2004policy},
  because this algorithm has been successfully applied to learn
  quadruped locomotion;
\item a self-modeling process inspired from~\cite{bongard2006resilient}.
\end{itemize}

To make the comparisons as fair as possible, we designed our
experiments to compare algorithms after the same amount of running
time or after the same number of real tests (see appendix
\ref{appendix:median_durations} for their median durations and their
median numbers of real tests). In all the test cases, the T-Resilience
algorithm required about 19 minutes and 25 tests on the robot (1000
generations of $100$ individuals). Consequently, two key values are
recorded for each algorithm (see Appendix \ref{appendix:details} for
exact procedures):
\begin{itemize}
\item the performance of the best controller obtained after about 25
  real tests\footnote{Depending on the algorithm, it is sometimes
    impossible to perform exactly 25 tests (for instance, if two tests
    are performed for each iteration).};
\item the performance of the best controller obtained after about 19 minutes.
\end{itemize}

The experiments for the four first cases (A, B, C and D) showed that
only the stochastic local search is competitive with the
T-Resilience. We therefore chose to only compare T-Resilience with the
local search algorithm for the two last failures (E and F).

Preliminary experiments with each algorithm showed that initializing
them with the parameters of the reference controller did not improve
their performance. We interpret these preliminary experiments as
indicating that the robot needs to use a qualitatively different gait,
which requires substantial changes in the parameters. This observation
is consistent with the gaits we tried to design for the damaged
robot. As a consequence, we chose to initialize each of the compared
algorithms with random parameters instead of initializing them with
the parameters of the reference controller. By thus starting with
random parameters, we do not rely on any a priori about the gaits for
the damaged robot: we start with the assumption that anything could
have happened.

We replicate each experiment $5$ times to obtain statistics. Overall,
this comparison requires the evaluation of about $4000$ different
controllers on the real robot.

We use $4$ Intel(R) Xeon(R) CPU E31230 @ 3.20GHz, each of them
including $4$ cores. Each algorithm is programmed in the Sferes$_{v2}$
framework~\citep{Mouret2010} and the source-code is available as
extension 10. The MOEA used in Bongard's algorithm and in the
T-Resilience algorithm is distributed on $16$ cores using MPI.

Final performance values are recorded with a CODA cx1 motion
capture system (Charnwood Dynamics Ltd, UK) so that reported results
do not depend on inaccuracies of the internal measurements. However,
all the tested algorithms have only access to the internal
measurements.

\section{Results}

\subsection{Reference controller}
\begin{figure*}
\centering
\subfloat[Undamaged hexapod robot (case A).]{\includegraphics[width=\columnwidth]{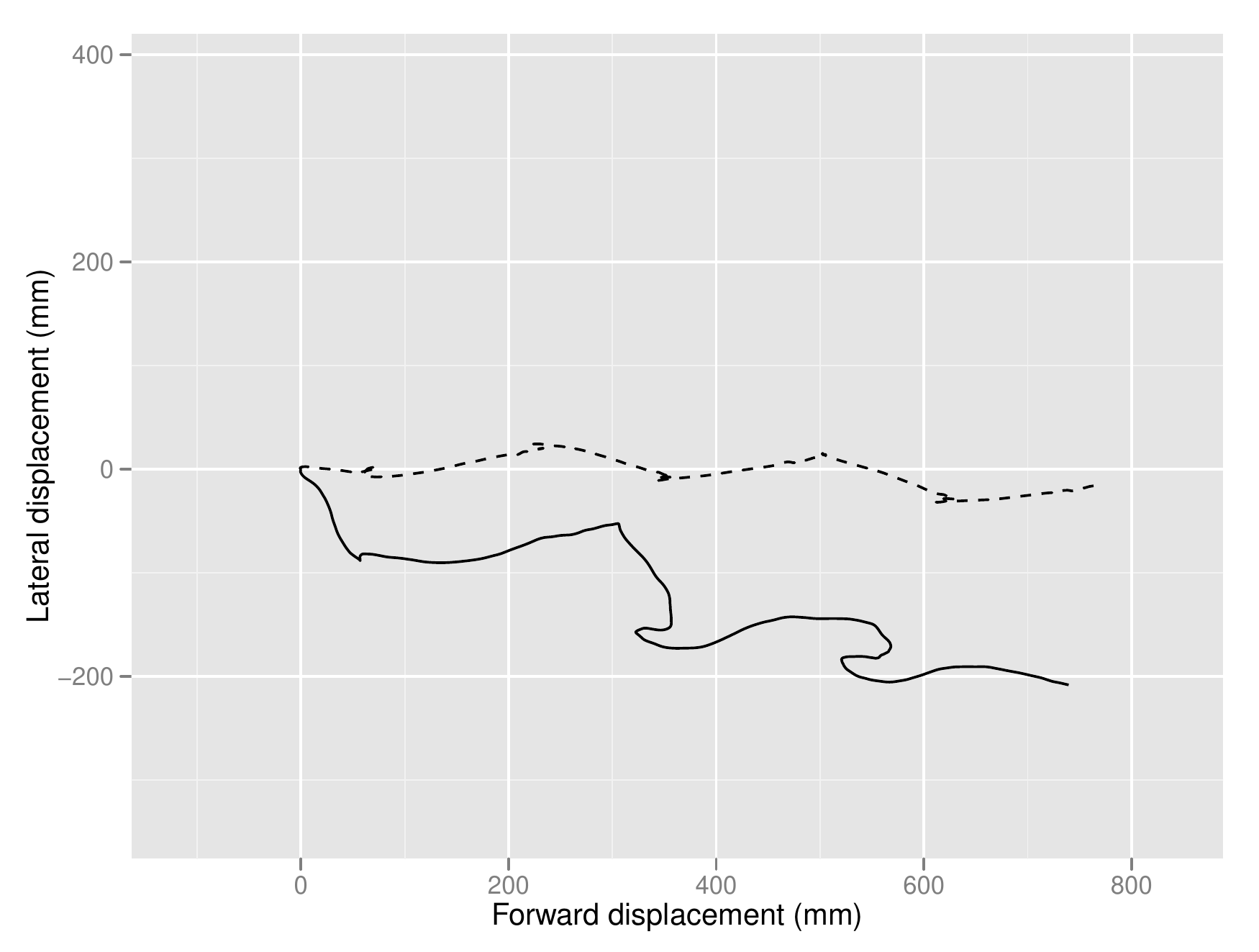}\label{fig:traj_both_undamaged}}
\subfloat[Middle left leg not powered (case B).]{\includegraphics[width=\columnwidth]{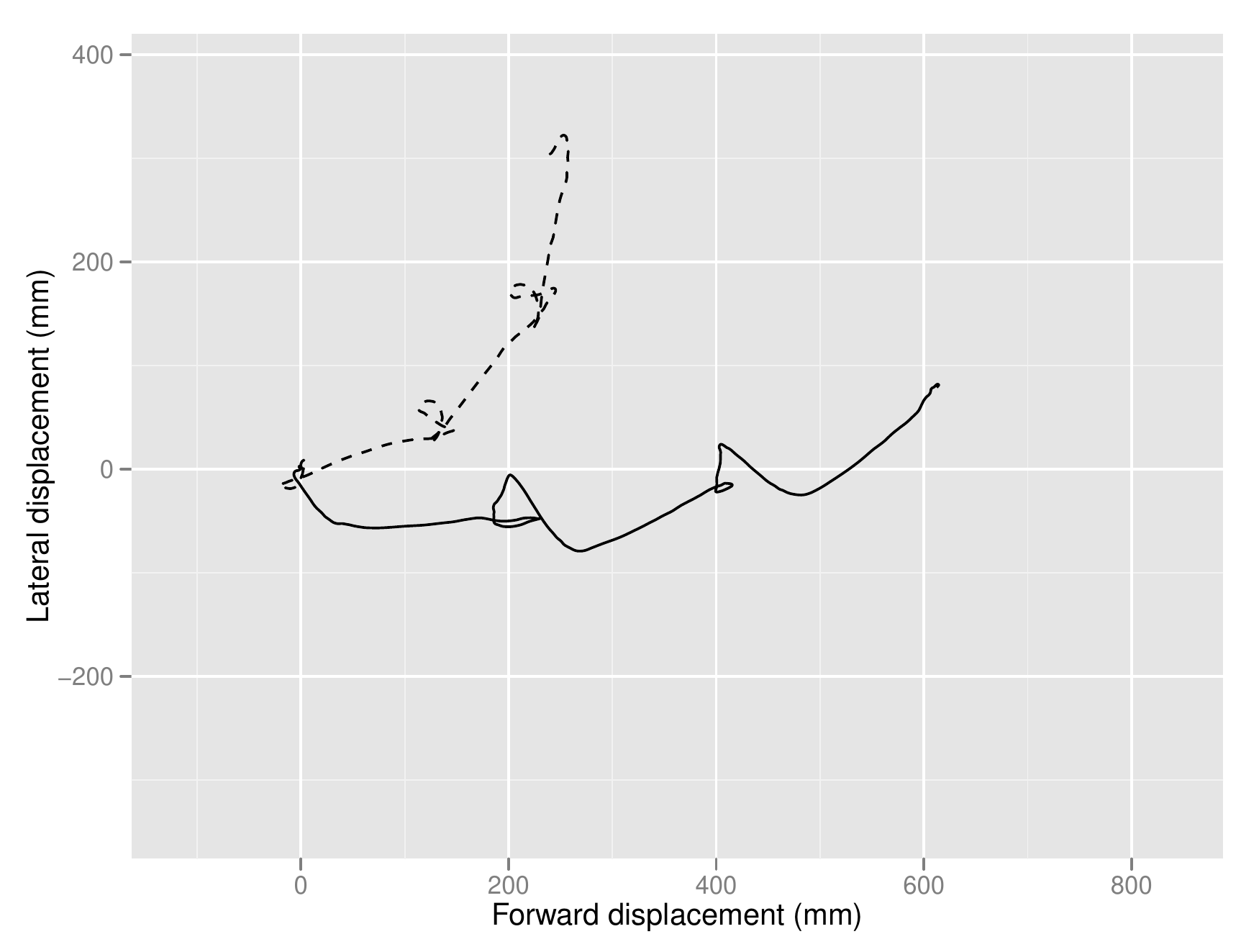}\label{fig:traj_both_unpowered}}\\
\subfloat[Front right leg shortened by half (case C).]{\includegraphics[width=\columnwidth]{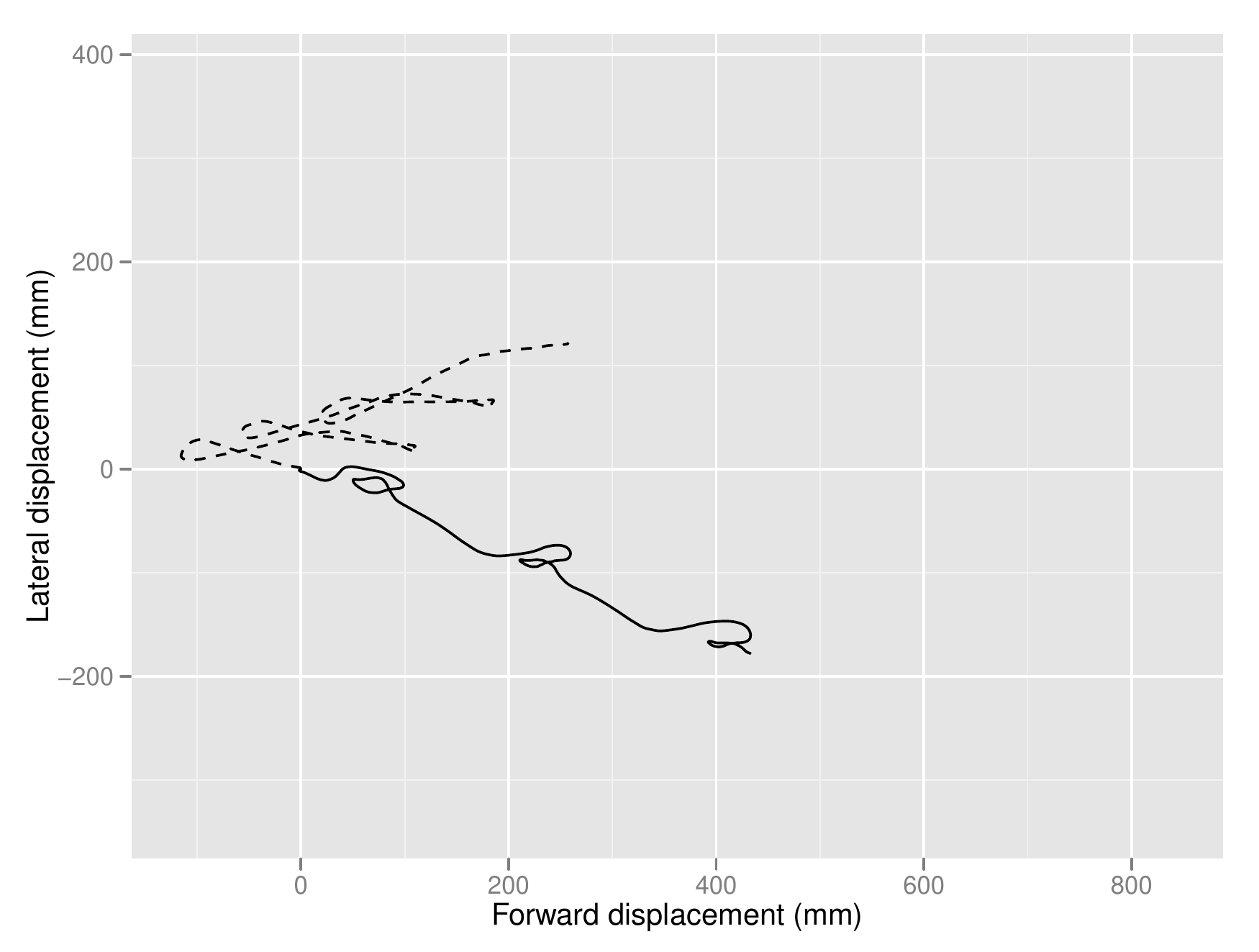}\label{fig:traj_both_shortened}}
\subfloat[Hind right leg lost (case D).]{\includegraphics[width=\columnwidth]{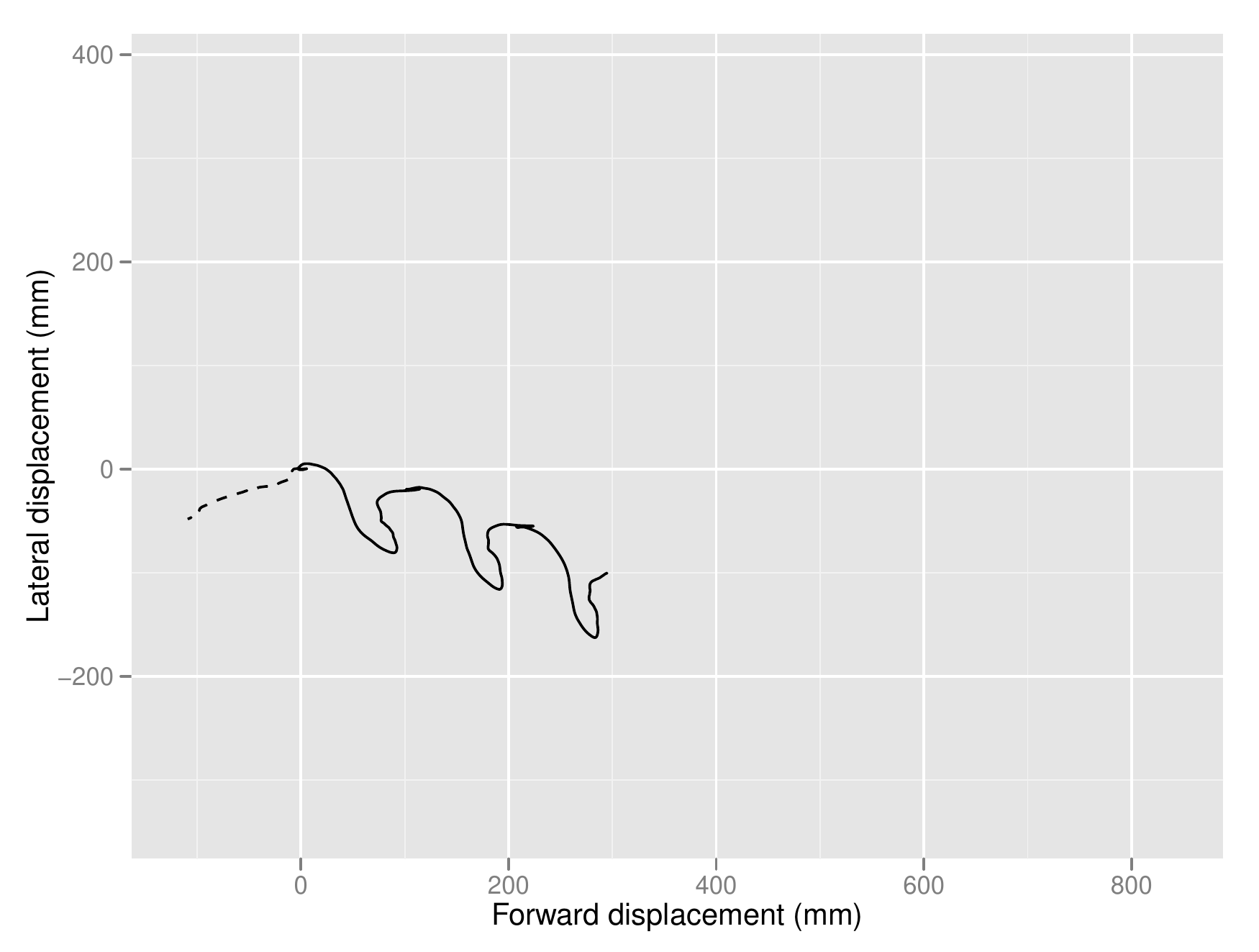}\label{fig:traj_both_lostleg}}\\
\subfloat[Middle right leg lost (case E).]{\includegraphics[width=\columnwidth]{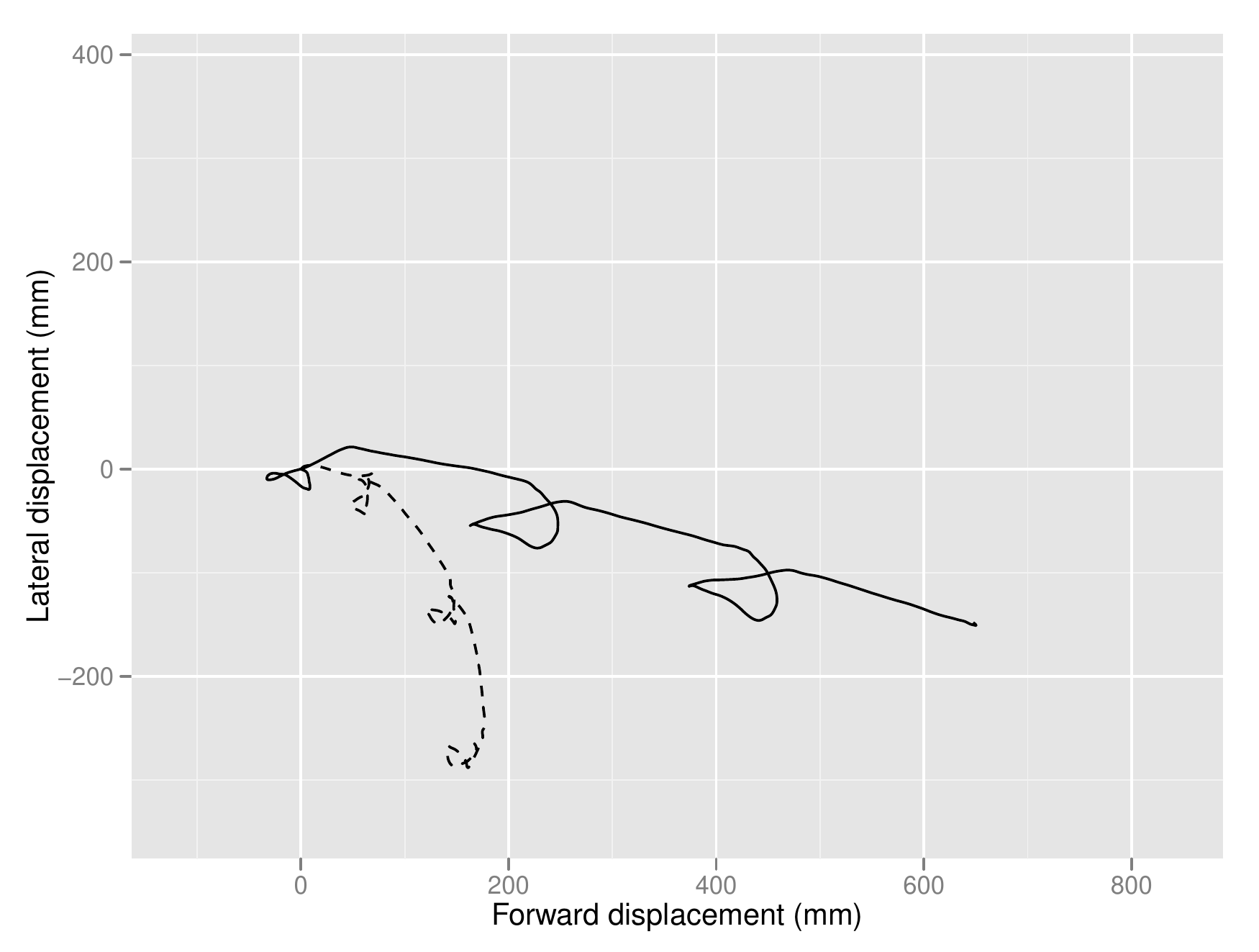}\label{fig:traj_both_middlelostleg}}
\subfloat[Middle right leg and front left leg lost (case F).]{\includegraphics[width=\columnwidth]{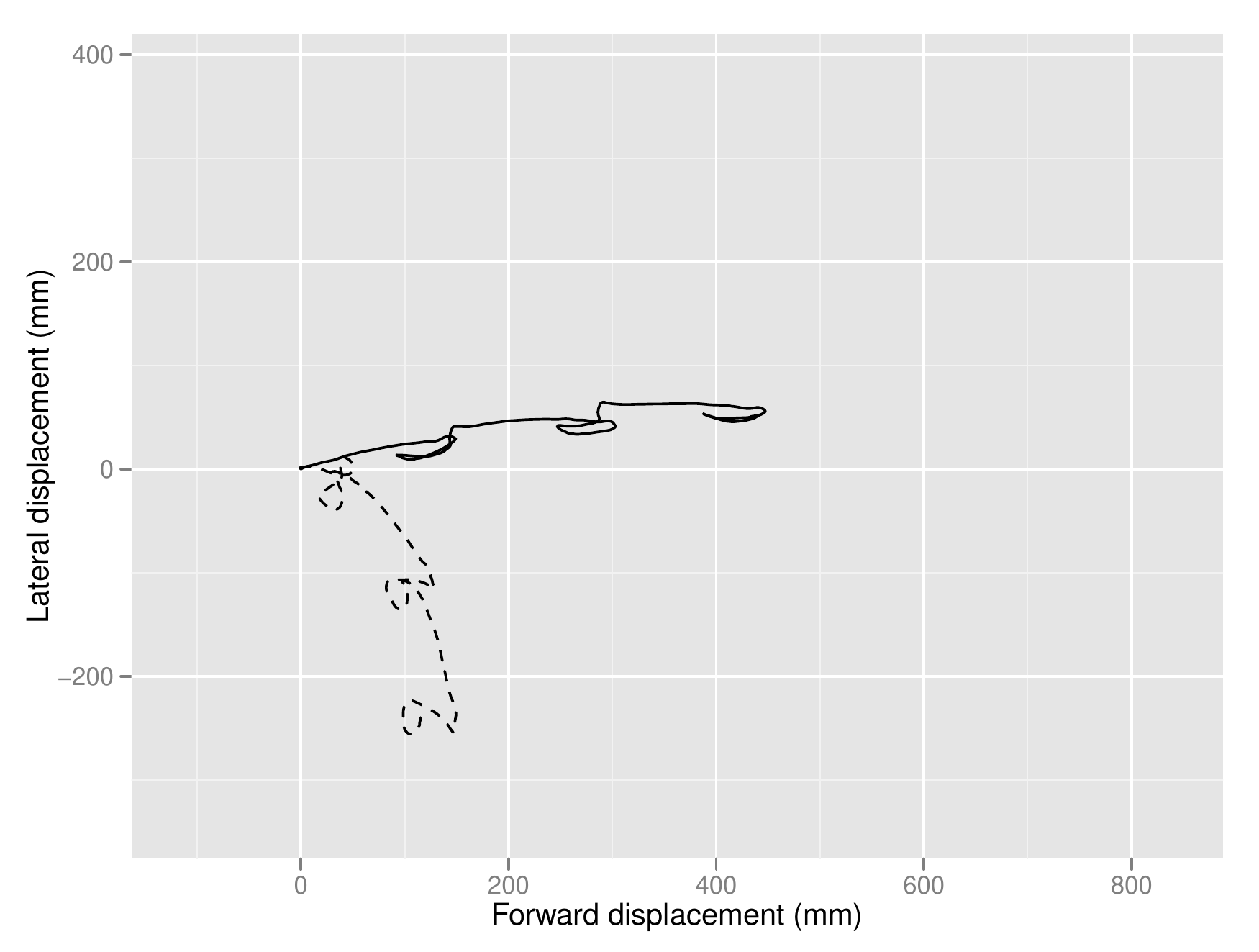}\label{fig:traj_both_twolostlegs}}
\caption{Typical trajectories (median performance) observed in every
  test case. Dashed line: reference gait. Solid line: controller with
  median performance value found by the T-Resilience algorithm. The
  poor performance of the reference controllers after any of the
  damages shows that adaptation is required in these situations. The
  trajectories obtained with the T-Resilience algorithm are not
  perfectly straight because our objective function does not
  explicitly reward straightness (see sections \ref{sec:impl_choices}
  and \ref{sec:results}). }
\label{fig:trajboth}
\end{figure*}

Table~\ref{table:hexapod} reports the performances of the reference
controller for each tested failure, measured with both the CODA
scanner and the on-board SLAM algorithm. At best, the damaged robot
covered $35\%$ of the distance covered by the undamaged robot ($0.78$
m with the undamaged robot, at best $0.26$ m after a failure). In
cases B, C and E, the robot also performs about a quarter turn
(Figure~\ref{fig:trajboth} (a), (b) and (e)); in case D, it falls
over; in case F, it alternates forward locomotion and backward
locomotion (figure~\ref{fig:trajboth} (f)). Videos of these behaviors
are available in appendix \ref{appendix:videos}.

This performance loss of the reference controller clearly shows
that an adaptation algorithm is required to allow the robot to pursue
its mission. Although not perfect, the distances reported by the
on-board RGB-D SLAM are sufficiently accurate to easily detect when
the adaptation algorithm must be launched.

\begin{table}
\centering
\begin{tabular}{|c|cccccc|}
\hline
Test cases & A & B & C & D & E & F\\
\hline
Perf. (CODA) & 0.78 & 0.26 & 0.25 & 0.00 & 0.15 & 0.10\\
Perf. (SLAM) & 0.75 & 0.17 & 0.26 & 0.00 & 0.04 & 0.16\\
\hline
\end{tabular}
\caption{Performances in meters obtained on the robot with the
  reference gait in all the considered test cases. Each test
  lasts 3 seconds. The CODA line corresponds to the distance covered
  by the robot according to the external motion capture system. The
  SLAM line corresponds to the performance of the same behaviors but
  reported by the SLAM algorithm. When internal measures are used
  (SLAM line), the robot can easily detects that a damage occurred
  because the difference in performance is very significant (column A
  versus the other columns).}
\label{table:hexapod}
\end{table}

%\begin{figure*}
%\centering
%\subfloat[Undamaged hexapod robot.]{\includegraphics[width=\columnwidth]{img/traj_hexapod_undamaged.pdf}\label{fig:traj_hexapod_undamaged}}
%\subfloat[Middle left leg not powered.]{\includegraphics[width=\columnwidth]{img/traj_hexapod_unpowered.pdf}\label{fig:traj_hexapod_unpowered}}\\
%\subfloat[Front right leg shortened by half.]{\includegraphics[width=\columnwidth]{img/traj_hexapod_shortened.pdf}\label{fig:traj_hexapod_shortened}}
%\subfloat[Hind right leg lost.]{\includegraphics[width=\columnwidth]{img/traj_hexapod_lostleg.pdf}\label{fig:traj_hexapod_lostleg}}\\
%\subfloat[Middle right leg lost.]{\includegraphics[width=\columnwidth]{img/traj_hexapod_middlelostleg.pdf}\label{fig:traj_hexapod_middlelostleg}}
%\subfloat[Middle right leg and front left leg lost.]{\includegraphics[width=\columnwidth]{img/traj_hexapod_twolostlegs.pdf}\label{fig:traj_hexapod_twolostlegs}}
%\caption{Trajectories obtained with the classic hexapod gait in all the test cases.}
%\label{fig:trajhexapod}
%\end{figure*}

\begin{figure*}
\centering
\subfloat[Undamaged hexapod robot (case A).]{\includegraphics[width=\columnwidth]{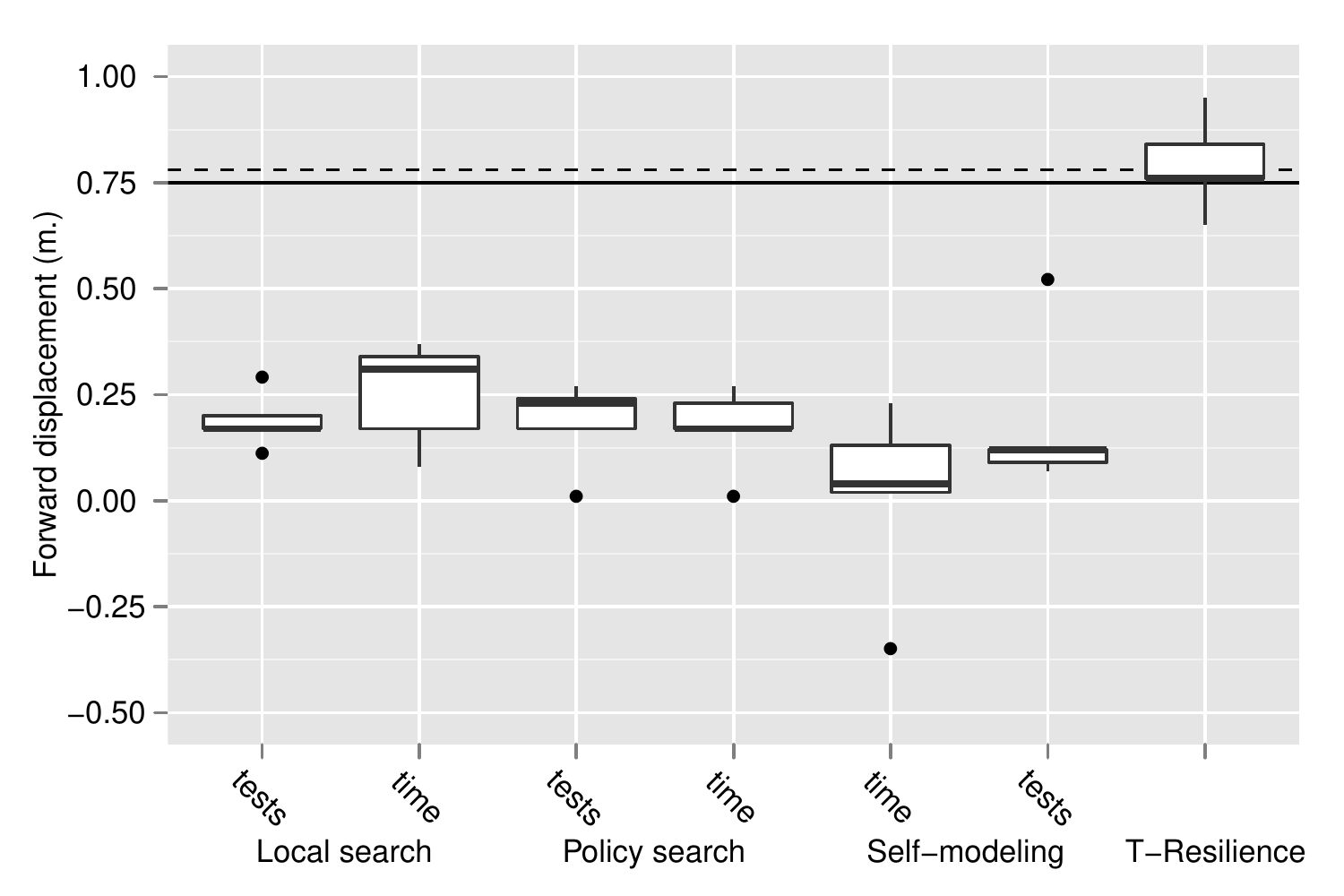}\label{fig:results_hexapod}}
\subfloat[Middle left leg not powered (case B).]{\includegraphics[width=\columnwidth]{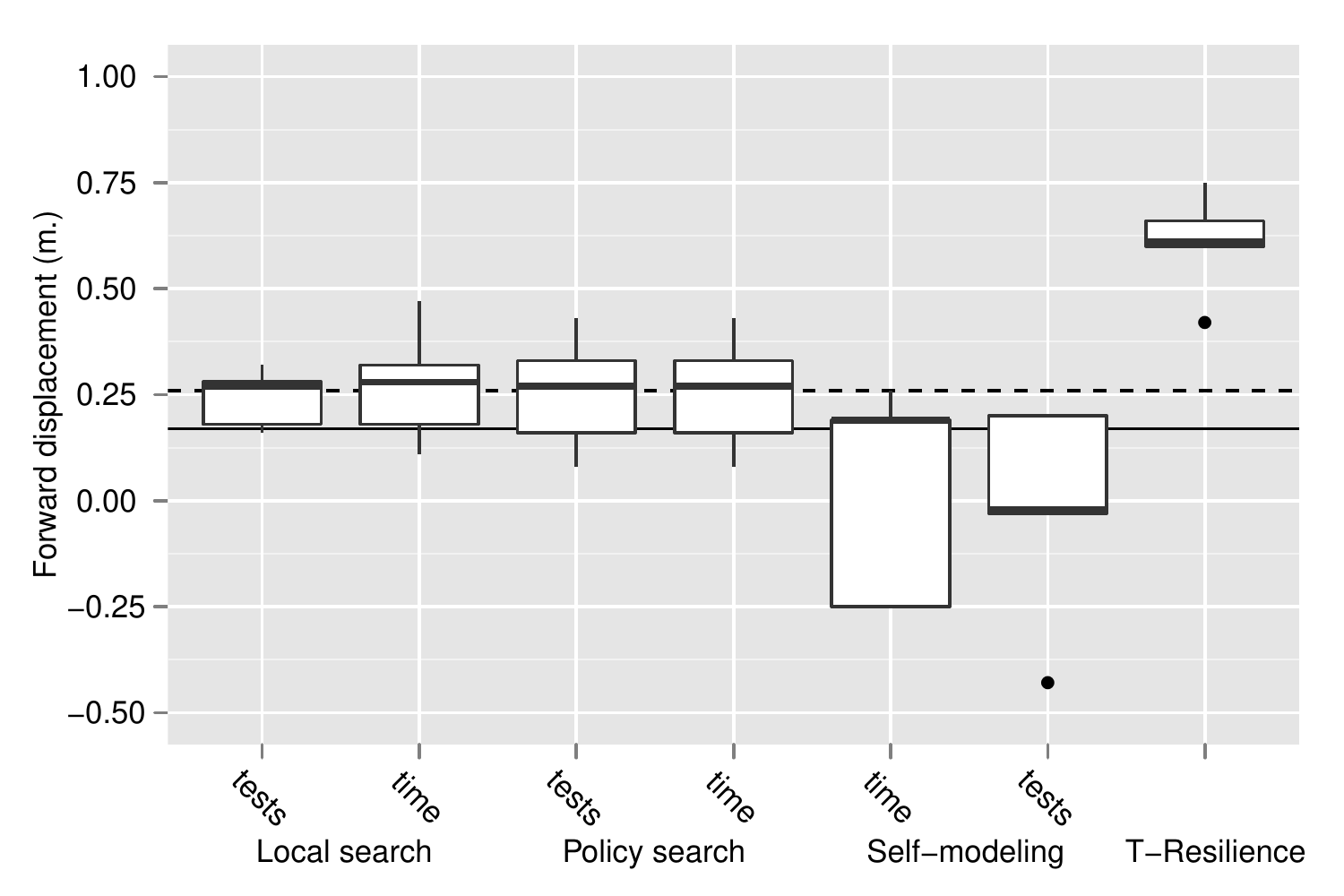}\label{fig:results_unpowered}}\\
\subfloat[Front right leg shortened by half (case C).]{\includegraphics[width=\columnwidth]{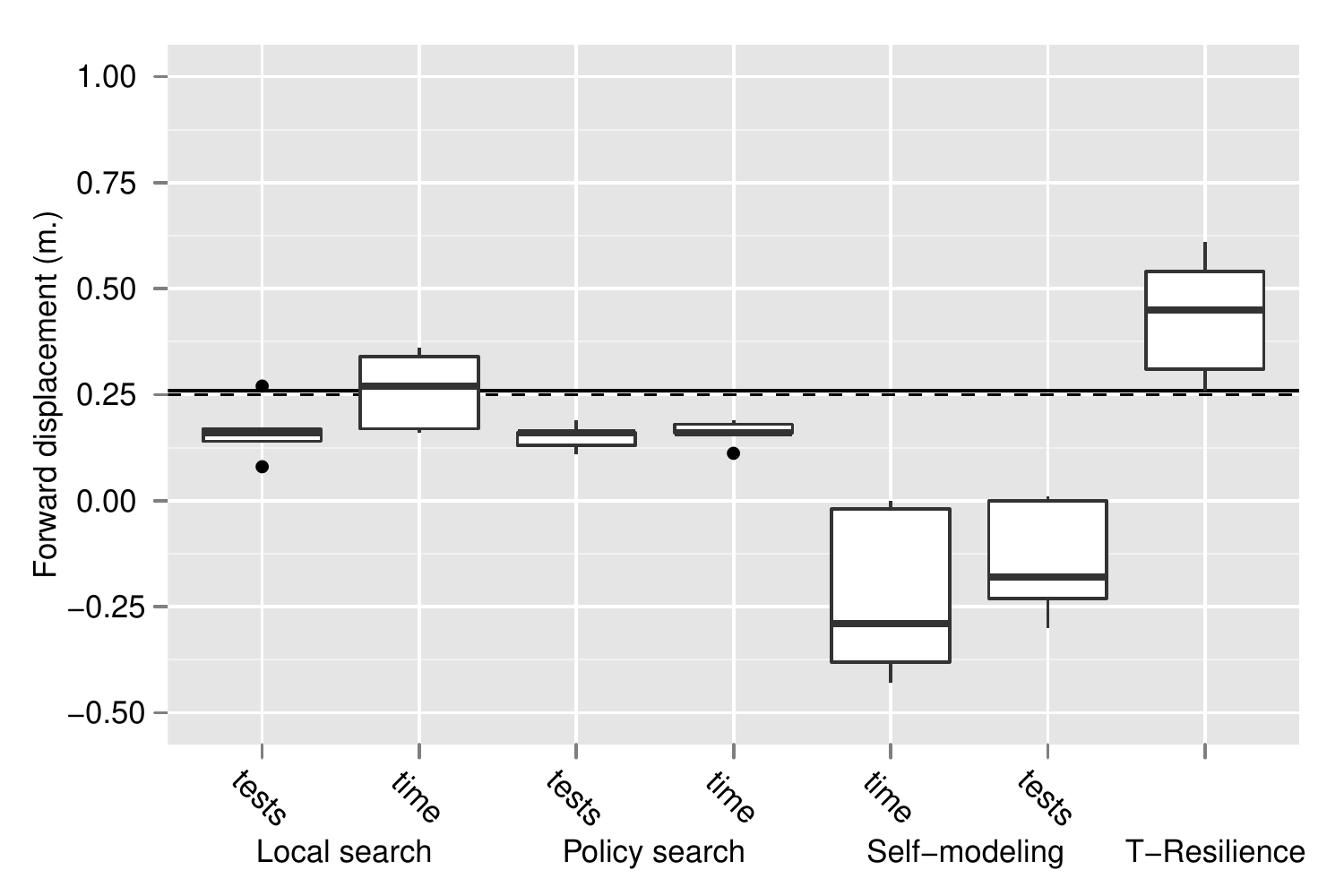}\label{fig:results_shortened}}
\subfloat[Hind right leg lost (case D).]{\includegraphics[width=\columnwidth]{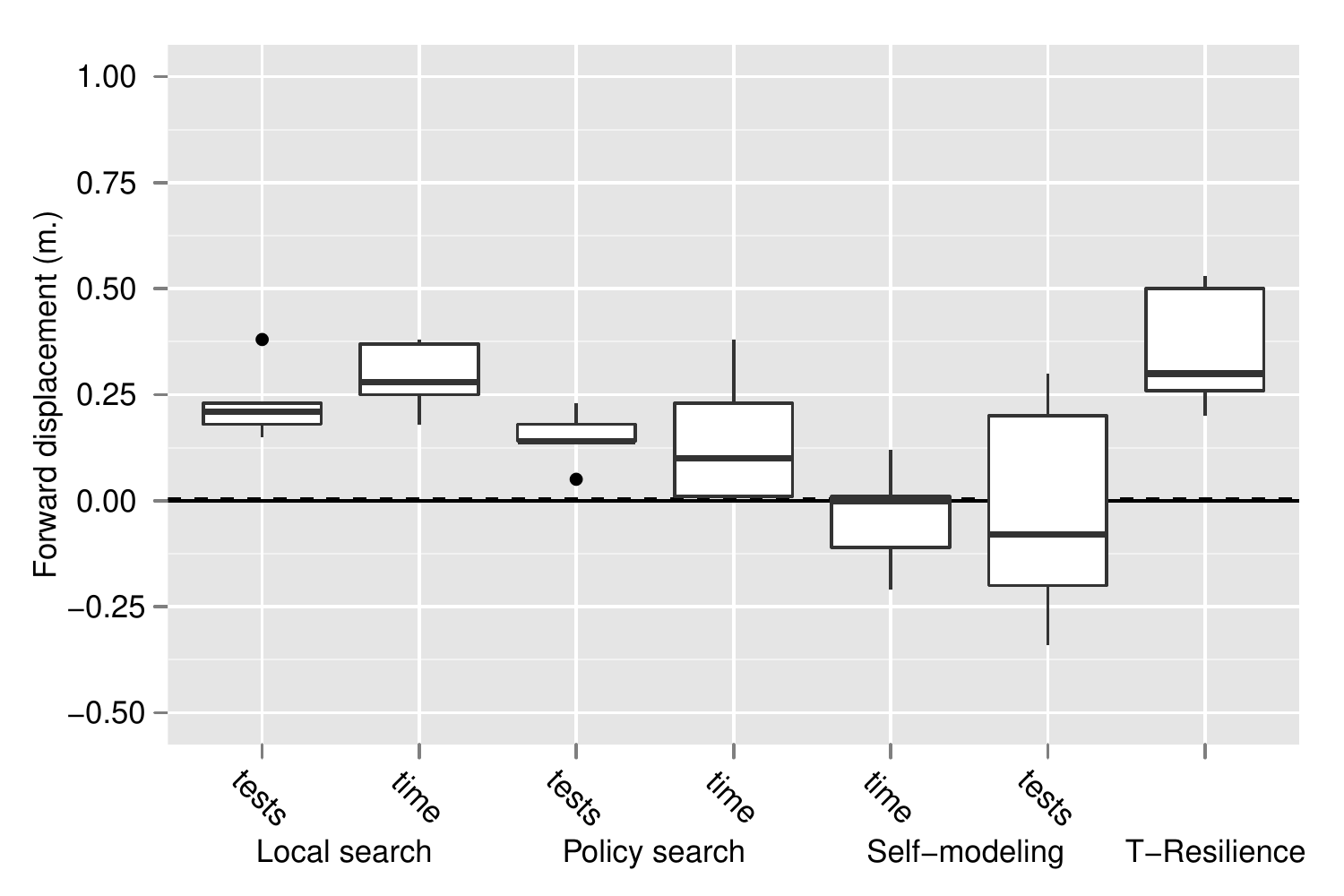}\label{fig:results_lostleg}}\\
\subfloat[Middle right leg lost (case E).]{\includegraphics[width=\columnwidth]{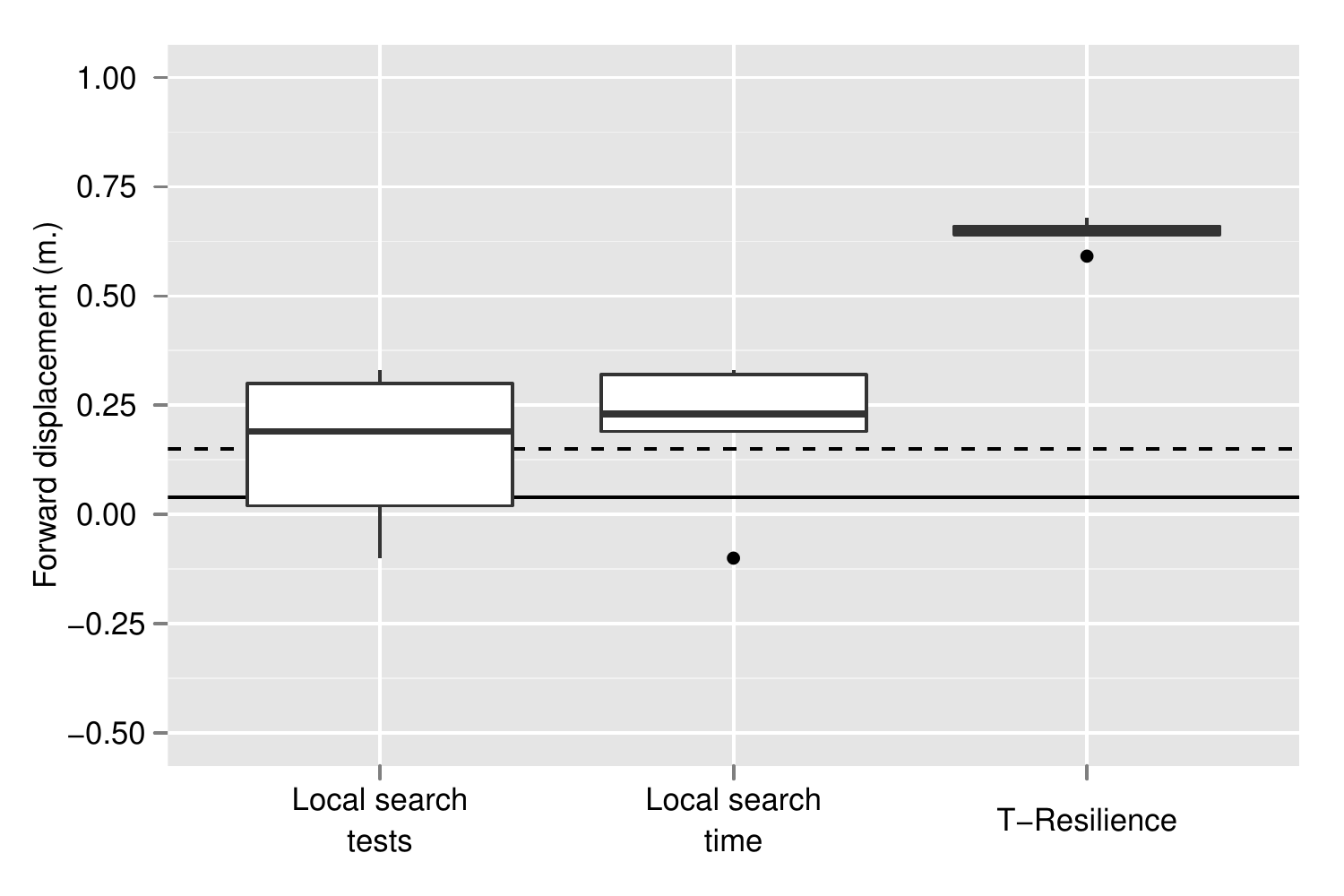}\label{fig:results_middlelostleg}}
\subfloat[Middle right leg and front left leg lost (case F).]{\includegraphics[width=\columnwidth]{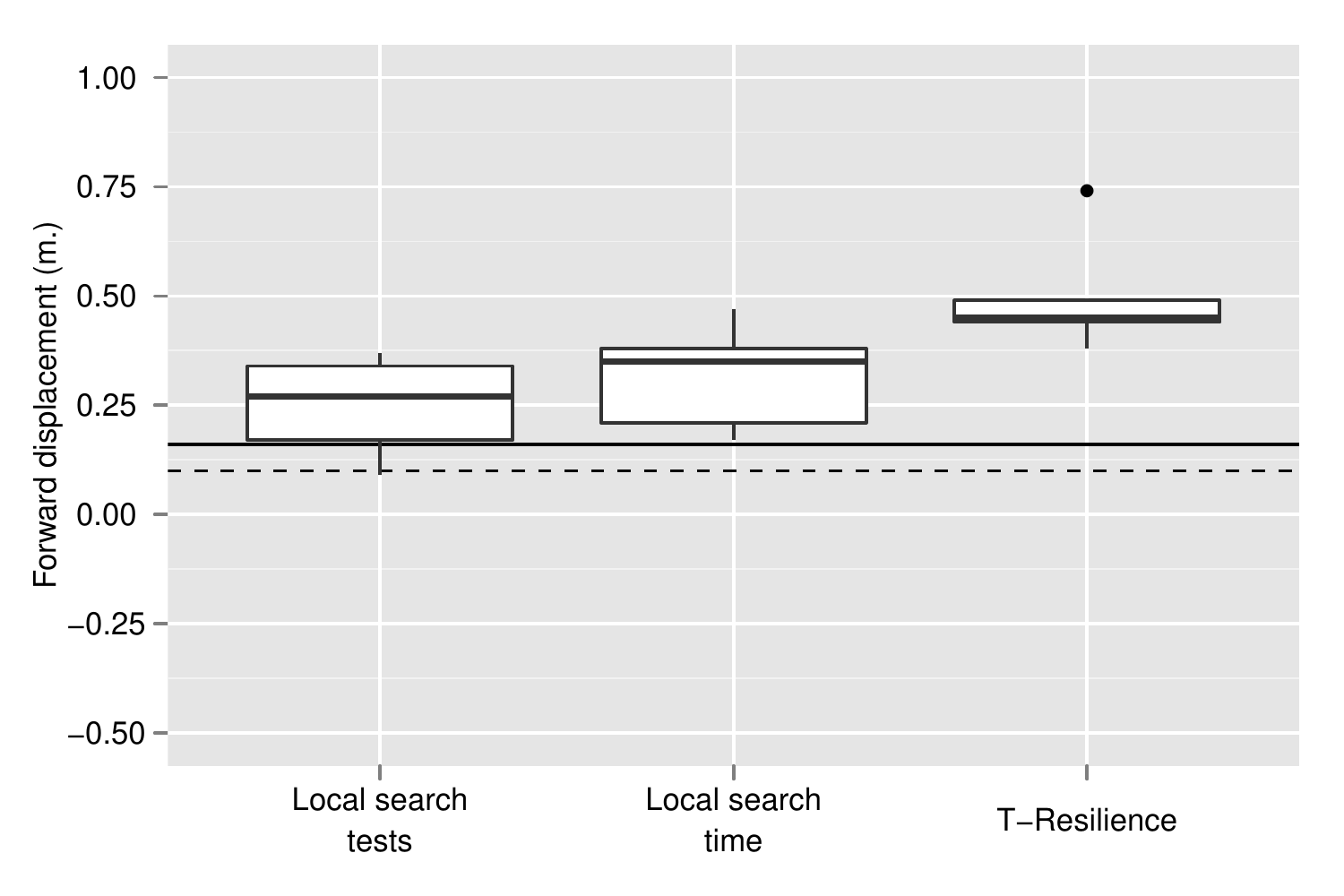}\label{fig:results_twolostlegs}}
\caption{Performances obtained in each test cases (distance covered in
  3 seconds). On each box, the central mark is the median, the edges
  of the box are the lower hinge (defined as the 25th percentile) and
  the upper hinge (the 75th percentile). The whiskers extend to the
  most extreme data point which is no more than 1.5 times the length
  of the box away from the box. Each algorithm has been run 5 times
  and distances are measured using the external motion capture
  system. Except for the T-Resilience, the performance of the
  controllers found after about 25 transfers ($tests$) and after about
  20 minutes ($time$) are depicted (all T-Resilience experiments last
  about 20 minutes and use 25 transfers). The horizontal lines denote
  the performances of the reference gait, according to the CODA
  scanner (dashed line) and according to the SLAM algorithm (solid
  line).}
\label{fig:all_results}
\end{figure*}

\begin{table*}
\centering

\subfloat[Ratios between median performance values.]{\begin{tabular}{|c|cc|cc|cc|c|}
\cline{2-8}
\multicolumn{1}{c|}{} & \multicolumn{2}{c|}{Local search} & \multicolumn{2}{c|}{Policy search} & \multicolumn{2}{c|}{Self-modeling} & \multirow{2}{*}{reference gait}\\
\multicolumn{1}{c|}{} &  \multicolumn{1}{c}{tests} &  \multicolumn{1}{c|}{time} &  \multicolumn{1}{c}{tests} &  \multicolumn{1}{c|}{time} &  \multicolumn{1}{c}{time} &  \multicolumn{1}{c|}{tests} &\\
\hline
A & 4.5 & 2.5 & 3.3 & 4.5 & 6.3 & 19.0 & 1.0\\
B & 2.3 & 2.2 & 2.3 & 2.3 & +++ &  3.2 & 2.3\\
C & 2.8 & 1.7 & 2.8 & 2.8 & +++ &  +++ & 1.8\\
D & 1.4 & 1.1 & 2.1 & 3.0 & +++ &  +++ & +++\\
E & 3.4 & 2.8 & & & & & 4.3\\
F & 1.7 & 1.3 & & & & & 4.5\\
\hline
global median & 2.8 & 2.0 & 2.6 & 2.9 & +++ & +++ & 3.3\\
\hline
\end{tabular}\vspace{10pt}}

\subfloat[Differences between median performance values (cm).]{\begin{tabular}{|c|cc|cc|cc|c|}
\cline{2-8}
\multicolumn{1}{c|}{} & \multicolumn{2}{c|}{Local search} & \multicolumn{2}{c|}{Policy search} & \multicolumn{2}{c|}{Self-modeling} & \multirow{2}{*}{reference gait}\\
\multicolumn{1}{c|}{} &  \multicolumn{1}{c}{tests} &  \multicolumn{1}{c|}{time} &  \multicolumn{1}{c}{tests} &  \multicolumn{1}{c|}{time} &  \multicolumn{1}{c}{time} &  \multicolumn{1}{c|}{tests} &\\
\hline
A & +59 & +45 & +53 & +59 & +64 & +72 & - 2\\
B & +34 & +33 & +34 & +34 & +63 & +42 & +35\\
C & +29 & +18 & +29 & +29 & +63 & +74 & +20\\
D & + 9 & + 2 & +16 & +20 & +38 & +30 & +30\\
E & +46 & +42 & & & & & +50\\
F & +18 & +10 & & & & & +35\\
\hline
global median & +32 & +26 & +32 & +32 & +63 & +57 & +33\\
\hline
\end{tabular}}

\caption{Performance improvements of the T-Resilience compared to
  other algorithms. For ratios, the symbol +++ indicates that the
  compared algorithm led to a negative or null median value.}
\label{table:performances_improvements}
\end{table*}

\subsection{Comparison of performances}
\label{sec:results}
Fig.~\ref{fig:all_results} shows the performance obtained for all
test cases and all the investigated
algorithms. Table~\ref{table:performances_improvements} reports the
improvements between median performance values. P-values are computed with
the Wilcoxon rank-sum tests (appendix
~\ref{appendix:p_values}). The horizontal lines in Figure
\ref{fig:all_results} show the efficiency of the reference gait in
each case.

The trajectories corresponding to controllers with median performance
values obtained with the T-Resilience are depicted on
figure~\ref{fig:trajboth}. Videos of the typical behaviors obtained
with the T-Resilience on every test case are available in extension
(Extensions 1 to 9).

\paragraph{Performance with the undamaged robot (case A).}
When the robot is not damaged, the T-Resilience algorithm discovered
controllers with the same level of performance than the reference
hexapod gait (p-value = 1). The obtained controllers are from 2.5 to
19 times more efficient than controllers obtained with other
algorithms (Table~\ref{table:performances_improvements}).

The poor performance of the other algorithms may appear surprising at
first sight.  Local search is mostly impaired by the very low number
of tests that are allowed on the robot, as suggested by the better
performance of the ``time'' variant (20 minutes / 50 tests) versus the
``tests'' variant (10 minutes / 25 tests). Surprisingly, we did not
observe any significant difference when we initialized the control
parameters with those of the reference controller (data not
shown). The policy gradient method suffers even more than local search
from the low number of tests because a lot of tests are required to
estimate the gradient. As a consequence, we were able to perform only
2 to 4 iterations of the algorithm. Overall, these results are
consistent with those of the literature because previous experiments
used longer experiments and often simpler systems. Similar
observations have been reported previously by other
authors~\citep{Yosinski2011}.

Bongard's algorithm mostly fails because of the reality gap between
the self-model and the real robot. Optimizing the behavior only in
simulation leads -- as expected -- to controllers that perform well
with the self-model but that do not work on the real robot. This
performance loss is sometimes high because the controllers make the
robot fall of over or go backward.

\paragraph{Resilience performance (cases B to F).}

When the robot is damaged, gaits found with the T-Resilience algorithm
are always faster than the reference gait ($p = 0.0625$, one-sample
Wilcoxon signed rank test).

After the same number of tests (variant $tests$ of each algorithm),
gaits obtained with T-Resilience are at least 1.4 times faster than
those obtained with the other algorithms (median of 3.0 times) with
median performance values from 30 to 65 cm in 3 seconds. These
improvements are all stastically significant (p <= 0.016) except for
the local search in the case D (loss of a hind leg; p = 0.1508).

After the same running time (variant $time$ of each algorithm), gaits
obtained with T-Resilience are also significantly faster (at least 1.3
times; median of 2.8 times; p <= 0.016) than those obtained with the
other algorithms in cases B, E and F. In cases C (shortened leg) and D
(loss of a hind leg), T-Resilience is not statistically different from
local search (shortened leg: p = 0.1508; loss of a hind leg: p =
0.5476). Nevertheless, these high p-values may stem from the low
number of replications (only $5$ replications for each
algorithm). Moreover, as section~\ref{section:duration} will show, the
execution time of the T-Resilience can be compressed because a large
part of the running time is spent in computer
simulations. Consequently, depending on the hardware, better
performances could be achieved in smaller amounts of time.

For all the tested cases, Bongard's self-modeling algorithm doesn't
find any working controllers. We observed that it suffers from two
difficulties: the optimized models do not always capture the actual
morphology of the robot, and reality gaps between the self-model and
the reality (see the comments about the undamaged robot). In the first
case, more time and more actions could improve the result. In the
second time, a better simulation model could make things better but it
is unlikely to fully remove the effect of the reality gap.

\begin{figure*}
\centering
\subfloat[Distribution of duration (median duration indicated below the graph).]{\includegraphics[width=\columnwidth]{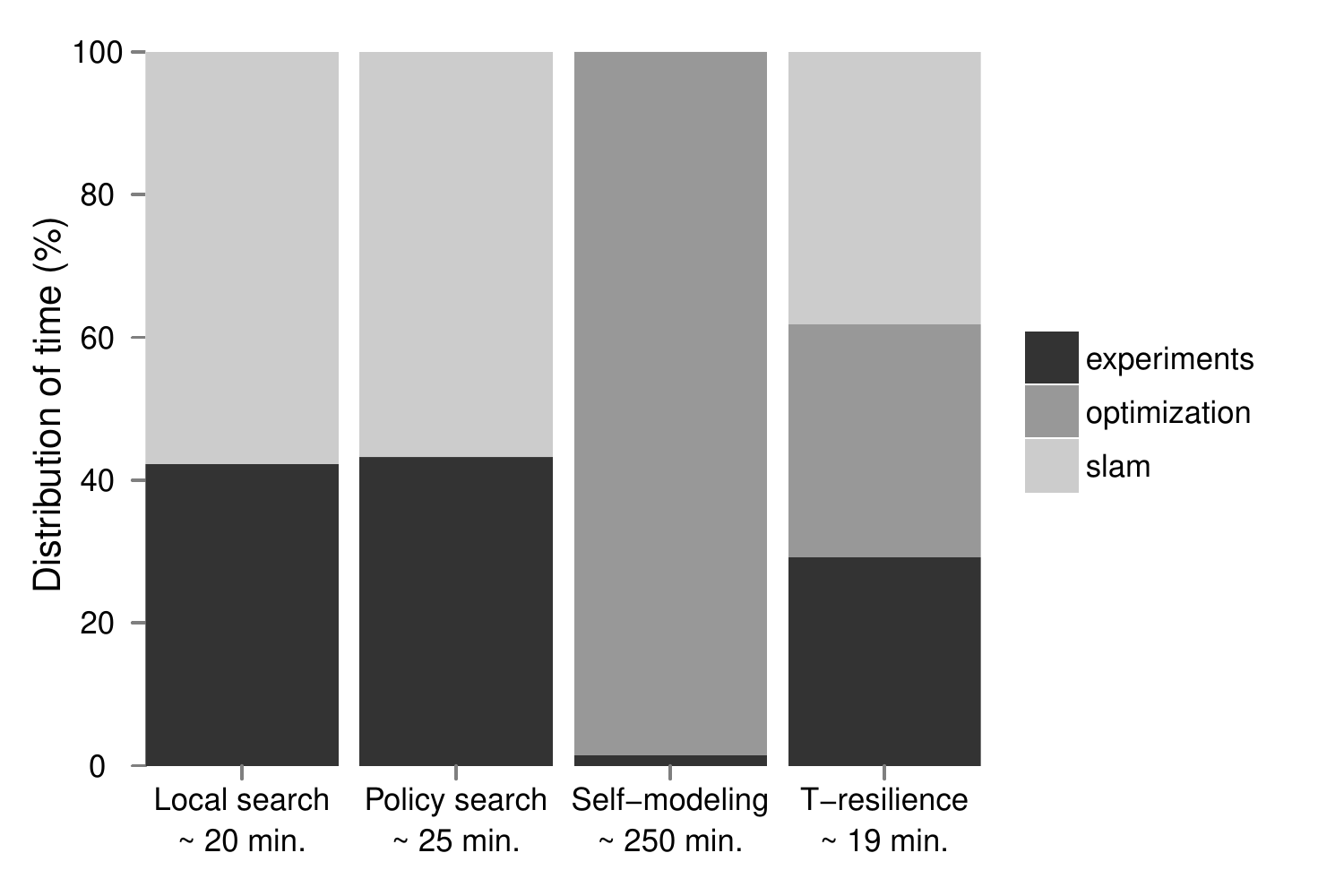}\label{fig:repartition_duration}}
\subfloat[Experimental time (experiments with the robot).]{\includegraphics[width=\columnwidth]{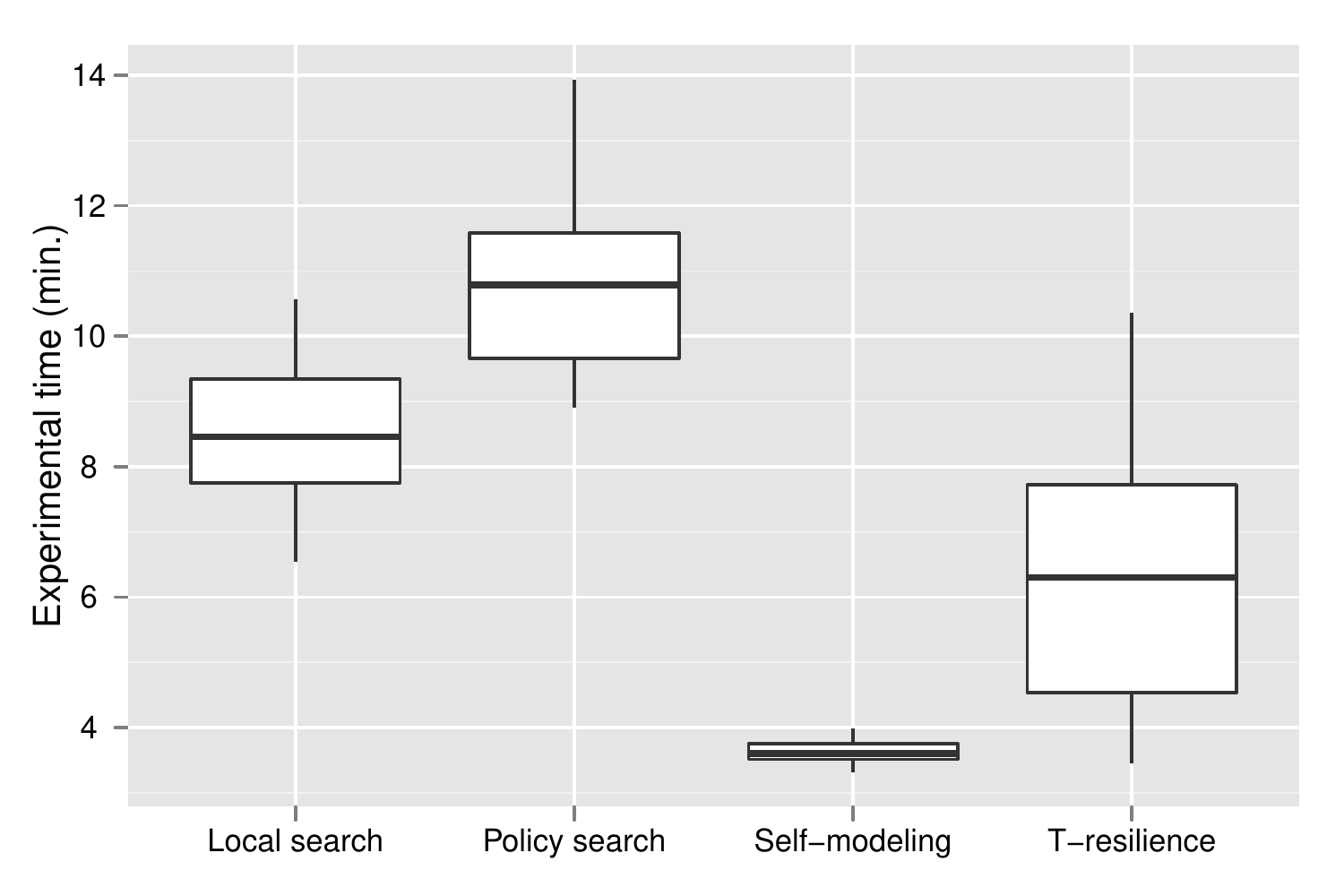}\label{fig:experimental_time}}
\caption{Distribution of duration and experimental time for each
  algorithm (median values on 5 runs of test cases A, B, C, D). All
  the differences between experimental times are statistically
  significant (p-values $< 2.5\times10^{-4}$ with Wilcoxon rank-sum
  tests).}
\label{fig:repartition_time}

\end{figure*}

\paragraph{Loss of a leg (case D and E).}
When the hind leg is lost, the T-Resilience leads controllers that
perform much better than the reference controller. Nevertheless, the
performances of the controller obtained with the T-Resilience are not
statistically different from those obtained with the local
search. These results stems from the fact that many of the transfers
made the robot tilt down (about half of the 25 transfers), whereas
these tests are ignored in our current implementation. This make the
estimation of the transferability function especially difficult. Fast
six-legged behaviors optimized on the self-model of the undamaged
robot are indeed often unstable without any of the hind legs. The
process has to use several transfers to find a transferability
function that allows the avoidance of all the gaits that use the hind
leg. Moreover, in this scenario, performances predicted by the
self-model and transferability values are clearly antagonistic. Since
transferred solutions are randomly chosen in the population and the
multi-objective optimization algorithm maintains a large spectrum of
Pareto-optimal trade-offs, transferred controllers have a high chance
to have a high performance score but a low and mostly uninformative
transferability score.

If the robot loses a less critical leg (middle leg in
case E), the T-Resilience is able to find fast gaits (about 3 times
faster than with the local search). This failure does not critically
impair the stability of the robot and the algorithm can conduct
informative real tests on the robot.

\paragraph{Straightness of the trajectories.} 
For all the damages, the gaits found by T-Resilience do not result in
trajectories that are perfectly aligned with the x-axis
(Fig.~\ref{fig:trajboth}, deviations from $10$ to $20$ cm). The
deviations mainly stem from the choice of the performance function and
they should not be overinterpreted as a weakness of the T-Resilience
algorithm. Indeed, the performance function only rewards the covered
distance during $3$ seconds and nothing is explicitly rewarding the
straightness of the trajectory. At first, it seems intuitive that the
fastest trajectories will necessarily be straight. However, the
fastest gaits achievable with this specific robot are unknown and
faster gaits may be achievable if the robot is not pointing in the
arbitrarily-chosen x-direction. For instance, the trajectory found
with the undamaged robot (Fig.~\ref{fig:trajboth}(a)) deviates from
the x-axis, but it has the same final performance as the reference
gait, which is perfectly straight. These two gaits cannot be
distinguished by the performance function and we have no way to know
if both faster and straighter trajectories are possible in our
system. The intuition that the straightest trajectories should be the
fastest is even more challenged for the damaged robots because the
robots are not symmetric anymore. In future work, we will investigate
alternative performance functions that encourage straight
trajectories.

Moreover, trajectories are actually mostly straight
(Fig.~\ref{fig:trajboth} and the videos in appendix)) but they do not
exactly point to the x-direction. This direction seems to be mainly
determined by the position of the robot at the beginning of the
experiment: at $t=0$, each degree of freedom is positioned according
to the value of the control function, that is, the robot often starts
in the middle of the gait pattern; because this position is
non-symmetric and sometimes unstable, we often observed that the first
step often makes the robot point in a different direction. Once the
gait is started, deviations along one directions are compensated by
symmetrical deviations at the next step and the gaits are mostly
straight.

\subsection{Comparison of durations and experimental time}
\label{section:duration}
The running time of each algorithm is divided into experimental time
(actual experiments on the robot), sensor processing time (computing
the robot's trajectory using RGB-D slam) and optimization time
(generating new potential solutions to test on the robot). The median
proportion of time allocated to each of this part of the algorithms is
pictured for each algorithm on
figure~\ref{fig:repartition_time}\footnote{Only test cases A, B, C and
  D are considered to compute these proportions (5 runs for each
  algorithm) because the policy search and the self-modeling process
  are not tested in test cases E and F}.

The durations of the SLAM algorithm and of the optimization processes
both only depend on the hardware specifications and can therefore be
substantially reduced by using faster computers or by parallelizing
computation. Only experimental time can not easily be reduced. The
median proportion of experimental time is 29\% for the T-Resilience,
whereas both the policy search and the local search leads to median
proportions higher than 40\% for a similar median duration by run
(about 20 minutes). The proportion of experimental time for the
self-modeling process is much lower (median value equals to 1\%)
because it requires much more time for each run (about 250 minutes for
each run, in our experiments).

The median experimental time of T-Resilience (6.3 minutes) is
significantly lower than those of local search and of policy search
(resp. 8.5 and 10.8 minutes, p-values of Wilcoxon rank-sum tests $<
2.5\times10^{-4}$). With the expected increases of computational
power, this difference will increase each year. The self-modeling
process requires significantly lower experimental time (median at 3.6
minutes, p-values of Wilcoxon rank-sum-tests $< 1.5\times10^{-11}$)
because it only test actions that involve a single leg, which is
faster than testing a full gait ($3$ second).

%\begin{figure*}
%\centering
%\subfloat[Undamaged hexapod robot.]{\includegraphics[width=\columnwidth]{img/traj_Tresilience_undamaged.pdf}\label{fig:traj_Tresilience_undamaged}}
%\subfloat[Middle left leg not powered.]{\includegraphics[width=\columnwidth]{img/traj_Tresilience_unpowered.pdf}\label{fig:traj_Tresilience_unpowered}}\\
%\subfloat[Front right leg shortened by half.]{\includegraphics[width=\columnwidth]{img/traj_Tresilience_shortened.pdf}\label{fig:traj_Tresilience_shortened}}
%\subfloat[Hind right leg lost.]{\includegraphics[width=\columnwidth]{img/traj_Tresilience_lostleg.pdf}\label{fig:traj_Tresilience_lostleg}}\\
%\subfloat[Middle right leg lost.]{\includegraphics[width=\columnwidth]{img/traj_Tresilience_middlelostleg.pdf}\label{fig:traj_Tresilience_middlelostleg}}
%\subfloat[Middle right leg and front left leg lost.]{\includegraphics[width=\columnwidth]{img/traj_Tresilience_twolostlegs.pdf}\label{fig:traj_Tresilience_twolostlegs}}
%\caption{Trajectories corresponding to controllers with median performance values obtained with the T-Resilience algorithm in all the test cases.}
%\label{fig:trajTresilience}
%\end{figure*}

\section{Conclusion and discussion}
% JBM : see avoidance learning
% TODO rework

All our experiments show that T-Resilience is a fast and efficient
learning approach to discover new behaviors after mechanical and
electrical damages (less than 20 minutes with only 6 minutes of
irreducible experimental time). Most of the time, T-Resilience leads
to gaits that are several times better than those obtained with direct
policy search, local search and Bongard's algorithm; T-Resilience
never obtained worse results. Overall, T-Resilience appears to be a
versatile algorithm for damage recovery, as demonstrated by the
successful experiments with many different types of damages. These
results validate the combination of the principles that underly our
algorithm: (1) using a self-model to transform experimental time with
the robot into computational time inside a simulation (2) learning a
transferability function that predicts performance differences between
reality and the self-model (instead of learning a new self-model) and,
(3) optimizing both the transferability and performance to learn
behaviors in simulation that will work well on the real robot, even if
the robot is damaged. These principles can be implemented with
alternative learning algorithms and alternative regression
models. Future work will identify whether better performance can be
achieved by applying the same principles with other machine learning
techniques.

During our experiments, we observed that the T-Resilience algorithm
was less sensitive to the quality of the SLAM than the other
investigated learning algorithms (policy gradient and local
search). Our preliminary analysis shows that the sensitivity of these
classic learning algorithms mostly stems from the fact they optimize
the SLAM measurements and not the real performance. For instance, in
several of our experiments, the local search algorithm found gaits
that make the SLAM algorithm greatly over-estimate the forward
displacement of the robot. The T-Resilience algorithm relies only on
internal sensors as well. However, these measures are not used to
estimate the performance but to compute the transferability
values. Gaits that lead to over-estimations of the covered distance
have low transferability scores because the measurement greatly
differs from the value predicted by the self-model. As a consequence,
they are avoided like all the behaviors for which the prediction of
the self-model does not match the measurement. In other words, the
self-model acts as a ``credibility check'' of the SLAM measurements,
which makes T-Resilience especially robust to sensor inaccuracies. If
sensors were redundant, this credibility check could also be used by
the robot to continue its mission when a sensor is unavailable.

%% A more frequent problem arises when inaccuracies of internal sensors
%% leads a transferable controller to be perceived as not transferable by
%% the algorithm. In this case, the T-Resilience process would simply
%% avoid it and focus on other transferable controllers.

%% Failures on the internal sensors can critically impair the performances of resilience approaches. Considering the T-Resilience, as the transferability function is directly estimated with internal measures of the forward displacement of the robot, failures on the RGB-D camera would lead to inaccurate estimations of the transferability function and impair discovery and selection of good trade-off controllers. To prevent such a situation, the robot has to :
%% \begin{itemize}
%% \item compute redundant estimations of the variables used to build the transferability function by using redundant sensors (e.g. forward displacement computed with a SLAM algorithm and odometry techniques);
%% \item detect which sensors are faulty and rely on the effective ones.
%% \end{itemize}

%TODO transition

The T-Resilience algorithm is designed to perform well with an
inaccurate self-model, but the better the self-model is, the more
efficient the algorithm is. For instance, in the case of the lost hind
leg, the stability properties of the robot were too different from
those of the self-model and efficient behaviors could hardly be found
in a limited amount of time. If the situation of the robot implies
that such a strong failure can not be repaired, then the T-Resilience
would benefit from an update of the self-model. In some extreme cases,
the T-Resilience algorithm could also be unable to find any satisfying
controller. To identify a new self model, the robot could launch
Bongard's algorithm and then use the updated model with the
T-Resilience algorithm -- to avoid reality gaps between the model and
the real robot. It seems also possible to take advantage of the data
recorded by the internal sensors during the T-Resilience transfers to
design a new self-model that would minimize the transferability
score. The combination of T-Resilience and update of the self-model
will be addressed in future work.

Overall, learning to predict what behaviors should be avoided is a
very general concept that can be applied to many situations in which a
robot has to autonomously adapt its behavior. On a higher level, this
concept could also share some similarities with what human do when
they are injured: if a movement is painful, humans do not fully
understand what cause the pain, but they identify the behaviors that
cause the pain\footnote{In the same way as learning a transferability
  function is easier than learning a new self-model, finding the cause
  of a pain requires a medical doctor whereas a patient can usually
  predict whether a move will be painful.}; once they know that some
move are painful, they learn to instinctively avoid them. Humans seem
reluctant to permanently change their self-model to reflect what
behaviors are possible: people with an immobilized leg still know how
to activate their muscles, amputated people frequently report pain in
their missing members~\citep{Ramachandran1998} and dream about
themselves in their intact body~\citep{Mulder2008}. Humans may
therefore learn by combining their self-model with a second model that
predicts which behaviors should be avoided, even if they are
possible. This model would be similar in essence to a transferability
function.

%Moreover, these
%people are not only planning new movements with a constrained
%self-model, they learn new, unconscious gaits and other habit
%movements. 

%% A simple way would rely on the data recorded with internal sensors
%% from the transfers occuring during the T-Resilience to off-line
%% identify a new morphological model of the robot starting from the
%% model of the undamaged robot. Assuming that the identification is
%% correct, the T-Resilience would next find better behaviors as the gap
%% between the self-model and the robot is clearly reduced.

\section*{Acknowledgements}
The authors thank Jeff Clune and St\'ephane Doncieux for helpful comments and suggestions.

\section*{Funding}
This work was supported by the ANR (project Creadadapt, ANR-12-JS03-0009); a UPMC-DGA
scholarship to A. Cully; and Polytech'Paris-UPMC.

\bibliographystyle{plainnat}
%\bibliography{resilience}

\begin{thebibliography}{90}
\providecommand{\natexlab}[1]{#1}
\providecommand{\url}[1]{\texttt{#1}}
\expandafter\ifx\csname urlstyle\endcsname\relax
  \providecommand{\doi}[1]{doi: #1}\else
  \providecommand{\doi}{doi: \begingroup \urlstyle{rm}\Url}\fi

\bibitem[Argall et~al.(2009)Argall, Chernova, Veloso, and Browning]{Argall2009}
B.~Argall, S.~Chernova, M.~Veloso, and B.~Browning.
\newblock A survey of robot learning from demonstration.
\newblock \emph{Robotics and Autonomous Systems}, 57\penalty0 (5):\penalty0
  469--483, 2009.

\bibitem[Barfoot et~al.(2006)Barfoot, Earon, and
  D'Eleuterio]{barfoot2006experiments}
T.~Barfoot, E.~Earon, and G.~D'Eleuterio.
\newblock Experiments in learning distributed control for a hexapod robot.
\newblock \emph{Robotics and Autonomous Systems}, 54\penalty0 (10):\penalty0
  864--872, 2006.

\bibitem[Bellingham and Rajan(2007)]{Bellingham2007}
J.~G. Bellingham and K.~Rajan.
\newblock {Robotics in remote and hostile environments.}
\newblock \emph{Science}, 318\penalty0 (5853):\penalty0 1098--102, November
  2007.
\newblock ISSN 1095-9203.

\bibitem[Berenson et~al.(2005)Berenson, Estevez, and
  Lipson]{berenson2005hardware}
D.~Berenson, N.~Estevez, and H.~Lipson.
\newblock Hardware evolution of analog circuits for in-situ robotic
  fault-recovery.
\newblock In \emph{Proceedings of NASA/DoD Conference on Evolvable Hardware},
  pages 12--19, 2005.

\bibitem[Bongard(2007)]{bongard2007action}
J.~Bongard.
\newblock Action-selection and crossover strategies for self-modeling machines.
\newblock In \emph{Proceedings of Genetic and Evolutionary Computation
  Conference (GECCO)}, pages 198--205. ACM, 2007.

\bibitem[Bongard and Lipson(2005)]{bongard2005nonlinear}
J.~Bongard and H.~Lipson.
\newblock Nonlinear system identification using coevolution of models and
  tests.
\newblock \emph{IEEE Transactions on Evolutionary Computation}, 9\penalty0
  (4):\penalty0 361--384, 2005.

\bibitem[Bongard et~al.(2006)Bongard, Zykov, and Lipson]{bongard2006resilient}
J.~Bongard, V.~Zykov, and H.~Lipson.
\newblock Resilient machines through continuous self-modeling.
\newblock \emph{Science}, 314\penalty0 (5802):\penalty0 1118--1121, 2006.

\bibitem[Caccavale and Villani(2002)]{Caccavale2002}
F.~Caccavale and L.~Villani, editors.
\newblock \emph{Fault Diagnosis and Fault Tolerance for Mechatronic Systems:
  Recent Advances}.
\newblock springer, 2002.

\bibitem[Cantu-Paz(2000)]{Cantu2000}
E.~Cantu-Paz.
\newblock \emph{{Efficient and accurate parallel genetic algorithms}}.
\newblock Kluwer Academic Publishers Norwell, MA, USA, 2000.

\bibitem[Chang and Lin(2011)]{chang2011libsvm}
C.~Chang and C.~Lin.
\newblock Libsvm: a library for support vector machines.
\newblock \emph{ACM Transactions on Intelligent Systems and Technology (TIST)},
  2\penalty0 (3):\penalty0 27, 2011.

\bibitem[Chernova and Veloso(2004)]{chernova2004evolutionary}
S.~Chernova and M.~Veloso.
\newblock An evolutionary approach to gait learning for four-legged robots.
\newblock In \emph{Proceedings of IEEE/RSJ International Conference on
  Intelligent Robots and Systems (IROS)}, volume~3, pages 2562--2567. IEEE,
  2004.

\bibitem[Clune et~al.(2011)Clune, Stanley, Pennock, and Ofria]{Clune2011}
J.~Clune, K.~Stanley, R.~Pennock, and C.~Ofria.
\newblock On the performance of indirect encoding across the continuum of
  regularity.
\newblock \emph{Evolutionary Computation, IEEE Transactions on}, 15\penalty0
  (3):\penalty0 346--367, 2011.

\bibitem[Connell and Mahadevan(1993)]{Connell1993}
J.~Connell and S.~Mahadevan.
\newblock \emph{Robot learning}.
\newblock Springer, 1993.

\bibitem[Corbato(2007)]{Corbato2007}
F.~Corbato.
\newblock {On Building Systems That Will Fail}.
\newblock \emph{ACM Turing award lectures}, 34\penalty0 (9):\penalty0 72--81,
  2007.

\bibitem[De~Jong(2006)]{de2006evolutionary}
K.~De~Jong.
\newblock \emph{Evolutionary computation: a unified approach}.
\newblock MIT Press, 2006.

\bibitem[Deb(2001)]{deb2001}
K.~Deb.
\newblock \emph{Multi-objective optimization using evolutionary algorithms}.
\newblock John Wiley and Sons, 2001.

\bibitem[Deb et~al.(2002)Deb, Pratap, Agarwal, and Meyarivan]{deb2002fast}
K.~Deb, A.~Pratap, S.~Agarwal, and T.~Meyarivan.
\newblock A fast and elitist multiobjective genetic algorithm: {NSGA-II}.
\newblock \emph{IEEE Transactions on Evolutionary Computation}, 6\penalty0
  (2):\penalty0 182--197, 2002.

\bibitem[Delcomyn(1971)]{Delcomyn1971}
F.~Delcomyn.
\newblock {The Locomotion of the Cockroach Pariplaneta americana}.
\newblock \emph{Journal of Experimental Biology}, 54\penalty0 (2):\penalty0
  443--452, 1971.

\bibitem[Ding et~al.(2010)Ding, Wang, Rovetta, and Zhu]{ding2010locomotion}
X.~Ding, Z.~Wang, A.~Rovetta, and J.~Zhu.
\newblock Locomotion analysis of hexapod robot.
\newblock \emph{Proceedings of Conference on Climbing and Walking Robots
  (CLAWAR)}, pages 291--310, 2010.

\bibitem[Doncieux et~al.(2011)Doncieux, Mouret, Bredeche, and
  Padois]{Doncieux2011}
S.~Doncieux, J.-B. Mouret, N.~Bredeche, and V.~Padois.
\newblock Evolutionary robotics: Exploring new horizons.
\newblock In \emph{New Horizons in Evolutionary Robotics: Extended
  Contributions from the 2009 EvoDeRob Workshop.}, pages 3--25. Springer, 2011.

\bibitem[Endres et~al.(2012)Endres, Hess, Engelhard, Sturm, Cremers, and
  Burgard]{endres12icra}
F.~Endres, J.~Hess, N.~Engelhard, J.~Sturm, D.~Cremers, and W.~Burgard.
\newblock An evaluation of the {RGB-D SLAM} system.
\newblock In \emph{Proceedings of the IEEE International Conference on Robotics
  and Automation (ICRA)}, 2012.

\bibitem[Goldberg and Chen(2001)]{Goldberg2001}
K.~Goldberg and B.~Chen.
\newblock Collaborative control of robot motion: robustness to error.
\newblock In \emph{Proceedings of IEEE/RSJ International Conference on
  Intelligent Robots and Systems (IROS)}, volume~2, pages 655--660, 2001.
\newblock ISBN 0-7803-6612-3.

\bibitem[G{\"o}rner and Hirzinger(2010)]{gorner2010analysis}
M.~G{\"o}rner and G.~Hirzinger.
\newblock Analysis and evaluation of the stability of a biologically inspired,
  leg loss tolerant gait for six-and eight-legged walking robots.
\newblock In \emph{Proceedings of the IEEE International Conference on Robotics
  and Automation (ICRA)}, pages 4728--4735, 2010.

\bibitem[Grefenstette et~al.(1999)Grefenstette, Schultz, and
  Moriarty]{Grefenstette1999}
J.~J. Grefenstette, A.~C. Schultz, and D.~E. Moriarty.
\newblock {Evolutionary algorithms for reinforcement learning}.
\newblock \emph{Journal of Artificial Intelligence Research}, pages 241--276,
  1999.

\bibitem[Hartland and Bredeche(2006)]{Hartland2006}
C.~Hartland and N.~Bredeche.
\newblock Evolutionary robotics, anticipation and the reality gap.
\newblock In \emph{Robotics and Biomimetics, 2006. ROBIO'06. IEEE International
  Conference on}, pages 1640--1645. IEEE, 2006.

\bibitem[Heidrich-Meisner and Igel(2009)]{Heidrich-Meisner2009a}
V.~Heidrich-Meisner and C.~Igel.
\newblock {Neuroevolution strategies for episodic reinforcement learning}.
\newblock \emph{Journal of Algorithms}, 64\penalty0 (4):\penalty0 152--168,
  October 2009.

\bibitem[Hemker et~al.(2009)Hemker, Stelzer, Von~Stryk, and
  Sakamoto]{hemker2009efficient}
T.~Hemker, M.~Stelzer, O.~Von~Stryk, and H.~Sakamoto.
\newblock Efficient walking speed optimization of a humanoid robot.
\newblock \emph{The International Journal of Robotics Research}, 28\penalty0
  (2):\penalty0 303--314, 2009.

\bibitem[Hoffmann et~al.(2010)Hoffmann, Marques, Arieta, Sumioka, Lungarella,
  and Pfeifer]{Hoffmann2010}
M.~Hoffmann, H.~Marques, A.~Arieta, H.~Sumioka, M.~Lungarella, and R.~Pfeifer.
\newblock {Body Schema in Robotics: A Review}.
\newblock \emph{IEEE Transactions on Autonomous Mental Development}, 2\penalty0
  (4):\penalty0 304--324, 2010.

\bibitem[Holland and Goodman(2003)]{Holland2003}
O.~Holland and R.~Goodman.
\newblock Robots with internal models a route to machine consciousness?
\newblock \emph{Journal of Consciousness Studies}, 10\penalty0 (4-5):\penalty0
  4--5, 2003.

\bibitem[Hoos and St{\"u}tzle(2005)]{hoos2005stochastic}
H.~H. Hoos and T.~St{\"u}tzle.
\newblock \emph{Stochastic local search: Foundations and applications}.
\newblock Morgan Kaufmann, 2005.

\bibitem[Hornby et~al.(2011)Hornby, Lohn, and Linden]{Hornby2011}
G.~S. Hornby, J.~D. Lohn, and D.~S. Linden.
\newblock {Computer-automated evolution of an X-band antenna for NASA's Space
  Technology 5 mission.}
\newblock \emph{Evolutionary computation}, 19\penalty0 (1):\penalty0 1--23,
  January 2011.
\newblock ISSN 1530-9304.

\bibitem[Hornby et~al.(2005)Hornby, Takamura, Yamamoto, and
  Fujita]{hornby2005autonomous}
G.~Hornby, S.~Takamura, T.~Yamamoto, and M.~Fujita.
\newblock Autonomous evolution of dynamic gaits with two quadruped robots.
\newblock \emph{IEEE Transactions on Robotics}, 21\penalty0 (3):\penalty0
  402--410, 2005.

\bibitem[Jakimovski and Maehle(2010)]{jakimovski2010situ}
B.~Jakimovski and E.~Maehle.
\newblock In situ self-reconfiguration of hexapod robot oscar using
  biologically inspired approaches.
\newblock \emph{Climbing and Walking Robots. InTech}, 2010.

\bibitem[Jakobi et~al.(1995)Jakobi, Husbands, and Harvey]{jakobi1995noise}
N.~Jakobi, P.~Husbands, and I.~Harvey.
\newblock {Noise and the reality gap: The use of simulation in evolutionary
  robotics}.
\newblock \emph{Proceedings of the European Conference on Artificial Life
  (ECAL)}, pages 704--720, 1995.

\bibitem[Kajita and Espiau(2008)]{Kajita2008}
S.~Kajita and B.~Espiau.
\newblock \emph{Handbook of Robotics}, chapter Legged Robots, pages 361--389.
\newblock Springer, 2008.

\bibitem[Kati{\'c} and Vukobratovi{\'c}(2003)]{katic2003survey}
D.~Kati{\'c} and M.~Vukobratovi{\'c}.
\newblock Survey of intelligent control techniques for humanoid robots.
\newblock \emph{Journal of Intelligent \& Robotic Systems}, 37\penalty0
  (2):\penalty0 117--141, 2003.

\bibitem[Kimura et~al.(2001)Kimura, Yamashita, and
  Kobayashi]{kimura2001reinforcement}
H.~Kimura, T.~Yamashita, and S.~Kobayashi.
\newblock Reinforcement learning of walking behavior for a four-legged robot.
\newblock In \emph{Proceedings of IEEE Conference on Decision and Control
  (CDC)}, volume~1, pages 411--416. IEEE, 2001.

\bibitem[Klaus et~al.(2012)Klaus, Glette, and T{\o}rresen]{Klaus2012}
G.~Klaus, K.~Glette, and J.~T{\o}rresen.
\newblock A comparison of sampling strategies for parameter estimation of a
  robot simulator.
\newblock \emph{Simulation, Modeling, and Programming for Autonomous Robots},
  pages 173--184, 2012.

\bibitem[Kober and Peters(2012)]{Kober2012}
J.~Kober and J.~Peters.
\newblock Reinforcement learning in robotics: A survey.
\newblock In \emph{Reinforcement Learning: State of the Art}, pages 579--610.
  Springer, 2012.

\bibitem[Kober and Peters(2010)]{Kober2010}
J.~Kober and J.~Peters.
\newblock {Imitation and Reinforcement Learning -- Practical Learning
  Algorithms for Motor Primitives in Robotics}.
\newblock \emph{IEEE Robotics and Automation Magazine}, 17\penalty0
  (2):\penalty0 1--8, 2010.

\bibitem[Kohl and Stone(2004)]{kohl2004policy}
N.~Kohl and P.~Stone.
\newblock Policy gradient reinforcement learning for fast quadrupedal
  locomotion.
\newblock In \emph{Proceedings of the IEEE International Conference on Robotics
  and Automation (ICRA)}, volume~3, pages 2619--2624. IEEE, 2004.

\bibitem[Koos et~al.(2012)Koos, Mouret, and Doncieux]{2012ACLI2214}
S.~Koos, J.-B. Mouret, and S.~Doncieux.
\newblock The transferability approach: Crossing the reality gap in
  evolutionary robotics.
\newblock \emph{IEEE Transactions on Evolutionary Computation}, 1:\penalty0
  1--25, 2012.

\bibitem[Koren and Krishna(2007)]{koren2007fault}
I.~Koren and C.~M. Krishna.
\newblock \emph{Fault-tolerant systems}.
\newblock Morgan Kaufmann, 2007.

\bibitem[Lin and Chen(2007)]{Lin2007}
C.-M. Lin and C.-H. Chen.
\newblock {Robust fault-tolerant control for a biped robot using a recurrent
  cerebellar model articulation controller}.
\newblock \emph{Systems, Man, and Cybernetics, Part B: Cybernetics},
  37\penalty0 (1):\penalty0 110--123, 2007.

\bibitem[Mahdavi and Bentley(2003)]{mahdavi2003evolutionary}
S.~Mahdavi and P.~Bentley.
\newblock An evolutionary approach to damage recovery of robot motion with
  muscles.
\newblock \emph{Advances in Artificial Life}, pages 248--255, 2003.

\bibitem[Mahdavi and Bentley(2006)]{mahdavi2006innately}
S.~Mahdavi and P.~Bentley.
\newblock Innately adaptive robotics through embodied evolution.
\newblock \emph{Autonomous Robots}, 20\penalty0 (2):\penalty0 149--163, 2006.

\bibitem[Metzinger(2004)]{Metzinger2004}
T.~Metzinger.
\newblock \emph{Being no one: The self-model theory of subjectivity}.
\newblock MIT Press, 2004.

\bibitem[Metzinger(2007)]{Metzinger2007}
T.~Metzinger.
\newblock Self models.
\newblock \emph{Scholarpedia}, 2\penalty0 (10):\penalty0 4174, 2007.

\bibitem[Moore(1975)]{moore1975progress}
G.~E. Moore.
\newblock Progress in digital integrated electronics.
\newblock In \emph{International Electron Devices Meeting}, volume~21, pages
  11--13. IEEE, 1975.

\bibitem[Mostafa et~al.(2010)Mostafa, Tsai, and Her]{mostafa2010alternative}
K.~Mostafa, C.~Tsai, and I.~Her.
\newblock Alternative gaits for multiped robots with leg failures to retain
  maneuverability.
\newblock \emph{International Journal of Advanced Robotic Systems}, 7\penalty0
  (4):\penalty0 31, 2010.

\bibitem[Mouret and Doncieux(2010)]{Mouret2010}
J.-B. Mouret and S.~Doncieux.
\newblock {Sferes\_v2: Evolvin' in the Multi-Core World}.
\newblock In \emph{Proceedings of IEEE Congress on Evolutionary Computation
  (CEC)}, pages 4079--4086, 2010.

\bibitem[Mouret and Doncieux(2012)]{Mouret2012div}
J.-B. Mouret and S.~Doncieux.
\newblock Encouraging behavioral diversity in evolutionary robotics: an
  empirical study.
\newblock \emph{Evolutionary Computation}, 20\penalty0 (1):\penalty0 91--133,
  January 2012.

\bibitem[Mouret et~al.(2012)Mouret, Koos, and Doncieux]{Mouret2012}
J.-B. Mouret, S.~Koos, and S.~Doncieux.
\newblock Crossing the reality gap: a short introduction to the transferability
  approach.
\newblock In \emph{Proceedings of ALIFE's workshop "Evolution in Physical
  Systems"}, pages 1--7, 2012.

\bibitem[Mulder et~al.(2008)Mulder, Hochstenbach, Dijkstra, and
  Geertzen]{Mulder2008}
T.~Mulder, J.~Hochstenbach, P.~Dijkstra, and J.~Geertzen.
\newblock {Born to adapt, but not in your dreams.}
\newblock \emph{Consciousness and cognition}, 17\penalty0 (4):\penalty0
  1266--71, 2008.

\bibitem[Nakamura et~al.(2007)Nakamura, Mori, Sato, and
  Ishii]{nakamura2007reinforcement}
Y.~Nakamura, T.~Mori, M.~Sato, and S.~Ishii.
\newblock Reinforcement learning for a biped robot based on a cpg-actor-critic
  method.
\newblock \emph{Neural Networks}, 20\penalty0 (6):\penalty0 723--735, 2007.

\bibitem[Nelson et~al.(2009)Nelson, Barlow, and Doitsidis]{nelson2009fitness}
A.~Nelson, G.~Barlow, and L.~Doitsidis.
\newblock Fitness functions in evolutionary robotics: A survey and analysis.
\newblock \emph{Robotics and Autonomous Systems}, 57\penalty0 (4):\penalty0
  345--370, 2009.

\bibitem[Nguyen-Tuong and Peters(2011)]{NguyenTuong2011}
D.~Nguyen-Tuong and J.~Peters.
\newblock {Model Learning for Robot Control : A Survey}.
\newblock \emph{Cognitive Processing}, 12\penalty0 (4):\penalty0 319--340,
  2011.

\bibitem[Palmer et~al.(2009)Palmer, Miller, and Blackwell]{palmer2009}
M.~Palmer, D.~Miller, and T.~Blackwell.
\newblock {An Evolved Neural Controller for Bipedal Walking: Transitioning from
  Simulator to Hardware}.
\newblock In \emph{Proceedings of IROS' workshop on Exploring new horizons in
  Evolutionary Design of Robots}, 2009.

\bibitem[Parker(2009)]{parker2009punctuated}
G.~Parker.
\newblock Punctuated anytime learning to evolve robot control for area
  coverage.
\newblock \emph{Design and Control of Intelligent Robotic Systems}, pages
  255--277, 2009.

\bibitem[Peters(2010)]{Peters2010}
J.~Peters.
\newblock Policy gradient methods.
\newblock \emph{Scholarpedia}, 5\penalty0 (10):\penalty0 3698, 2010.

\bibitem[Peters and Schaal(2008)]{peters2008reinforcement}
J.~Peters and S.~Schaal.
\newblock Reinforcement learning of motor skills with policy gradients.
\newblock \emph{Neural Networks}, 21\penalty0 (4):\penalty0 682--697, 2008.

\bibitem[Prassler and Kosuge(2008)]{Prassler2008}
E.~Prassler and K.~Kosuge.
\newblock \emph{Handbook of Robotics}, chapter Domestic Robotics, pages
  1253--1281.
\newblock Springer, 2008.

\bibitem[Pretorius et~al.(2012)Pretorius, du~Plessis, and
  Cilliers]{Pretorius2012}
C.~Pretorius, M.~du~Plessis, and C.~Cilliers.
\newblock Simulating robots without conventional physics: A neural network
  approach.
\newblock \emph{Journal of Intelligent \& Robotic Systems}, pages 1--30, 2012.

\bibitem[Qu et~al.(2003)Qu, Ihlefeld, Jin, and Saengdeejing]{Qu2003}
Z.~Qu, C.~M. Ihlefeld, Y.~Jin, and A.~Saengdeejing.
\newblock {Robust fault-tolerant self-recovering control of nonlinear uncertain
  systems}.
\newblock \emph{Automatica}, 39\penalty0 (10):\penalty0 1763--1771, 2003.

\bibitem[Quigley et~al.(2009)Quigley, Conley, Gerkey, Faust, Foote, Leibs,
  Wheeler, and Ng]{Ros2009}
M.~Quigley, K.~Conley, B.~P. Gerkey, J.~Faust, T.~Foote, J.~Leibs, R.~Wheeler,
  and A.~Y. Ng.
\newblock {ROS}: an open-source robot operating system.
\newblock In \emph{Proceedings of ICRA's workshop on Open Source Software},
  2009.

\bibitem[Ramachandran and Hirstein(1998)]{Ramachandran1998}
V.~Ramachandran and W.~Hirstein.
\newblock The perception of phantom limbs.
\newblock \emph{Brain}, 121\penalty0 (9):\penalty0 1603--1630, 1998.

\bibitem[Saranli et~al.(2001)Saranli, Buehler, and Koditschek]{saranli2001rhex}
U.~Saranli, M.~Buehler, and D.~Koditschek.
\newblock Rhex: A simple and highly mobile hexapod robot.
\newblock \emph{The International Journal of Robotics Research}, 20\penalty0
  (7):\penalty0 616--631, 2001.

\bibitem[Schleyer and Russell(2010)]{schleyer2010adaptable}
G.~Schleyer and A.~Russell.
\newblock Adaptable gait generation for autotomised legged robots.
\newblock In \emph{Proceedings of Australasian Conference on Robotics and
  Automation (ACRA)}, 2010.

\bibitem[Schmitz et~al.(2001)Schmitz, Dean, Kindermann, Schumm, and
  Cruse]{schmitz2001biologically}
J.~Schmitz, J.~Dean, T.~Kindermann, M.~Schumm, and H.~Cruse.
\newblock A biologically inspired controller for hexapod walking: simple
  solutions by exploiting physical properties.
\newblock \emph{The biological bulletin}, 200\penalty0 (2):\penalty0 195--200,
  2001.

\bibitem[Smola and Vapnik(1997)]{smola1997support}
A.~Smola and V.~Vapnik.
\newblock Support vector regression machines.
\newblock \emph{Advances in neural information processing systems}, 9:\penalty0
  155--161, 1997.

\bibitem[Smola and Sch{\"o}lkopf(2004)]{smola2004tutorial}
A.~Smola and B.~Sch{\"o}lkopf.
\newblock A tutorial on support vector regression.
\newblock \emph{Statistics and computing}, 14\penalty0 (3):\penalty0 199--222,
  2004.

\bibitem[Sproewitz et~al.(2008)Sproewitz, Moeckel, Maye, and
  Ijspeert]{sproewitz2008learning}
A.~Sproewitz, R.~Moeckel, J.~Maye, and A.~Ijspeert.
\newblock Learning to move in modular robots using central pattern generators
  and online optimization.
\newblock \emph{The International Journal of Robotics Research}, 27\penalty0
  (3-4):\penalty0 423--443, 2008.

\bibitem[Steingrube et~al.(2010)Steingrube, Timme, W{\"o}rg{\"o}tter, and
  Manoonpong]{steingrube2010self}
S.~Steingrube, M.~Timme, F.~W{\"o}rg{\"o}tter, and P.~Manoonpong.
\newblock Self-organized adaptation of a simple neural circuit enables complex
  robot behaviour.
\newblock \emph{Nature Physics}, 6\penalty0 (3):\penalty0 224--230, 2010.

\bibitem[Sturm et~al.(2008)Sturm, Plagemann, and Burgard]{sturm2008adaptive}
J.~Sturm, C.~Plagemann, and W.~Burgard.
\newblock Adaptive body scheme models for robust robotic manipulation.
\newblock In \emph{Robotics: Science and Systems}, 2008.

\bibitem[Sutton and Barto(1998)]{Sutton1998}
R.~S. Sutton and A.~G. Barto.
\newblock \emph{{Introduction to Reinforcement Learning}}.
\newblock MIT Press, 1998.

\bibitem[Sutton et~al.(2000)Sutton, McAllester, Singh, and
  Mansour]{sutton2000policy}
R.~Sutton, D.~McAllester, S.~Singh, and Y.~Mansour.
\newblock Policy gradient methods for reinforcement learning with function
  approximation.
\newblock \emph{Advances in neural information processing systems}, 12\penalty0
  (22), 2000.

\bibitem[Tedrake et~al.(2005)Tedrake, Zhang, and Seung]{tedrake2005learning}
R.~Tedrake, T.~Zhang, and H.~Seung.
\newblock Learning to walk in 20 minutes.
\newblock In \emph{Proceedings of Yale workshop on Adaptive and Learning
  Systems}, 2005.

\bibitem[Toffolo and Benini(2003)]{Toffolo2003}
A.~Toffolo and E.~Benini.
\newblock {Genetic diversity as an objective in multi-objective evolutionary
  algorithms}.
\newblock \emph{Evolutionary Computation}, 11\penalty0 (2):\penalty0 151--167,
  2003.

\bibitem[Togelius et~al.(2009)Togelius, Schaul, Wierstra, Igel, and
  Schmidhuber]{Togelius2009}
J.~Togelius, T.~Schaul, D.~Wierstra, C.~Igel, and J.~Schmidhuber.
\newblock {Ontogenetic and phylogenetic reinforcement learning}.
\newblock \emph{Kuenstliche Intelligenz}, pages 30--33, 2009.

\bibitem[Turing(1950)]{Turing1950}
A.~M. Turing.
\newblock Computing machinery and intelligence.
\newblock \emph{Mind}, 59\penalty0 (236):\penalty0 433--460, 1950.

\bibitem[Visinsky et~al.(1994)Visinsky, Cavallaro, and Walker]{Visinsky1994}
M.~Visinsky, J.~Cavallaro, and I.~Walker.
\newblock {Robotic fault detection and fault tolerance: A survey}.
\newblock \emph{Reliability Engineering \& System Safety}, 46\penalty0
  (2):\penalty0 139--158, January 1994.

\bibitem[Vogeley et~al.(1999)Vogeley, Kurthen, Falkai, and Maier]{Vogeley1999}
K.~Vogeley, M.~Kurthen, P.~Falkai, and W.~Maier.
\newblock {Essential functions of the human self model are implemented in the
  prefrontal cortex.}
\newblock \emph{Consciousness and cognition}, 8\penalty0 (3):\penalty0 343--63,
  1999.

\bibitem[Weingarten et~al.(2004)Weingarten, Lopes, Buehler, Groff, and
  Koditschek]{weingarten2004automated}
J.~Weingarten, G.~Lopes, M.~Buehler, R.~Groff, and D.~Koditschek.
\newblock Automated gait adaptation for legged robots.
\newblock In \emph{Proceedings of the IEEE International Conference on Robotics
  and Automation (ICRA)}, volume~3, pages 2153--2158, 2004.

\bibitem[Whiteson(2012)]{Whiteson2012}
S.~Whiteson.
\newblock Evolutionary computation for reinforcement learning.
\newblock In \emph{Reinforcement Learning: State of the Art}, pages 326--355.
  Springer, 2012.

\bibitem[Wilson(1966)]{Wilson1966}
D.~Wilson.
\newblock Insect walking.
\newblock \emph{Annual Review of Entomology}, 11\penalty0 (1):\penalty0
  103--122, 1966.

\bibitem[Yosinski et~al.(2011)Yosinski, Clune, Hidalgo, Nguyen, Zagal, and
  Lipson]{Yosinski2011}
J.~Yosinski, J.~Clune, D.~Hidalgo, S.~Nguyen, J.~Zagal, and H.~Lipson.
\newblock {Evolving Robot Gaits in Hardware: the HyperNEAT Generative Encoding
  Vs. Parameter Optimization}.
\newblock \emph{Proceedings of the European Conference on Artificial Life
  (ECAL)}, pages 1--8, 2011.

\bibitem[Zagal et~al.(2004)Zagal, Ruiz-del Solar, and Vallejos]{Zagal2004}
J.~Zagal, J.~Ruiz-del Solar, and P.~Vallejos.
\newblock Back to reality: Crossing the reality gap in evolutionary robotics.
\newblock In \emph{Proceedings of IFAC Symposium on Intelligent Autonomous
  Vehicles (IAV)}, 2004.

\bibitem[Zagal et~al.(2009)Zagal, Delpiano, and Ruiz-del Solar]{zagal2009self}
J.~Zagal, J.~Delpiano, and J.~Ruiz-del Solar.
\newblock Self-modeling in humanoid soccer robots.
\newblock \emph{Robotics and Autonomous Systems}, 57\penalty0 (8):\penalty0
  819--827, 2009.

\bibitem[Zykov(2008)]{zykov2008morphological}
V.~Zykov.
\newblock \emph{Morphological and behavioral resilience against physical damage
  for robotic systems}.
\newblock PhD thesis, Cornell University, 2008.

\bibitem[Zykov et~al.(2004)Zykov, Bongard, and Lipson]{zykov2004evolving}
V.~Zykov, J.~Bongard, and H.~Lipson.
\newblock Evolving dynamic gaits on a physical robot.
\newblock In \emph{Proceedings of Genetic and Evolutionary Computation
  Conference, Late Breaking Paper (GECCO)}, volume~4, 2004.

\end{thebibliography}

\appendix
\section{Index to Multimedia Extensions}
\label{appendix:videos}
\label{appendix:index}
See table \ref{tab:index}.
\begin{table*}
\begin{tabular}{lll}
\hline
Extension & Media type & Description                                                                                          \\
\hline
1         & Video      & The median and best gaits obtained with T-Resilience in case A (undamaged robot)                     \\
2         & Video      & The median and best gaits obtained with T-Resilience in case B (unpowered leg)                       \\
3         & Video      & The median and best gaits obtained with T-Resilience in case C (shortened leg)                       \\
4         & Video      & The median and best gaits obtained with T-Resilience in case D (Hind leg lost)                       \\
5         & Video      & The median and best gaits obtained with T-Resilience in case E (middle leg lost)                     \\
6         & Video      & The median and best gaits obtained with T-Resilience in case F (middle and hind legs lost)          \\
7         & Video      & A full T-Resilience run (25 tests, accelerated 2.5 times) in case B (unpowered leg)                  \\
8         & Video      & A full T-Resilience run (25 tests, accelerated 2.5 times) in case F (both middle and hind legs lost) \\
9         & Video      & Behavior of the reference controller in each of the six test cases                                   \\
10        & Code       & Source code (C++) for all the experiments                                                            \\
\hline
\end{tabular}
\caption{\label{tab:index}Multimedia extensions are available online at: \protect\url{http://chronos.isir.upmc.fr/~mouret/t_resilience}}
\end{table*}

\section{NSGA-II}
\label{appendix:nsga2}
\begin{figure*}
  \includegraphics[width=\textwidth]{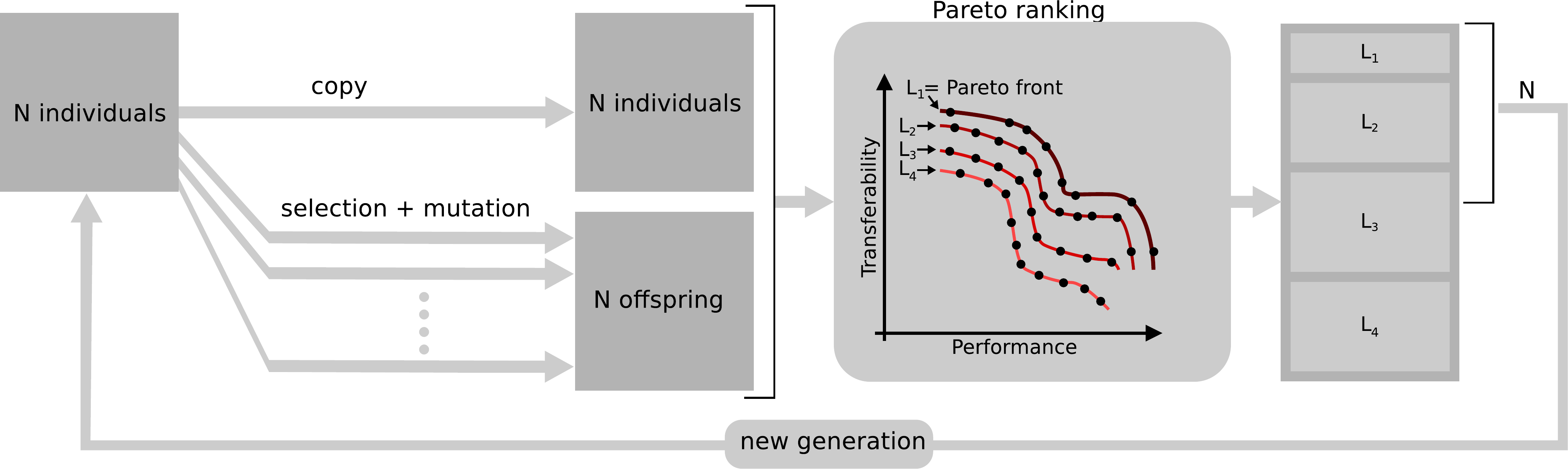}
  \caption{\label{fig:nsga2}The stochastic multi-objective
    optimization algorithm NSGA-II~\citep{deb2002fast}. Starting with
    a population of $N$ randomly generated individuals, an offspring
    population of $N$ new candidate solutions is generated using the
    best candidate solutions of the current population. The union of
    the offspring and the current population is then ranked according
    to Pareto dominance (here represented by having solutions in
    different ranks connected by lines labeled $L_{1}$, $L_{2}$, etc.)
    and the best $N$ candidate solutions form the next generation.}
\end{figure*}

Figure \ref{fig:nsga2} describes the main principle of the NSGA-II algorithm~\citep{deb2002fast}.
%% \section{Stochastic Multi-Objective Optimization}
%% \label{appendix:moea}
%% To find efficient controllers that perform similarly in the self-model
%% and in reality, the learning algorithm has to optimize both the
%% performance and the approximated transferability.

%% Similarly to many MOEAs, this
%% algorithm first generates a set of $N$ random points, called a
%% population. Then it enters a loop of four steps until a convergence
%% criteria is met (in our implementation, a fixed number of iterations is
%% performed):
%% \begin{enumerate}
%% \item Sort population with respect to dominance such that
%%   non-dominated candidate solutions are ranked $1$, those which are
%%   only dominated by non-dominated ones ranked $2$, etc. Candidate
%%   solutions that are attributed the same rank are then sorted with
%%   regard to a diversity measure (in objective space) to favor
%%   solutions in the less crowded parts.
%% \item Keep only the best $N$ solutions (during the first iteration,
%%   this step is useless).
%% \item Use the sorted population to generate new candidate solutions by
%%   perturbing the kept ones. In our implementation, solution are
%%   perturbed using polynomial mutation~\citep{Deb1994}.
%% \item Merge the newly generated candidate solutions and the previous
%%   population; this gives the new population.
%% \end{enumerate}

%% Once a good approximation of the Pareto front is reached, the choice
%% of the final solution depends on high-level criteria that is usually
%% application-dependent.

\section{Reference controller}
\label{appendix:ref}
Table \ref{table:param_hexapode} shows the parameters used for
the reference controller (section
\ref{section:robot_and_control}). The amplitude orientation parameters
($\alpha^i_1$) are set to $1$ to produce a gait as fast as possible,
while the amplitude elevation parameters ($\alpha^i_2$) are set to a
small value (0.25) to keep the gait stable. The phase elevation
parameters ($\phi^i_2$) define two tripods: 0.25 for legs 0-2-4; 0.75
for legs 1-3-5. To achieve a cyclic motion of the leg, the phase
orientation values ($\phi^i_1$) are chosen by subtracting $0.25$
to the phase elevation values ($\phi^i_2$), plus a $0.5$ shift for
legs 3-4-5 that are on the left side of the robot.
\begin{table}[H]
\centering
\begin{tabular}{|cc|cccccc|}
\hline
\multicolumn{2}{|c|}{Leg number} & 0 & 1 & 2 & 3 & 4 & 5\\
\hline
\multirow{2}*{Orientation}& $\alpha^i_1$& 1.00 & 1.00 & 1.00 & 1.00 & 1.00 & 1.00\\
&$\phi^i_1$& 0.00 & 0.50 & 0.00 & 0.00 & 0.50 & 0.00\\
\multirow{2}*{Elevation}& $\alpha^i_2$& 0.25 & 0.25 & 0.25 & 0.25 & 0.25 & 0.25\\
&$\phi^i_2$& 0.25 & 0.75 & 0.25 & 0.75 & 0.25 & 0.75\\
\hline
\end{tabular}
\caption{Parameters of the reference controller.}
\label{table:param_hexapode}
\end{table}

\section{Implementation details}
\label{appendix:details}

\subsection{Local search}

% At each step, a new controller $c'$ is derived from the current best controller $c$: $\forall j,\ c'_j = c_j + \delta_j$, where $\delta_j$ is either -0.25, 0 or 0.25 with probability $\frac{1}{3}$. Each parameter $c'_j$ remains in $[0,\ 1]$.

Our implementation of the local search (algorithm~\ref{algo:local}) starts from a randomly generated initial controller. A random perturbation $c'$ is derived from the current best controller $c$. The controller $c'$ is next tested on the robot for 3 seconds and the corresponding performance value $\mathcal{F}_{real}(c')$ is estimated with a SLAM algorithm using the RGB-D camera. If $c'$ performs better than $c$, $c$ is replaced by $c'$, else $c$ is kept.

For both the stochastic local search and the policy gradient method (section \ref{annexe:policy}), a random perturbation $c'$ from a controller $c$ is obtained as follows:
\begin{itemize}
\item each parameter $c'_j$ is obtained by adding to $c_j$ a random deviation $\delta_j$, uniformely picked up in $\{-0.25,\ 0,\ 0.25\}$;
\item if $c'_j$ is greater (resp. lower) than 1 (resp. 0), it takes the value 1 (resp. 0).
\end{itemize} 

The process is iterated during 20 minutes to match the median duration of the T-Resilience (Table~\ref{table:durations}; variant $time$). For comparison, the best controller found after 25 real tests is also kept (variant $tests$).

\begin{algorithm}
\caption{Stochastic local search ($T$ real tests)}
\label{algo:local}
\begin{algorithmic}
\STATE $c \leftarrow$ random controller
\vspace{5pt}
\FOR{$i = 1 \to T$}
\STATE $c' \leftarrow$ random perturbation of $c$
\IF{$\mathcal{F}_{real}(c') > \mathcal{F}_{real}(c)$}
\STATE $c \leftarrow c'$
\ENDIF
\ENDFOR
\vspace{5pt}
\STATE new controller: $c$
\end{algorithmic}
\end{algorithm}

\subsection{Policy gradient method}
\label{annexe:policy}

Our implementation of the policy gradient method is based on~\cite{kohl2004policy} (algorithm~\ref{algo:policy}). It starts from a randomly generated controller $c$. At each iteration, 15 random perturbations $c'^i$ from this controller are tested for 3 seconds on the robot and their performance values are estimated with the SLAM algorithm, using the RGB-D camera. The number of random perturbations (15) is the same as in \citep{kohl2004policy}, in which only 12 parameters have to be found. For each control parameter $j$, the average performance $A_{+,j}$ (resp. $A$ or $A_{-,j}$) of the controllers whose parameter value $c'^i_j$ is greater than (resp. equal to or less than) the value of $c_j$ is computed. If $A$ is not greater than both $A_{+,j}$ and $A_{-,j}$, the control parameter $c_j$ is modified as follows:
\begin{itemize}
\item $c_j$ is increased by 0.25, if $A_{+,j} > A_{-,j}$ and $c_j < 1$;
\item $c_j$ is decreased by 0.25, if $A_{-,j} > A_{+,j}$ and $c_j > 0$.
\end{itemize}

Once all the control parameters have been updated, the newly generated controller $c$ is used to start a new iteration of the algorithm.

The whole process is iterated 4 times (i.e. 60 real tests; variant $tests$) with a median duration of 24 minutes to match the median duration of the T-Resilience (Table~\ref{table:durations}). For comparison, the best controller found after 2 iterations (i.e. 30 real tests; variant $time$) is also kept.

\begin{algorithm}
\caption{Policy gradient method ($T \times S$ real tests)}
\label{algo:policy}
\begin{algorithmic}
\STATE $c \leftarrow$ random controller
\vspace{5pt}
\FOR{$i = 1 \to T$}
\STATE $\{c'^1, c'^2, \ldots, c'^S\} \leftarrow$ $S$ random perturbations of $c$
\FOR{$j = 1 \to S$}
\STATE $A_{\hspace{1pt}0\hspace{1pt},j} =$ average of $\mathcal{F}_{real}(c'^i)$ for $c'^i$ such as $c'^i_j = c_j$
\STATE $A_{+,j} =$ average of $\mathcal{F}_{real}(c'^i)$ for $c'^i$ such as $c'^i_j > c_j$
\STATE $A_{-,j} =$ average of $\mathcal{F}_{real}(c'^i)$ for $c'^i$ such as $c'^i_j < c_j$
\IF{$A_{0,j} > \max(A_{+,j}, A_{-,j})$}
\STATE $c_j$ remains unchanged
\ELSE
\IF{$A_{+,j} > A_{-,j}$}
\STATE $c_j = \min(c_j + 0.25,\ 1)$
\ELSE
\STATE $c_j = \max(c_j - 0.25,\ 0)$
\ENDIF
\ENDIF
\ENDFOR
\ENDFOR
\vspace{5pt}
\STATE new controller: $c$
\end{algorithmic}
\end{algorithm}

\subsection{Self-modeling process (Bongard's algorithm)}

\begin{algorithm*}[ht]
\caption{Self-modeling approach ($T$ real tests)}
\label{algo:bongard}
\begin{algorithmic}
\STATE $pop_{model} \leftarrow \{m^1, m^2, \ldots, m^{S_{model}}\}$ (randomly generated or not)
\STATE empty training set of actions $\Omega$
\vspace{5pt}
\FOR{$i = 1 \to T$}
\STATE \begin{tabular}{@{}l|l} selection of the action which maximises variance of predictions in $pop_{model}$ & \\
  execution of the action on the robot & \\
  recording of robot's orientation based on internal measurements (accelerometer) & (1) Self-modeling\\
  addition of the action to the training set $\Omega$ & \\
  $N_{model}$ iterations of MOEA on $pop_{model}$ evaluated on $\Omega$ & 
\end{tabular}
\ENDFOR
\vspace{5pt}
\STATE selection of the new self-model
\STATE $pop_{ctrl} \leftarrow \{c^1, c^2, \ldots, c^{S_{ctrl}}\}$ (randomly generated)
\vspace{5pt}
\STATE \begin{tabular}{@{}l|l}  $N_{ctrl}$ iterations of MOEA on $pop_{ctrl}$ in the self-model & (2) Optimization of controllers
\end{tabular}
\vspace{5pt}
\STATE selection of the new controller in the Pareto front
\end{algorithmic}
\end{algorithm*}

Our implementation of the self-modeling process is based on \cite{bongard2006resilient} (algorithm \ref{algo:bongard}). Unlike the implementation of \cite{bongard2006resilient}, we use internal measurements to assess the consequences of actions. This measure is performed with a 3-axis accelerometer (ADXL345) placed at the center of the robot, thus allowing the robot to measure its orientation.

%% To sum up, the algorithm follows these steps:

%% \begin{itemize}
%% \item[1.1.] action selection loop (exploration):
%% \begin{itemize}
%% \item each of the 36 possible actions is tested on each of the $16$
%%   candidate models to observe the orientation of robot's body
%%   predicted by the model;
%% \item the action for which models of the population disagree at most
%%   is selected;
%% \item this action is tested on the robot and the corresponding exact
%%   orientation of robot's body is recorded by an internal accelerometer;
%% \end{itemize}
%% \item[1.2.] model selection loop (estimation):
%% \begin{itemize}
%% \item a stochastic optimization algorithm (an EA) is used to optimize the population of models so that they accurately predict what was measured with the robot, for each tested action.
%% \item if less than $15$ actions have been performed, the action selection loop is started again;
%% \end{itemize}
%% \end{itemize}
%% Once $15$ actions have been performed, the best model found so far is
%% used to learn a new behavior using an EA.

%% \begin{itemize}
%% \item [2.] controller optimization (exploitation):
%% \begin{itemize}
%% \item a stochastic optimization algorithm (an EA) is used to optimize a
%%   population controllers that maximize forward displacement within the
%%   simulation of the self-model.
%% \end{itemize}
%% \end{itemize}

\paragraph{Robot's model.}

The self-model of the robot is a dynamic simulation of the hexapod
built with the Open Dynamics Engine (ODE); it is the same model as the
one used for T-Resilience experiments. However this self-model is
parametrized in order to discover some damages or morphological
modifications. For each leg of the robot, the algorithm has to find
optimal values for 5 parameters:
% the length of the two last parts of the leg (2 parameters) and if the actuators (the servos) are powered or not (3 parameters):

\begin{itemize}
\item length of middle part of the leg (float)
\item length of the terminal part of the leg (float)
\item activation of the first actuator (boolean)
\item activation of the second actuator (boolean)
\item activation of the last actuator (boolean)
\end{itemize}

The length parameters have 6 different values: \{0, 0.5, 0.75, 1, 1.25, 1.5\}, which represents a scale factor with respect to the original size.
If the length parameter of one part is zero, the part is deleted in the simulation and all other parts only attached to it are deleted too. We therefore have a model with 30 parameters.

\paragraph{Action set.}

\begin{figure}
 \centering
 \includegraphics[width=\columnwidth]{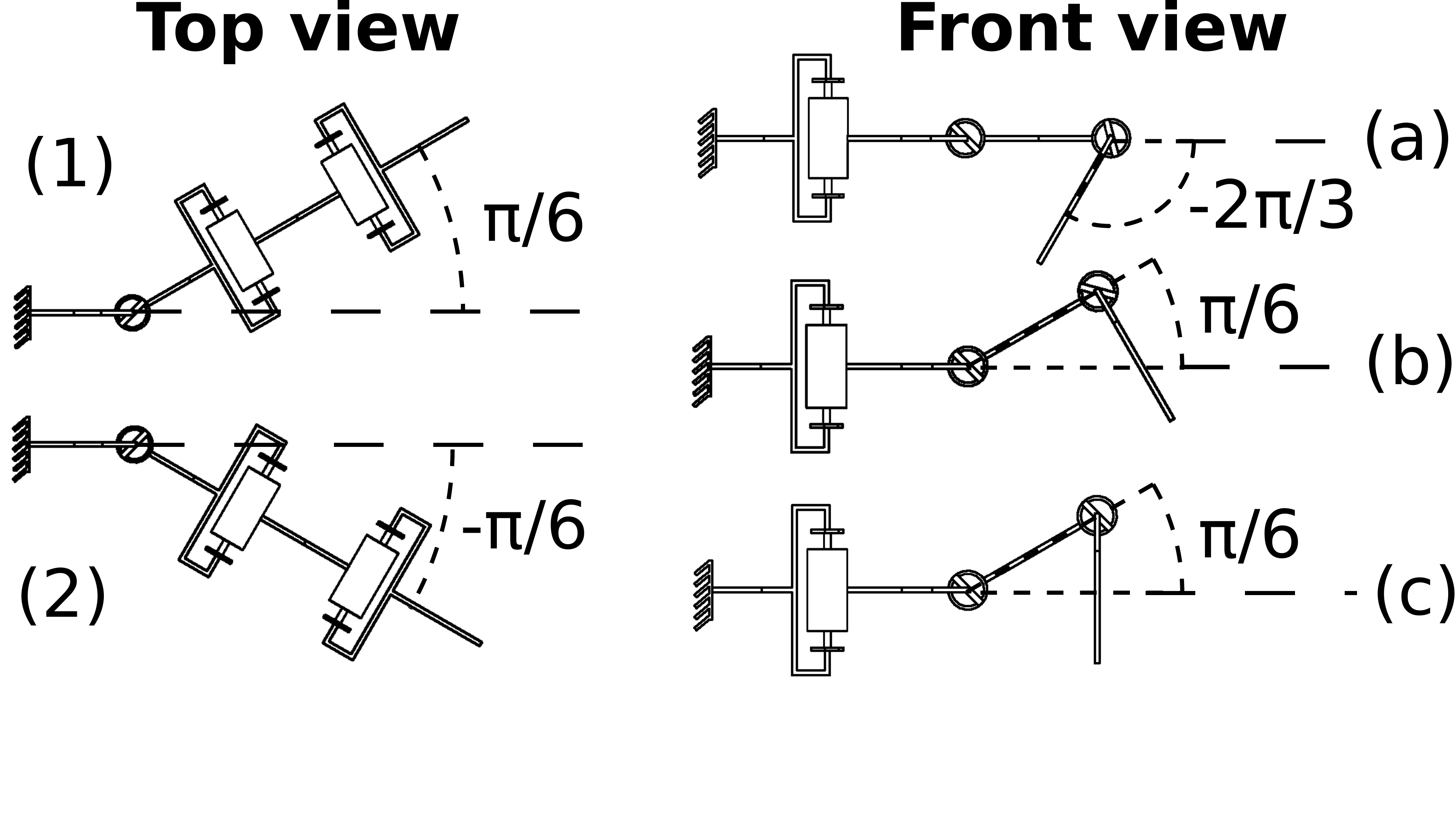}
 % hexapode.JPG: 3264x2448 pixel, 72dpi, 115.15x86.36 cm, bb=0 0 3264 2448
 \caption{The six possible actions of a leg that can be tested on the robot: (1,a), (2, a), (1,b), (2, b), (1,c), (2, c).}
\label{fig:actions}
\end{figure}

As advised by \cite{bongard2007action} (variant II), we use a set of
actions where each action uses only one leg. The first servo has 2
possible positions (1,2): $-\pi/6$ and $\pi/6$. For each of these two
positions, we have 3 possible actions (a,b,c) as shown on Figure
\ref{fig:actions}. There are consequently 6 possible actions for each
leg, that is, 36 actions in total.

\paragraph{Parameters.}

A population of 36 models is evolved during 2000 generations. The initial population is randomly generated for the initial learning scenario. For other scenarios, the population is initialized with the self-model of the undamaged robot. A new action is tested every 80 generations, which leads to a total of 25 actions tested on the real robot. Applying a new action on the robot implies making an additionnal simulation for each model at each generation, leading to arithmetic progression of the number of simulation needed per generation. Moreover, $36 \times 36$ additionnal simulations are needed each time a new action has to be selected and transferred (the whole action set applyed to the whole population). In total, about one million simulations have been done per run ($(25 \times 26/2) \times 36 \times 80 + 36\times36\times 25 = 968400$).

The self-modeling process is iterated 25 times (i.e. 25 real tests; variant $tests$) before the optimization of controllers occurs, which leads to a median duration of 250 minutes on overall. (Table~\ref{table:durations}). For comparison, the best controller optimized with the self-model obtained after 25 minutes of self-modeling is also kept (i.e. after 11 real tests; variant $time$).

%\begin{algorithm*}
%\caption{Self-modeling approach ($T$ real tests)}
%\label{algo:bongard}
%\begin{algorithmic}
%\STATE $pop_{model} \leftarrow \{m^1, m^2, \ldots, m^{S_{model}}\}$ (randomly generated or not)
%\STATE empty training set of actions $\Omega$
%\FOR{$i = 1 \to T$}
%\STATE selection of the action which maximises variance of predictions in $pop_{model}$
%\STATE execution of the action on the robot
%\STATE recording of robot's orientation based on internal measurements (accelerometer)
%\STATE addition of the action to the training set $\Omega$
%\STATE $N_{model}$ iterations of MOEA on $pop_{model}$ evaluated on $\Omega$
%\ENDFOR
%\STATE selection of the new self-model
%\STATE $pop_{ctrl} \leftarrow \{c^1, c^2, \ldots, c^{S_{ctrl}}\}$ (randomly generated)
%\STATE $N_{ctrl}$ iterations of MOEA on $pop_{ctrl}$ in the self-model
%\STATE selection of the new controller in the Pareto front
%\end{algorithmic}
%\end{algorithm*}

\section{Median durations and number of tests}
\label{appendix:median_durations}
\begin{table}[H]
  \centering
  \begin{tabular}{|c|c|c|}
    \hline
    \multirow{2}{*}{Algorithms} & Median duration & Median number\\
    & (min.) & of real tests\\
    \hline
    Local search & 20 (10) & 50 (25) \\
    Policy search & 25 (13) & 60 (30)\\
    T-Resilience & 19 (19) & 25 (25)\\
    Self-modeling& 250 (250) & 25 (25)\\
    \hline
  \end{tabular}
  \caption{Median duration and median number of real tests on the
    robot during a full run for each algorithm, for the ``time''
    variant. Number in parenthesis correspond to the ``tests''
    variant.}
  \label{table:durations}
\end{table}

\section{Statistical tests}
\label{appendix:p_values}
\begin{table}[H]
\centering
\begin{tabular}{|c|cc|cc|cc|c|}
\cline{2-8}
\multicolumn{1}{c|}{\textcolor{white}{}} & \multicolumn{2}{c|}{Local search} & \multicolumn{2}{c|}{Policy search} & \multicolumn{2}{c|}{Self-modeling} & \multirow{2}{*}{Ref.}\\
\multicolumn{1}{c|}{} &  \multicolumn{1}{c}{tests} &  \multicolumn{1}{c|}{time} &  \multicolumn{1}{c}{tests} &  \multicolumn{1}{c|}{time} &  \multicolumn{1}{c}{time} &  \multicolumn{1}{c|}{tests} &\\
\hline
A & 0.008 & 0.008 & 0.008 & 0.008 & 0.008 & 0.008 & 1.000\\
B & 0.008 & 0.016 & 0.016 & 0.016 & 0.008 & 0.008 & 0.063\\
C & 0.016 & 0.151 & 0.008 & 0.008 & 0.008 & 0.008 & 0.063\\
D & 0.151 & 0.548 & 0.016 & 0.087 & 0.063 & 0.008 & 0.063\\
E & 0.008 & 0.008 & & & & & 0.063\\
F & 0.008 & 0.063 & & & & & 0.063\\
\hline
\end{tabular}
\caption{Statistical significance when comparing performances between
  the T-Resilience and the other algorithms (Ref. corresponds to the
  reference gait). P-values are computed with Wilcoxon rank-sum
  tests.}
\label{table:pvalues_performances}
\end{table}

\end{document}